\documentclass{article}
\usepackage{PRIMEarxiv}
\usepackage[T1]{fontenc}
\usepackage[utf8]{inputenc}
\usepackage{microtype}
\usepackage{graphicx}
\usepackage[font=small,labelfont=bf]{caption}
\usepackage{booktabs}
\usepackage{longtable}
\usepackage{array}
\usepackage{tabularx}
\usepackage{amsmath,amssymb}
\usepackage{xcolor}
\usepackage{enumitem}
\usepackage{float}
\usepackage[numbers,sort&compress]{natbib}
\usepackage[hidelinks]{hyperref}
\usepackage{url}
\usepackage{orcidlink}
\usepackage{fancyhdr}
\usepackage{placeins}
\usepackage{needspace}
\usepackage{tikz}
\usepackage{listings}
\usetikzlibrary{arrows.meta,positioning,calc}
\hypersetup{
  pdftitle={Reading and Steering Representations of Materials-Science Mechanisms in an Open-Weight Language Model},
  pdfauthor={Markus J. Buehler}
}
\setlist{nosep,leftmargin=*}
\newcommand{\jlens}{Jacobian lens}

\newcommand{\code}[1]{\texttt{#1}}

\usepackage{soul}
\newif\ifshowrevisions
\showrevisionsfalse

\soulregister\textbf7
\soulregister\textit7
\soulregister\texttt7
\soulregister\emph7
\soulregister\ref7
\soulregister\pageref7
\soulregister\path7
\soulregister\code7
\soulregister\textsuperscript7

\ifshowrevisions
  \newcommand{\hlyellow}[1]{{\sethlcolor{yellow!45}\hl{#1}}}
\else
  \newcommand{\hlyellow}[1]{#1}
\fi

\soulregister\code7
\lstdefinestyle{cli}{
  basicstyle=\ttfamily\scriptsize,
  breaklines=true,
  columns=fullflexible,
  keepspaces=true,
  showstringspaces=false,
  frame=single,
  rulecolor=\color{gray!45},
  backgroundcolor=\color{gray!4},
  xleftmargin=0.5em,
  xrightmargin=0.5em,
  aboveskip=0.45em,
  belowskip=0.45em
}
\ifshowrevisions
  \lstdefinestyle{revisioncli}{style=cli,backgroundcolor=\color{yellow!20}}
\else
  \lstdefinestyle{revisioncli}{style=cli}
\fi

\title{
Reading and Steering Representations of Materials-Science Mechanisms in an Open-Weight Language Model
}
\author{
  \orcidlink{0000-0002-4173-9659} \textbf{Markus J. Buehler}$^{*}$ \\ 
  \\
  Department of Civil and Environmental Engineering \\
  Department of Mechanical Engineering \\
  Schwarzman College of Computing \\
  Massachusetts Institute of Technology \\
  Cambridge, MA, USA \\
  \\
  $^{*}$Corresponding author: \texttt{mbuehler@MIT.EDU}
}

\date{}

\begin{document}
\maketitle

\begin{abstract}
Large language models can answer scientific questions, yet a correct output does not reveal whether the model represents or uses the governing physics. Here, using three open-weight Gemma 4 models (\code{google/gemma-4-E4B-it}, \code{google/gemma-4-12B-it}, \code{google/gemma-4-31B-it}) we identify three experimentally separable signatures of materials-science mechanism information: selective concept readability, relational encoding of qualitative constitutive orientation, and causal, context-dependent control of constrained engineering answers. We combine matched direct and Jacobian vocabulary readouts, option-free state geometry, a 60-law counterfactual benchmark and causal interventions. In 50 held-out materials descriptions, three independently fitted Jacobian lenses reproduced concept ranks, and target-free word sets from both readouts enabled blinded identification of 9 of 10 mechanism families. A separate 72-prompt benchmark produced mechanism-specific hidden-state neighborhoods, but an exact graph audit showed that this apparent physical organization was equally explained by numerical comparison. We therefore compared otherwise identical prompts in which only the direction of the physical input was reversed, asking whether the resulting hidden-state movement followed the supplied constitutive law. These state transformations ordered direct, physically neutral and inverse laws across 60 frozen relations and correctly oriented 39 of 40 directional laws, whereas lexical controls were near chance. Bidirectional interventions shifted answer probabilities toward or away from the physically appropriate outcome across all 12 matched cases, while counterfactual state patches transferred opposing decision signals across mechanisms and answer formats. Physical relationships were therefore more visible in controlled state changes than in absolute states alone.
\end{abstract}


\section{Introduction}

Artificial intelligence is becoming an increasingly capable component of scientific research, both supporting and leading open-ended investigations~\citep{stanley_open_endedness_2017,wang_poet_2019,jablonka2024predictive,ghafarollahi_sciagents_2025,ghafarollahi_sparks_2025,lu_ai_scientist_2024,lu_ai_scientist_v2_2025,agarwal_autodiscovery_2025,schmidhuber_creativity_2010}. Foundation models can retrieve and synthesize literature, generate code, propose hypotheses, predict properties, and support scientific decision-making across chemistry, materials science, biology, and engineering. Yet successful behavior alone does not establish that a model represents or uses the scientific mechanism relevant to its answer. A correct prediction may arise from a physically meaningful internal representation, but it may also reflect lexical cues, memorized associations, numerical shortcuts, or features of the requested output format. Distinguishing these possibilities is important for evaluating model reliability, diagnosing failure, and developing AI systems whose scientific decisions remain robust under changes in context.

Materials science provides a particularly demanding setting in which to study this problem because scientific reasoning often requires connecting indirect evidence and mechanisms across scales. A loading history, fracture surface, heat treatment, or lattice morphology must be translated into an underlying mechanism and then used compositionally to predict a physical consequence. The governing concepts span atomic defects, mesoscale processes, and structural performance: Griffith fracture, Hall-Petch strengthening, cleavage, diffusion-controlled creep, and toughness are related, but they are not interchangeable~\citep{griffith1921rupture,hall1951deformation,petch1953cleavage,herring1950viscosity,callister2018materials,nepal_hierarchically_2023,gao2004flaw,ebrahimi2015silk,wegst_bioinspired_2015}. The scientific challenge is therefore not simply to associate a description with a familiar term, but to infer the relevant mechanism and apply its governing relation to a new material condition.

That inference contains at least two logically distinct operations. The model must first identify how an input quantity changed and then apply the appropriate constitutive relation. An increase in dislocation density raises Taylor strengthening, whereas an increase in porosity lowers elastic modulus; the same numerical direction therefore implies opposite property changes under direct and inverse laws. An interpretability measurement that recovers only that ``the stated quantity increased'' has identified a useful comparative variable, but not yet the mechanism-dependent physical consequence. Separating these stages is central to the study reported in this paper.

Artificial intelligence (AI) is increasingly part of the materials science workflow. High-throughput databases and graph models accelerate materials screening~\citep{jain2013materialsproject,merchant2023scaling,buehler_accelerating_2024,pal2026graphnativereinforcementlearningenables,buehler_preflexor_2025,doi:10.1021/acsengchemau.3c00053}; unsupervised embeddings and domain-adapted language models and multi-agent systems recover structure from the materials literature~\citep{tshitoyan2019embeddings,gupta2022matscibert,luu_bioinspiredllm_2024,hage_beamperl_2026,ghafarollahi_protagents_2024,buehler_mechgpt_2024,buehler_melm_2023,ghafarollahi_sciagents_2025,wang2026autonomousagentscoordinatingdistributed}; and language models are being tested for property prediction, low-data chemistry, and advanced graduate-level materials problem solving \citep{rubungo2023llmprop,jablonka2024predictive,yoshitake2024materialbench}. These advances establish useful behavior, but behavior alone does not reveal how a model arrived at an answer. A correct term can result from a represented physical mechanism, a conspicuous word in the prompt, memorized text, or a shallow association. Those possibilities have different implications for extrapolation, failure diagnosis, and future model training.

This distinction motivates our investigation of mechanistic interpretability. A transformer updates a high-dimensional residual vector through successive attention and feed-forward blocks \citep{vaswani2017attention}. Prior work has analyzed residual-stream communication, feed-forward memories, learned sparse features, and circuit-level pathways \citep{elhage2021framework,geva2021feedforward,bricken2023monosemanticity,templeton2024scaling,lindsey2025biology}. Other work asks whether latent knowledge or a reasoning structure can be recovered from activations \citep{burns2022latent,hou2023multistep}. A recurring warning is that a flexible probe can learn a task that the underlying representation did not make readily available; controls and matched baselines are therefore essential \citep{hewitt2019probes}.

The present study uses a deliberately constrained readout strategy. At every token position, Gemma's 42 layers update a 2,560-dimensional hidden vector; we number these transformer layers from 0 through 41 throughout. Gemma normally converts only the final vector into vocabulary scores by applying its final normalization and fixed output matrix, also called the decoder or language-model head. Intermediate vectors are not sentences. A lens reuses that fixed decoder as a measuring instrument, so the result is a vocabulary ranking rather than a free-form explanation generated by another model.

The simplest lens is direct unembedding or a logit lens where we send an intermediate state directly through the final decoder, implicitly assuming that an earlier layer already uses the final layer's coordinate system. A tuned lens learns a layer-specific translation \citep{belrose2023tuned}. The \jlens{} instead estimates how a perturbation to an earlier state propagates, on average, through the remaining network, and applies that average downstream map before the same fixed decoder \citep{gurnee2026workspace}. Direct and Jacobian readouts therefore produce the same measurable object, a ranking over Gemma's 262,144-token vocabulary, while differing only in whether downstream transport is modeled. Rank 1 is strongest; a smaller numerical rank is therefore considered better.

The Jacobian-lens study \citep{gurnee2026workspace} used Claude models and broad language tasks. Here we test whether its measurement transfers faithfully to an open Gemma checkpoint and to indirect materials science descriptions. Our aim is to determine whether this transfer enables deeper interpretation of how LLMs represent and use scientific concepts. We also test whether the resulting internal directions can steer model behavior and consider how these measurements could inform future training objectives, including reinforcement-learning reward signals.
Scientifically appropriate terms such as \emph{sensitization}, \emph{dislocation}, and \emph{toughness} may be absent from the observation; several mechanisms share words such as \emph{crack}, \emph{boundary}, or \emph{deformation}; and some technical terms split into multiple tokenizer pieces. A useful analysis must therefore distinguish a predetermined search for expected words from a target-free search for whatever vocabulary assembles naturally while the model computes an answer.

\subsection{Research Design}

We separate (i) readability, (ii) representation, and (iii) causal use. \hlyellow{The complete analysis is conducted at 4B, and selected readability, relational, and causal-coordinate tests are repeated at 12B and 31B. Direct and Jacobian vocabulary readouts recover selected scientific terms from Gemma's hidden states; matched reversals of a physical input produce state changes that track whether a constitutive relation is direct or inverse; and neutralizing or reversing the corresponding relational coordinate changes the model's engineering answers.} 
Controlled recovery asks whether a physical word declared before execution becomes highly ranked. While this approach is sensitive, it searches for a known answer. Instead open discovery supplies no preconceived answer list; it asks which words assemble repeatedly across layers, prompt positions, phrasings, and independent lens fits. Relational tests avoid choosing a vocabulary word and instead compare complete hidden vectors. We first treat every prompt as a point defined by Gemma's 2,560 internal numbers and connect it to eligible prompts from other material cases. That graph is a map of similarity among complete questions, not a graph of words or reasoning steps. It asks whether different alloys and microstructures form comparative neighborhoods and whether those neighborhoods retain the mechanism-specific sign required to translate numerical change into physical consequence. We then ask a distinct relational question: rather than interpreting either prompt state in isolation, does the change between two matched states preserve the orientation of the constitutive law when the stated numerical change is reversed? Because a readable, neighboring, or relationally organized state need not affect the answer, steering and activation patching separately perturb a frozen direction or transplant a naturally occurring state and measure the downstream engineering decision.

\begin{figure}[ht!]
\centering
\resizebox{\linewidth}{!}{
\begin{tikzpicture}[
  x=1cm,
  y=1cm,
  font=\sffamily\small,
  >={Latex[length=2.4mm,width=1.7mm]},
  outer/.style={draw=black!25,rounded corners=3.2pt,line width=0.75pt,
    fill=black!1,inner sep=0pt},
  card/.style={draw=black!28,rounded corners=2.6pt,line width=0.65pt,
    fill=white,inner sep=2.4mm},
  heading/.style={font=\sffamily\large\bfseries,text=black!88},
  subhead/.style={font=\sffamily\bfseries},
  takeaway/.style={anchor=base,align=center,inner sep=0pt,
    font=\sffamily\footnotesize\bfseries},
  flow/.style={->,draw=black!48,line width=1.2pt},
  teal/.style={draw=teal!72!black,fill=teal!72!black},
  blue/.style={draw=blue!68!black,fill=blue!68!black},
  purple/.style={draw=violet!72!black,fill=violet!72!black}
]
  \node[outer,anchor=north west,minimum width=4.75cm,minimum height=8.15cm]
    (promptbox) at (0,0) {};
  \node[outer,anchor=north west,minimum width=5.35cm,minimum height=8.15cm]
    (modelbox) at (5.30,0) {};
  \node[outer,anchor=north west,minimum width=11.75cm,minimum height=8.15cm]
    (evidencebox) at (11.20,0) {};

  \node[heading,anchor=north west] at (0.35,-0.34) {Prompt};
  \node[heading,anchor=north west] at (5.65,-0.34) {Open-weight model};
  \node[heading,anchor=north west] at (11.55,-0.34) {Interpretable evidence};

  \node[card,anchor=north west,text width=3.82cm,minimum height=4.50cm,
    align=left,font=\fontfamily{pcr}\selectfont\small] at (0.35,-1.12) {%
    \raggedright\hyphenpenalty=10000\exhyphenpenalty=10000
    Austenitic sheet aged near $650\,^{\circ}\mathrm{C}$ showed continuous
    carbide films around grains. An aggressive liquid subsequently produced
    narrow grooves along the same paths.};
  \node[anchor=north west,text width=4.00cm,align=left,font=\sffamily\footnotesize,
    text=black!68] at (0.38,-6.00)
    {Mechanism terms are absent from the description.};
  \node[anchor=north west,text width=4.00cm,align=left,font=\sffamily\footnotesize\bfseries,
    text=teal!65!black] at (0.38,-6.70)
    {Read the state at the final prompt token.};

  \node[card,anchor=north west,minimum width=4.65cm,minimum height=1.45cm]
    at (5.65,-1.10) {};
  \draw[teal,line width=0.75pt] (5.98,-1.80) -- (6.32,-1.46) -- (6.66,-1.80)
    -- (6.32,-2.14) -- cycle;
  \draw[teal,line width=0.75pt] (5.98,-1.80) -- (6.66,-1.80);
  \draw[teal,line width=0.75pt] (6.32,-1.46) -- (6.32,-2.14);
  \foreach \x/\y in {5.98/-1.80,6.32/-1.46,6.66/-1.80,6.32/-2.14}
    \fill[white,draw=teal!72!black,line width=0.65pt] (\x,\y) circle (0.095cm);
  \node[anchor=west,font=\sffamily\bfseries,text=teal!65!black]
    at (7.02,-1.62) {Gemma-4 E4B-it};
  \node[anchor=west,font=\sffamily\footnotesize,text=black!60]
    at (7.02,-2.03) {42-layer transformer};

  \draw[black!30,line width=0.8pt] (6.08,-3.10) -- (6.08,-6.75);
  \foreach \y/\c in {-3.28/teal!72!black,-4.75/blue!68!black,-6.22/violet!72!black}
    \fill[\c] (6.08,\y) circle (0.105cm);
  \node[anchor=west,align=left,text width=3.90cm] at (6.40,-3.28)
    {\textbf{Decode internal concepts}\\[-1pt]
     \footnotesize direct + Jacobian decoder};
  \node[anchor=west,align=left,text width=3.90cm] at (6.40,-4.75)
    {\textbf{Compare matched changes}\\[-1pt]
     \footnotesize reverse one physical input};
  \node[anchor=west,align=left,text width=3.90cm] at (6.40,-6.22)
    {\textbf{Intervene causally}\\[-1pt]
     \footnotesize add a direction or replace a state};

  \draw[flow] (4.80,-4.08) -- (5.22,-4.08);
  \draw[flow] (10.70,-4.08) -- (11.12,-4.08);

  \node[card,anchor=north west,minimum width=3.35cm,minimum height=6.70cm]
    (readcard) at (11.50,-1.12) {};
  \node[card,anchor=north west,minimum width=3.35cm,minimum height=6.70cm]
    (relcard) at (15.25,-1.12) {};
  \node[card,anchor=north west,minimum width=3.35cm,minimum height=6.70cm]
    (causalcard) at (19.00,-1.12) {};

  \node[subhead,text=teal!65!black] at (13.17,-1.57) {Readable};
  \draw[teal!65!black,line width=0.7pt] (11.80,-1.88) -- (14.55,-1.88);
  \node[font=\sffamily\scriptsize,text=black!56] at (13.17,-2.16)
    {boundary-attack example};
  \node[anchor=west,font=\sffamily\footnotesize] at (11.82,-2.72) {corrosion};
  \fill[teal!66] (11.82,-3.02) rectangle (14.20,-3.22);
  \node[anchor=west,font=\sffamily\footnotesize] at (11.82,-3.66) {destructive};
  \fill[teal!54] (11.82,-3.96) rectangle (13.16,-4.16);
  \node[anchor=west,font=\sffamily\footnotesize] at (11.82,-4.60) {damage};
  \fill[teal!43] (11.82,-4.90) rectangle (13.13,-5.10);
  \node[align=center,font=\sffamily\scriptsize,text=black!56,text width=2.65cm]
    at (13.17,-5.78) {three-fit consensus prominence};
  \node[takeaway,text=teal!65!black] at (13.17,-6.80) {materials terms};
  \node[takeaway,text=teal!65!black] at (13.17,-7.14) {are readable};

  \node[subhead,text=blue!64!black] at (16.92,-1.57) {Relational};
  \draw[blue!64!black,line width=0.7pt] (15.55,-1.88) -- (18.30,-1.88);
  \node[font=\sffamily\scriptsize,text=black!56] at (16.92,-2.16)
    {60-materials science laws};
  \draw[black!55,line width=0.65pt] (15.74,-5.73) -- (15.74,-2.63);
  \draw[black!55,line width=0.65pt] (15.74,-5.73) -- (18.26,-5.73);
  \draw[black!44,dashed,line width=0.55pt] (15.74,-5.02) -- (18.26,-5.02);
  \node[anchor=east,font=\sffamily\scriptsize,text=black!56] at (15.67,-5.73) {$-3$};
  \node[anchor=east,font=\sffamily\scriptsize,text=black!56] at (15.67,-5.02) {$0$};
  \node[anchor=east,font=\sffamily\scriptsize,text=black!56] at (15.67,-2.68) {$10$};
  \fill[blue!64!black] (16.18,-5.56) circle (0.105cm);
  \fill[blue!64!black] (17.00,-4.99) circle (0.105cm);
  \fill[blue!64!black] (17.82,-2.76) circle (0.105cm);
  \node[font=\sffamily\scriptsize,text=blue!64!black] at (16.18,-5.32) {$-2.27$};
  \node[font=\sffamily\scriptsize,text=blue!64!black] at (17.00,-4.75) {$0.11$};
  \node[font=\sffamily\scriptsize,text=blue!64!black] at (17.82,-2.52) {$9.67$};
  \node[font=\sffamily\scriptsize,rotate=24] at (16.18,-6.03) {inverse};
  \node[font=\sffamily\scriptsize,rotate=24] at (17.00,-6.03) {neutral};
  \node[font=\sffamily\scriptsize,rotate=24] at (17.82,-6.03) {direct};
  \node[takeaway,text=blue!64!black] at (16.92,-6.80) {state changes};
  \node[takeaway,text=blue!64!black] at (16.92,-7.14) {track relations};

  \node[subhead,text=violet!68!black] at (20.67,-1.57) {Causal};
  \draw[violet!68!black,line width=0.7pt] (19.30,-1.88) -- (22.05,-1.88);
  \node[font=\sffamily\scriptsize,text=black!56] at (20.67,-2.16)
    {raw higher-minus-lower shift};
  \draw[black!55,line width=0.65pt,<->] (19.82,-5.40) -- (21.92,-5.40);
  \draw[black!42,dashed,line width=0.55pt] (20.85,-2.70) -- (20.85,-5.48);
  \node[anchor=east,font=\sffamily\scriptsize,text=black!60] at (20.20,-3.42) {refine};
  \node[anchor=west,font=\sffamily\scriptsize,text=black!60] at (21.20,-4.50) {coarsen};
  \foreach \x/\dy in {21.163/0.00,21.522/0.08,21.631/-0.05,21.491/0.02,21.616/-0.10,21.428/0.11} {
    \draw[-{Latex[length=1.2mm,width=0.85mm]},draw=violet!48!black,line width=0.45pt]
      (20.85,{-3.42+\dy}) -- (\x,{-3.42+\dy});
    \fill[violet!70!black] (\x,{-3.42+\dy}) circle (0.085cm);
  }
  \foreach \x/\dy in {20.178/0.00,20.334/0.09,20.006/-0.05,20.241/0.02,20.569/-0.10,19.850/0.11} {
    \draw[-{Latex[length=1.2mm,width=0.85mm]},draw=violet!48!black,line width=0.45pt]
      (20.85,{-4.50+\dy}) -- (\x,{-4.50+\dy});
    \node[draw=violet!70!black,fill=white,line width=0.75pt,minimum size=0.13cm,
      inner sep=0pt] at (\x,{-4.50+\dy}) {};
  }
  \node[anchor=north,font=\sffamily\scriptsize,text=black!58] at (19.92,-5.54)
    {lower};
  \node[anchor=north,font=\sffamily\scriptsize,text=black!58] at (20.85,-5.54)
    {$0$};
  \node[anchor=north,font=\sffamily\scriptsize,text=black!58] at (21.82,-5.54)
    {higher};
  \node[takeaway,text=violet!68!black] at (20.67,-6.80) {interventions};
  \node[takeaway,text=violet!68!black] at (20.67,-7.14) {shift decisions};
\end{tikzpicture}}
\caption{\textbf{Overview of the study.} A materials-science prompt is evaluated with a frozen open-weight language model while its intermediate states are recorded. We study those states in three complementary ways: vocabulary readout tests which scientific concepts are readable; matched counterfactual comparison tests whether state changes preserve constitutive orientation; and causal intervention tests whether changing an internal representation changes the model's engineering decision. The right-hand panels illustrate the corresponding evidence types (decoded terms, relational trends, and intervention-induced answer shifts). In the causal card, arrows show the raw change in ``higher''-minus-``lower'' answer preference: refinement shifts toward ``higher'' and coarsening toward ``lower'', so the physically correct effects have opposite signs. Full experiments and statistical tests are reported in the following sections. \hlyellow{The three branches represent complementary evidence types; the diagram does not imply that one shared feature is followed across all experiments.}}
\label{fig:design}
\end{figure}
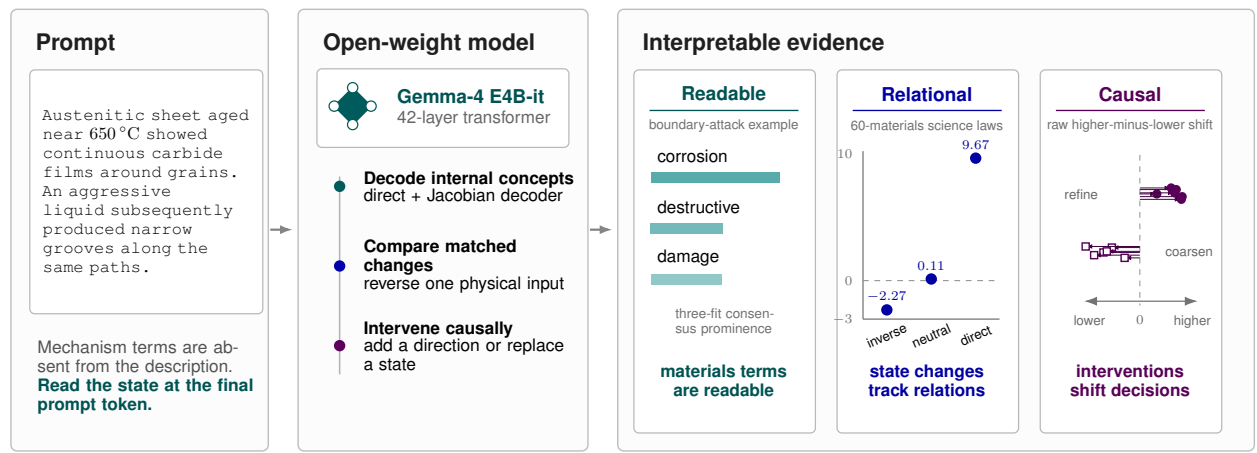

The held-out readout dataset is not 50 unrelated concepts. It contains ten mechanism families, each expressed by five independent short descriptions. The five phrasings test whether a result follows the physical situation rather than one favorable sentence. Before any held-out output was inspected, we fixed the prompts, omitted terms, final-prompt readout position, layer band, rank thresholds, candidate filters, model revision, and statistical tests. We then applied three independently fitted Jacobian lenses to the same frozen Gemma model and compared every result with direct unembedding of the identical state. Figure~\ref{fig:design} summarizes the study's progression from readable vocabulary to relational and causal evidence.

To test physical equivalence more aggressively, we later froze a separate 24-triplet cohort. Each triplet contained an anchor, a physically equivalent paraphrase with changed wording and units, and a near-verbatim counterfactual in which only the physical relation was reversed. Word and character TF-IDF both verified before model execution that the counterfactual was the closer lexical match in every triplet. The registered broad endpoint failed, but its retained layer curve motivated a new, disjoint cohort of six mechanisms and 24 material systems. The late window was then frozen before running that second cohort. This chronology lets us distinguish the failed broad claim, the exploratory late observation, and the prospective disjoint replication.

After those state arrays had been inspected, we designed a graph reanalysis. Its primary protocol and subsequent falsification amendments were each specified before their corresponding graph calculations, but the graph remains post hoc because the underlying states had been seen. A positional audit then used the same 72 scientific stems at three boundaries: the natural end of the question with no suffix, an artificial checkpoint before any answer mapping, and the final state after semantic answer choices. A frozen cross-mechanism test paired governing laws for which the same numerical increase can imply opposite property changes. We then audited the graph's identifiability by proving the exact relationship among numerical direction, law orientation, and physical outcome; matching graph shapes without labels; holding out whole mechanisms in graph learning; detecting unlabeled spectral communities; and enumerating every balanced partition. These later analyses reuse the same prompts. They ask not only whether a favorable graph exists, but what information that graph can actually identify.

The negative identifiability result motivated a different, explicitly scaffolded test of relational abstraction. One invariant prompt asked Gemma to determine the monotonic sign of a supplied equation, determine whether a numerical control rose or fell, and silently compose the two before emitting one answer word. A supervised centroid direction and layer were selected on 16 development laws and then frozen. The final 60-law benchmark balanced 20 direct, 20 inverse, and 20 physically neutral relations across 13 domains. Every law crossed two equivalent equation forms, two material cases, both numerical directions, and two answer orders. Ten neutral laws defined an empirical zero and robust scale; ten different neutral laws tested that calibration. Thus the new test breaks the earlier label alias and asks whether a controlled state displacement, rather than an absolute location or an unlabeled similarity graph, transfers constitutive orientation.

\hlyellow{The benchmarks were purposively constructed to cover a range of expert-curated materials science topics. The vocabulary suite spans ten families across failure, deformation, degradation, strengthening, and transformation, with five indirect descriptions per family, whereas the relational suite exactly balances 20 direct, 20 inverse, and 20 neutral relations across 13 physical domains, including separate neutral calibration and validation sets. These designed cohorts are not intended to be statistically representative of all materials-science mechanisms.}

\hlyellow{To test whether that relational result depended on the explicit reasoning scaffold or one checkpoint size, we removed every equation, law name, orientation label, and prescribed sign-decomposition instruction while retaining the material scenario, changed control quantity, response quantity, numerical endpoints, and constrained answer set. The same 60 law identities were then evaluated at matched normalized depth in Gemma 4B, 12B, and 31B, using a separate development-only direction in each model's own hidden coordinates. This design tests prompt-format and scale robustness of a known relational measurement; it does not ask the models to discover unfamiliar governing laws.}

The causal stage was separated chronologically and by data. First, a frozen broad screen tested three mechanism directions on 30 entirely new physical conditions, two answer-word orders, five symmetric perturbation doses, and matched controls. That screen revealed a post hoc relational hypothesis: the grain-size direction appeared to move toward higher strength after refinement but toward lower strength after coarsening. We then froze a new confirmation before any corresponding model output. It used six disjoint material pairs with matched alloy identity, grain sizes, covariates, answer words, direction, layer, doses, controls, and success criteria; only the direction of grain-size change was reversed. After those results were known, we returned to the same grain cohort for activation patching, transplanting an entire hidden state from a reversed-relation donor at one layer. A later frozen factorial patch used natural question-end states from six mechanisms, crossed answer vocabularies, and deliberately reversed the relationship between numerical direction and physical outcome. It tests whether patching transfers a general scientific relation or a narrower late numerical/decision feature. \hlyellow{Finally, a 24-law target cohort disjoint from the direction-fitting laws tested whether neutralizing, reversing, and restoring one relational coordinate changes ordinary equation-free decisions at 4B, 12B, and 31B.}

\hlyellow{A key finding is that selected materials-science terms can be reproducibly readable, although 36 of 50 prompts in the primary 4B controlled-readout suite recover no declared term within the top 100 under either readout; readable vocabulary also does not imply that the measured geometry uniquely encodes constitutive physics.} Apparent organization may be compatible with lexical identity or numerical comparison, so causal intervention is required to test whether an internal direction actually changes a scientific decision. The experiments below establish both sides of this distinction: strong boundaries on global geometric interpretation and localized, context-dependent steering.

In this paper we ask five questions:
\begin{enumerate}
    \item What scientific vocabulary is reproducibly readable, both with and without predeclared target words?
    \item Which improvements are specific to Jacobian transport rather than direct unembedding or raw Gemma states?
    \item Does option-free state geometry organize comparative relations, and can either absolute geometry or a matched state transformation retain constitutive orientation when numerical direction and governing law are separated?
    \item Can a frozen internal direction or transplanted state causally change a scientific answer?
    \item Which results transfer across phrasings, answer vocabularies, materials systems, and governing mechanisms?
\end{enumerate}

\begin{table}[ht!]
\centering
\scriptsize
\setlength{\tabcolsep}{3pt}
\caption{\textbf{Study map and evidential status.} Here ``frozen'' means that the listed endpoint and success rule were fixed before its output; it does not imply that every underlying state array or motivating cohort was unseen. ``Jacobian-specific'' requires a matched advantage over direct unembedding or a direct-direction control, rather than merely a positive Jacobian result.}
\label{tab:study-map}
\begin{tabularx}{\linewidth}{>{\raggedright\arraybackslash}p{0.22\linewidth}>{\raggedright\arraybackslash}p{0.18\linewidth}>{\raggedright\arraybackslash}X>{\centering\arraybackslash}p{0.12\linewidth}}
\toprule
Experiment & Status & Primary interpretation & Jacobian-specific? \\
\midrule
Controlled term recovery & Frozen held-out & Selective gains; family-level interval includes zero & No universal gain \\
Target-free vocabulary & Frozen extraction; secondary rating & Recognizable mechanism words for both readouts & No \\
Lexical-adversarial geometry & Frozen discovery and disjoint follow-up & Absolute endpoints fail; late-minus-middle shift is uniform & No \\
Natural question-end graph & Frozen positional robustness on inspected cohort & Within-mechanism comparative topology, AUC 0.642 & No \\
Cross-mechanism graph & Frozen falsification & Counter-numeric transfer fails, AUC 0.471 & No \\
Graph identifiability audit & Post hoc state audit; component protocols frozen & Even with numerical direction supplied, physical polarity and exact partitions fail & No \\
Neutral-anchored relational laws & Frozen 60-law cohort; direction and layer fixed earlier & Matched state changes recover narrow constitutive orientation, 39/40 directional laws & No; raw-state readout \\
Broad steering & Frozen screen & Integrated effect; only corrosion passes every family gate & Selective \\
Grain steering & Prospective frozen confirmation & Correct refinement/coarsening reversal in 12/12 conditions & Yes, in this format \\
Full-state patching & Exploratory grain study; frozen factorial follow-up & Causal late decision feature; cross-mechanism breadth fails & Lens not used to patch \\
\hlyellow{Relational-coordinate intervention} & \hlyellow{Frozen disjoint target cohort; controls amended before control output} & \hlyellow{Selective contribution, counterfactual control, and partial error repair across scale} & \hlyellow{Lens independent} \\
\bottomrule
\end{tabularx}
\end{table}

\begin{table}[ht!]
\centering
\footnotesize
\setlength{\tabcolsep}{4pt}
\caption{\textbf{Complete structure of the held-out dataset.} Every row contains five independent descriptions of the listed physical evidence. The right column gives all terms declared before execution and omitted from every corresponding input. Daggers mark documented multi-token terms excluded from the one-token rank endpoint. The Supplementary Information prints all 50 exact prompts and their terms.}
\label{tab:families}
\begin{tabularx}{\linewidth}{>{\raggedright\arraybackslash}p{0.20\linewidth}>{\raggedright\arraybackslash}X>{\raggedright\arraybackslash}p{0.29\linewidth}}
\toprule
Family & Physical evidence described without naming the mechanism & Omitted terms assessed \\
\midrule
Ductile failure & Necking, large elongation, dimples, inclusions, and joined particle-centered holes & \code{ductile}, \code{nucleation}, \code{coalescence}, \code{void} \\
Boundary attack & Stainless steel, carbide-decorated interfaces, chromium depletion, and attack following grain edges & \code{boundary}, \code{corrosion}, \code{sensitization} \\
Cyclic damage & Repeated loading, beach or arrest marks, striations, and incremental front advance & \code{fatigue}, \code{crack}, \code{propagation} \\
Cleavage & Little macroscopic plasticity, planar facets, river markings, and paths through grains & \code{brittle}, \code{cleavage}, \code{transgranular}$^{\dagger}$ \\
High-temperature deformation & Long time under heat and stress, slow strain, interface cavities, and grain-shape change & \code{creep}, \code{diffusion}, \code{cavity} \\
Particle strengthening & Nonshearable particles, bowed line defects, and reduced obstacle spacing & \code{precipitation}, \code{bowing}, \code{strengthening} \\
Rapid transformation & Fast cooling, suppressed redistribution, laths or plates, and a distorted body-centered cell & \code{martensite}$^{\dagger}$, \code{shear}, \code{Bain}, \code{tetragonal} \\
Line-defect motion & A linear lattice imperfection moving on a plane and leaving permanent offset & \code{dislocation}, \code{glide}, \code{slip} \\
Notch resistance & Energy absorption, stable notch-tip damage, flaw tolerance, and high critical intensity & \code{toughness}, \code{crack}, \code{fracture} \\
Surface oxidation & Oxygen-bearing exposure, a ceramic-like film, ion transport, mass gain, and spallation & \code{oxidation}, \code{oxide}, \code{scale} \\
\bottomrule
\end{tabularx}
\end{table}

\section{Results}

\subsection{Controlled concept recovery is reproducible but heterogeneous}

The first experiment is a frozen suite that contained 50 prompts, 150 tokenizer-resolved prompt-concept pairs, and no model-output exclusions. For each pair, we recorded the best full-vocabulary rank in the fixed 38-92\% layer band. At each cutoff $k\in\{1,2,5,10,20,50,100\}$, recovery is the fraction of declared terms ranked within the top $k$. The primary summary is area under this recovery curve against $\log k$.

\begin{figure}[ht!]
\centering
\includegraphics[width=0.98\linewidth]{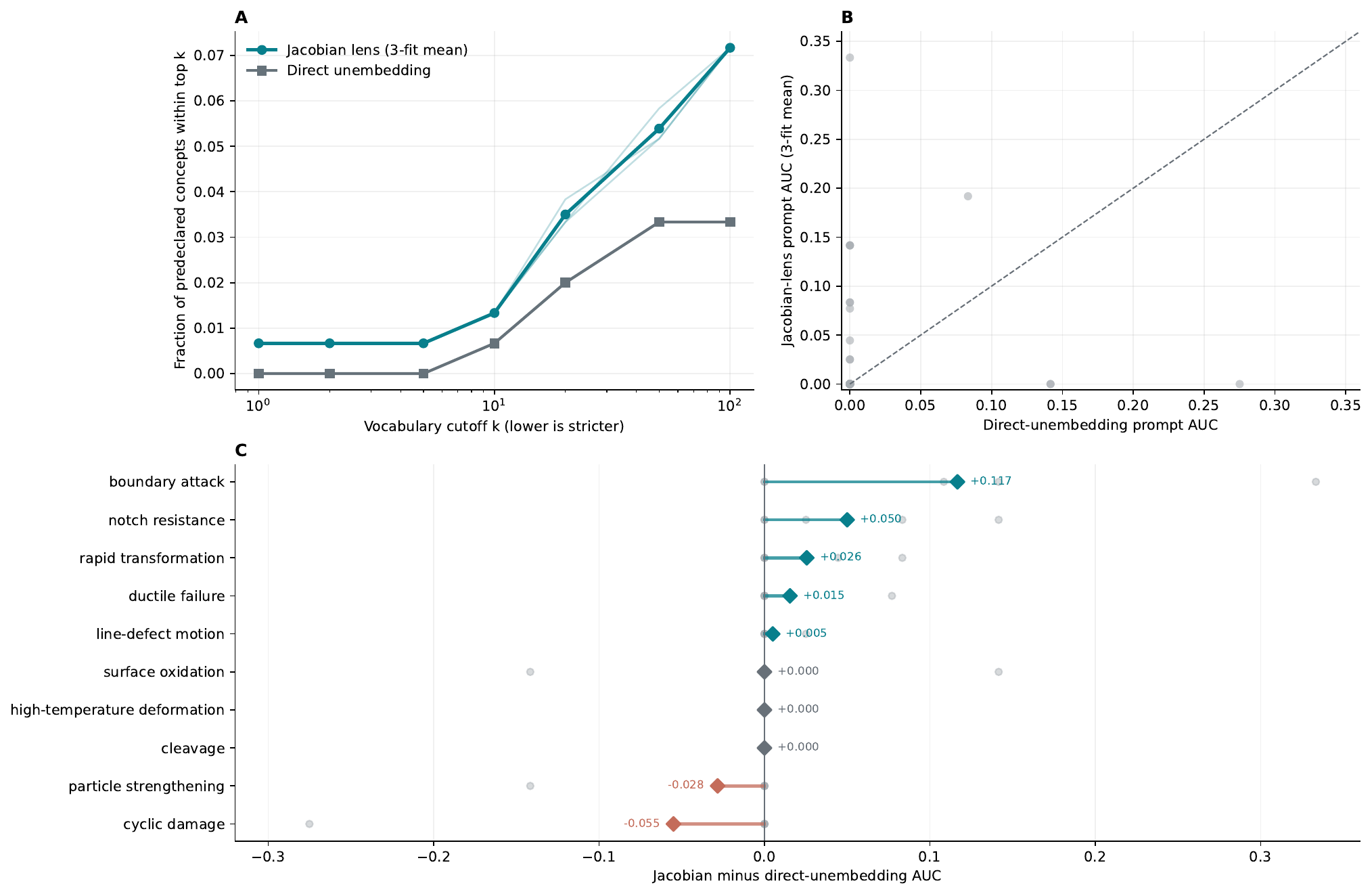}
\caption{\textbf{Can either readout recover an engineering term that was deliberately omitted from the description?} Before running the 50 prompts, we declared 150 scientifically appropriate one-token terms and then asked where each term ranked among Gemma's 262,144 vocabulary entries. (A) The cutoff $k$ is relaxed from rank 1 to rank 100; a higher curve means more declared terms have become strongly readable. Thin teal curves are the three independent Jacobian fits, the thick teal curve is their mean, and gray is direct unembedding of the same states. At $k=100$, the fractions are 7.2\% and 3.3\%, respectively. (B) Each point is one prompt's recovery area under the curve (AUC); a point above the diagonal favors Jacobian transport. Eleven prompts improve, 36 tie, and three decline. (C) The same differences are grouped at the scientifically appropriate level: gray circles are the five phrasings and diamonds are the ten family means. Positive values favor the Jacobian lens. Boundary attack and notch resistance produce the largest gains, whereas cyclic damage and particle strengthening favor direct unembedding. Across families, the mean difference is $+0.0129$, but the hierarchical 95\% interval spans $-0.0187$ to $+0.0486$ and the one-sided family sign-flip test gives $p=0.234$. Thus the experiment finds selective, mechanism-dependent gains rather than a universal Jacobian advantage.}
\label{fig:controlled}
\end{figure}

Mean Jacobian recovery AUC was 0.025765, compared with 0.012832 for direct unembedding (Figure~\ref{fig:controlled}A). The absolute difference was $+0.012933$, a 100.8\% relative increase. This average did not establish a universal advantage: a hierarchical bootstrap that resampled the ten physical families and then their five phrasings gave a 95\% interval of $-0.018674$ to $+0.048609$, and the exact one-sided family sign-flip test gave $p=0.2344$. At the prompt level, Jacobian transport won 11 comparisons, tied 36, and lost 3; these counts are descriptive because the five prompts within a family are related.

The ties are scientifically important as 36 of 50 prompts had zero recovery AUC under both readouts. Only 11 prompts had nonzero Jacobian recovery, compared with 4 under direct unembedding. Retrospective layer-robustness checks therefore asked whether the positive events were isolated spikes. Eleven prompt-concept units ever entered the Jacobian top 100 and nine stayed there for at least two consecutive sampled layers. Across families, the Jacobian advantage was $0.212$ log$_{10}$-rank units for the median layer ($p=0.0195$) and $0.197$ for the geometric mean across layers ($p=0.0117$), whereas the best-layer advantage was not significant ($p=0.197$). The effect is sparse, but most top-100 events are not one-layer accidents.

The family structure explains the uncertainty (Figure~\ref{fig:controlled}C). Boundary attack improved most ($\Delta$AUC $=+0.1167$), followed by notch resistance ($+0.0500$), rapid transformation ($+0.0256$), ductile failure ($+0.0154$), and line-defect motion ($+0.0050$). Cyclic damage ($-0.0550$) and particle strengthening ($-0.0283$) favored direct unembedding. Three families tied. The appropriate conclusion is therefore selective improvement, not a general Jacobian victory across materials mechanisms.

A post hoc influence check dropped one complete family at a time. The mean Jacobian-minus-direct AUC remained positive in all ten deletions, ranging from $+0.0014$ to $+0.0205$, but omitting boundary attack reduced the mean from $+0.0129$ to $+0.0014$. Thus no single family reverses the sign, while the size of the average gain remains strongly family dependent; this sensitivity analysis does not change the nonsignificant registered family-level test.

Individual terms show what those family averages mean physically. For a carbide-decorated austenitic sheet attacked along its grain edges, \code{corrosion} ranked 1/1/1 under the three Jacobian lenses, versus 111 under direct unembedding. For a copper specimen in which particle-centered holes enlarged and joined, \code{coalescence} ranked 24/25/18, versus 17,418. A flaw-tolerant steel placed \code{toughness} at 11/11/11, versus 12,063; a composition-preserving plate transformation placed \code{tetragonal} at 42/42/43, versus 35,954; and motion of a linear lattice imperfection placed \code{dislocation} at 87/92/94, versus 5,262. Hot-gas surface reaction was similarly specific: \code{oxidation} ranked 17/19/20, versus 195. These are not synonyms chosen after viewing the output; every term was declared before the held-out runs.

The counterexamples are equally informative. For a compressor blade with arrest lines after millions of vibration cycles, direct unembedding placed \code{fatigue} at rank 7 while the Jacobian lenses placed it at 1,279-1,345. For nonshearable particles and strongly curved line defects, direct unembedding placed \code{strengthening} at 11 versus 351-402. A cleavage prompt improved \code{brittle} from 5,290 to 222-251 and \code{cleavage} from 21,134 to 977-1,171, yet neither term entered the prespecified top 100. The aggregate therefore distinguishes spectacular readable events from families in which the same measurement is neutral or worse.

The strongest instrument-level result is reproducibility. Across all 150 controlled pairs, ranks from the three independently fitted lenses had Spearman correlations of 0.9987, 0.9980, and 0.9978 (family-clustered bootstrap lower bounds 0.9975, 0.9959, and 0.9945; Supplementary Figure~S1). By comparison, the correlation between the three-fit Jacobian mean and direct unembedding was only 0.212. Independent WikiText samples therefore led to almost identical judgments about which materials terms were easy or difficult to read.

\hlyellow{A matched supplementary audit of the same 50 prompts found higher absolute controlled readability at the larger Gemma checkpoints but no consistent Jacobian-over-direct advantage: the contrast changed sign across 4B, 12B, and 31B, and every family-hierarchical interval included zero (Supplementary Figure~S29).}

As shown in Figure~\ref{fig:controlled}A, we ask a progressively easier question as $k$ moves from 1 to 100. A term counts at $k=1$ only if it is the strongest token in the vocabulary; at $k=100$, it may have 99 stronger tokens. Figure~\ref{fig:controlled}B compares the two readouts prompt by prompt. Panel C is the inferential level: the diamonds are ten family averages, so the few highly positive boundary-attack cases cannot be mistaken for 50 independent wins. Supplementary Figure~S1 separately tests fit stability rather than validity. High fit-to-fit correlation means the instrument repeats; it does not prove that the readable term caused the model's behavior.

\subsection{Target-free decoding recovers useful materials neighborhoods}

The open-vocabulary analysis never received Table~\ref{tab:families}'s omitted terms. It scanned unrestricted top decoded tokens across all prompt positions and registered layers, removed tokens already present in the input or one-token continuation, retained 64 candidates per stored run, required agreement across all three lens records, removed a frozen list of ordinary function words, and downweighted words occurring across many prompts. The final lists were aggregated over the five phrasings in each family.

\begin{figure}[p]
\centering
\includegraphics[width=0.98\linewidth]{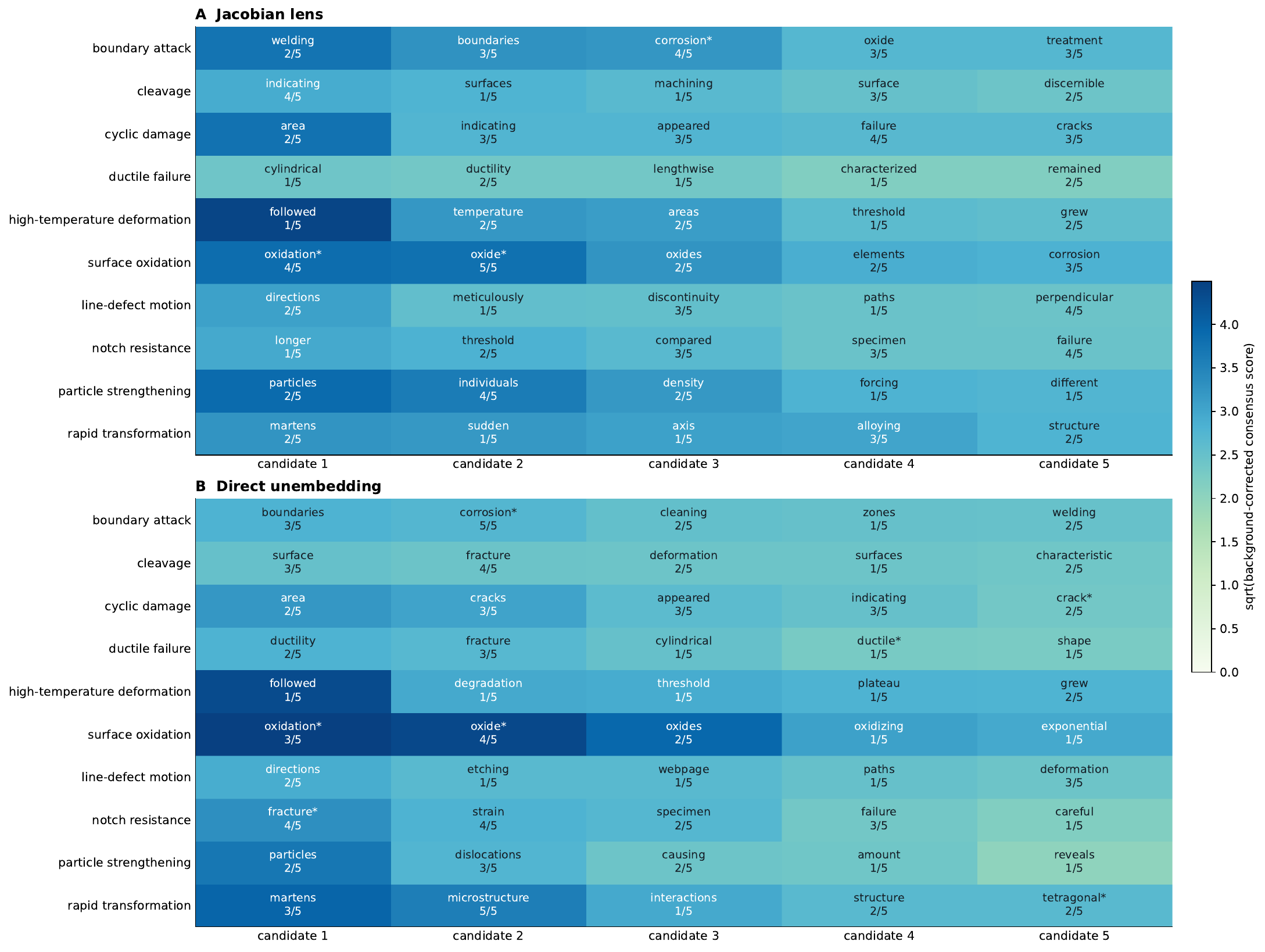}
\caption{\textbf{Target-free family vocabularies.} Each row is one physical family and each cell is a word selected without the declared concept list. Support is the number of independent phrasings, out of five, in which the word appeared under the three-fit consensus rule; darker blue-green cells indicate stronger background-corrected support. Asterisks mark exact declared-word overlap added only after ranking. Read across a row to ask whether the words form a recognizable materials neighborhood. Boundary attack, surface oxidation, and rapid transformation show coherent mechanism or microstructure terms; high-temperature deformation and cleavage show why stable decoding can still be generic. The figure therefore measures what assembles naturally, not whether the algorithm can find a supplied answer.}
\label{fig:openvocab}
\end{figure}

Figure~\ref{fig:openvocab} exposes both useful and imperfect discoveries. For boundary attack, the Jacobian list contains \code{welding}, \code{boundaries}, \code{corrosion}, \code{oxide}, and \code{treatment}; \code{corrosion} appears in four of five phrasings. Surface oxidation is the clearest lexical neighborhood: \code{oxidation} appears in four of five phrasings, \code{oxide} in all five, followed by \code{oxides}, \code{elements}, and \code{corrosion}. Rapid transformation produces fragments and neighbors rather than the predeclared full word: \code{martens}, \code{alloying}, \code{structure}, and \code{microstructure}. Ductile failure surfaces \code{ductility} but also shape words such as \code{cylindrical} and \code{lengthwise}; line-defect motion yields \code{discontinuity}, \code{directions}, and \code{perpendicular} rather than \code{dislocation} among its five leading family candidates. The notch family emphasizes \code{specimen}, \code{failure}, and \code{threshold}. These are plausible but broader neighborhoods around the physical situations.

Weak families remain visible. The high-temperature-deformation list is dominated by generic words including \code{followed}, \code{temperature}, \code{areas}, and \code{threshold}. For cleavage, \code{indicating} outranks \code{surface}, \code{machining}, and \code{discernible}. Cyclic damage contains \code{failure} and \code{cracks}, but the simpler direct readout recovers the exact word \code{fatigue} much more strongly in the controlled test. Open discovery therefore does not merely reproduce the target list: it reveals when the internal vocabulary is a coherent materials neighborhood, when it is morphological but indirect, and when it is generic.

Exact overlap with a declared word was annotated only after ranking. Such a word appeared among the top eight family candidates for 2 of 10 Jacobian lists and 6 of 10 direct lists. Exact overlap therefore favored the simpler baseline. Yet exact word matching understates semantic information: \code{boundaries} is useful for the declared singular \code{boundary}, and \code{martens} can indicate a martensitic neighborhood even though the full word is multi-token. We consequently treat family identification, not exact overlap, as the semantic endpoint.

A retrospective secondary test quantified this information at the individual-prompt level rather than asking a rater to interpret the ten family summaries. In each of five folds, a cosine nearest-centroid classifier learned from four phrasings of every mechanism and predicted the held-out fifth phrasing using only its complete target-free candidate list. Candidate vocabulary and inverse-document-frequency weights were learned inside the training fold. The strictest filter then removed every candidate that exactly matched a prompt word and conservative prefix-related variants such as \code{oxidation}/\code{oxide}. No declared concept list entered the features.

\begin{figure}[ht!]
\centering
\includegraphics[width=0.72\linewidth]{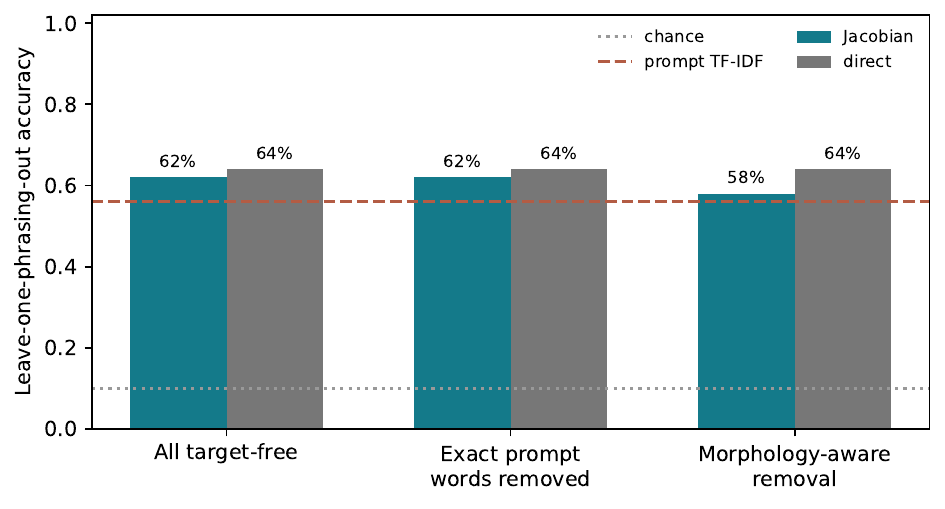}
\caption{\textbf{Do the freely discovered words identify the materials mechanism without a supplied answer list?} Each bar is leave-one-phrasing-out accuracy across the 50 prompts and ten balanced mechanism families. A classifier learned four phrasings per family and predicted the held-out fifth using only target-free candidate words from the corresponding readout. ``Exact prompt words removed'' deletes literal input copies; ``morphology-aware removal'' additionally deletes conservative prefix-related variants. The dotted line is 10\% chance. The dashed line is a TF-IDF classifier that instead sees the prompt words themselves. The target-free Jacobian and direct lists remain well above chance after the strictest removal, at 58\% and 64\%, but the prompt baseline reaches 56\% and direct does not differ significantly from Jacobian. Thus the discovered vocabulary contains cross-phrasing mechanism information, while the comparison rules out a lens-specific superiority claim. The complete confusion matrix and per-prompt predictions are in the Supplementary Information.}
\label{fig:targetfree-classification}
\end{figure}

The decontaminated Jacobian lists identified 29 of 50 prompts (58\%; macro-F1 0.568), while direct lists identified 32 of 50 (64\%; macro-F1 0.615; Figure~\ref{fig:targetfree-classification}). Both exceeded the 10\% balanced-class null after correction for the six tested readout/filter configurations ($p<10^{-4}$ for each). A prompt-word TF-IDF classifier reached 56\%, however, and direct-only correctness exceeded Jacobian-only correctness by 8 to 5 prompts (exact paired $p=0.581$). The target-free vocabulary therefore carries mechanism-family signal that survives removal of obvious prompt copies, but this analysis does not show that the Jacobian lens is superior or that all signal is independent of the input wording.

\subsection{Open-vocabulary streams localize specific materials terms}

\begin{figure}[ht!]
\centering
\includegraphics[width=0.98\linewidth]{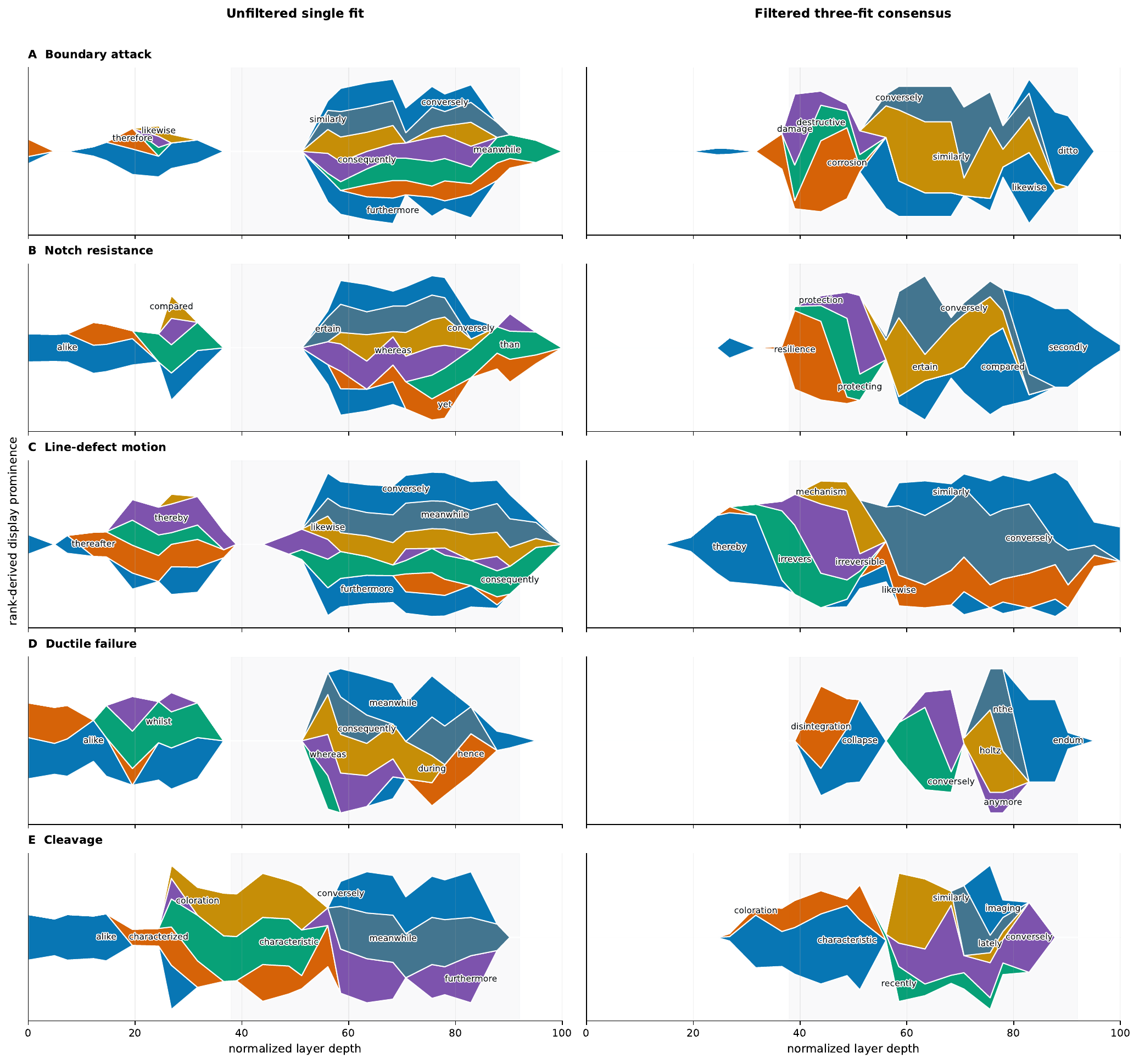}
\caption{\textbf{What does the stricter target-free filter remove, and what materials vocabulary remains?} Rows A-E correspond to the five exact prompts printed in the text. Each left panel shows the original-style unrestricted leading words from lens fit 0. Each right panel removes input and output words, the frozen 214-word function list, and the globally common readout scaffold, then requires the word to occur in all three independent fits at that layer. No declared concept was supplied to either selection. The gray region marks the fixed 38-92\% analysis band. The comparison shows that much of the visually continuous unfiltered stream is generic discourse vocabulary, while the stricter view retains localized physical neighborhoods: \code{corrosion}/\code{damage} (A), \code{resilience}/\code{protection} (B), \code{irreversible}/\code{mechanism} (C), and \code{collapse}/\code{disintegration} (D). The generic cleavage stream (E) is a useful negative case. Gaps mean that none of the displayed words passed the strict rule at that layer; they are not evidence of absent computation. Ribbon thickness is a rank-derived visualization score, not token probability, causal influence, or a literal chain of thought.}
\label{fig:streams}
\end{figure}

To show exactly what the filtering changes, Figure~\ref{fig:streams} now compares two views of the same five frozen prompts. The left view recreates the original style: it takes leading decoded words from one lens fit without removing prompt words, continuations, function words, or globally common scaffold vocabulary. The right view removes those sources of leakage and noise and retains a word at a layer only when all three independent lens fits contain it. Both views select their displayed words from the decoded rankings; neither receives a materials target list. Ribbon thickness is a rank-derived display score, not probability mass.

The five exact prompts, reproduced here so that the reader can interpret every ribbon carefully, are:
\begin{enumerate}[label=(\Alph*),leftmargin=2.4em]
    \item \textbf{Boundary attack:} ``Austenitic sheet aged near 650 degrees Celsius showed continuous carbide films around grains. An aggressive liquid subsequently produced narrow grooves along the same paths.''
    \item \textbf{Notch resistance:} ``Two steels have comparable yield stress, yet one pre-notched plate absorbs several times more energy before sudden separation and survives a much larger flaw.''
    \item \textbf{Line-defect motion:} ``Once the resolved driving stress exceeds a threshold, mobile lattice lines traverse their favored planes and produce permanent strain.''
    \item \textbf{Ductile failure:} ``A copper tensile specimen lost much of its cross-sectional area before failure. Small particle-centered holes had enlarged and joined across the final ligament.''
    \item \textbf{Cleavage:} ``A body-centered metal failed suddenly below its transition temperature. Microscopy revealed broad planar facets connected by sharp ridges through individual crystallites.''
\end{enumerate}

The paired views expose both signal and measurement noise. The unfiltered panels are dominated by connecting words such as \code{furthermore}, \code{meanwhile}, and \code{conversely}. After the frozen filters and three-fit agreement rule, prompt A contains the mechanism word \code{corrosion} together with \code{destructive} and \code{damage}; prompt B assembles \code{resilience}, \code{protection}, and \code{protecting}; prompt C contains \code{irreversible} and \code{mechanism}; and prompt D contains \code{collapse} and \code{disintegration}. These are useful engineering neighborhoods even when they do not reproduce the textbook target noun. Prompt E remains dominated by \code{characteristic}, \code{coloration}, and \code{imaging}, visually reinforcing the weak cleavage result in the family-level discovery analysis. Residual fragments and discourse words in the filtered panels are not hidden: they delimit what this readout can support.

Blank or narrow intervals in the filtered column have a precise procedural meaning. None of the seven displayed words survived all filters and appeared in all three fits at that depth. They do not mean that the model was inactive or contained no other readable word; conversely, continuity of a ribbon does not mean that the model composed a sentence. The plots show when particular vocabulary directions are linearly readable as depth changes, not that one word caused the next.

The original development stream that motivated this display is retained as Supplementary Figure~S2. Its dimpled-fracture prompt placed \code{coalescence} among the unrestricted leading decoded tokens, together with generic connective words. The stream renderer was not given a target list, but \code{coalescence} had been declared separately when that development prompt was designed. The example is visually useful for showing how vocabulary assembles and where noise enters; it is not an unanticipated held-out discovery and is not used for population statistics.

\hlyellow{A matched 31B single-lens case series reuses these same five prompts and paired layout in Supplementary Figure~S30. Its left column applies the identical unfiltered renderer; because only one 31B lens is available, its decontaminated right column uses a documented prompt/output, English-frequency, and across-prompt specificity filter instead of three-fit consensus.}

These ribbons are an exploratory open-vocabulary readout, not a literal chain of thought. They show which vocabulary directions are readable under a linear measurement as layer depth changes. They do not reveal private prose, a serial reasoning transcript, or the model's subjective experience.

\subsection{Mid-layer states carry mechanism geometry that remains partly lexical}

\begin{figure}[ht!]
\centering
\includegraphics[width=0.98\linewidth]{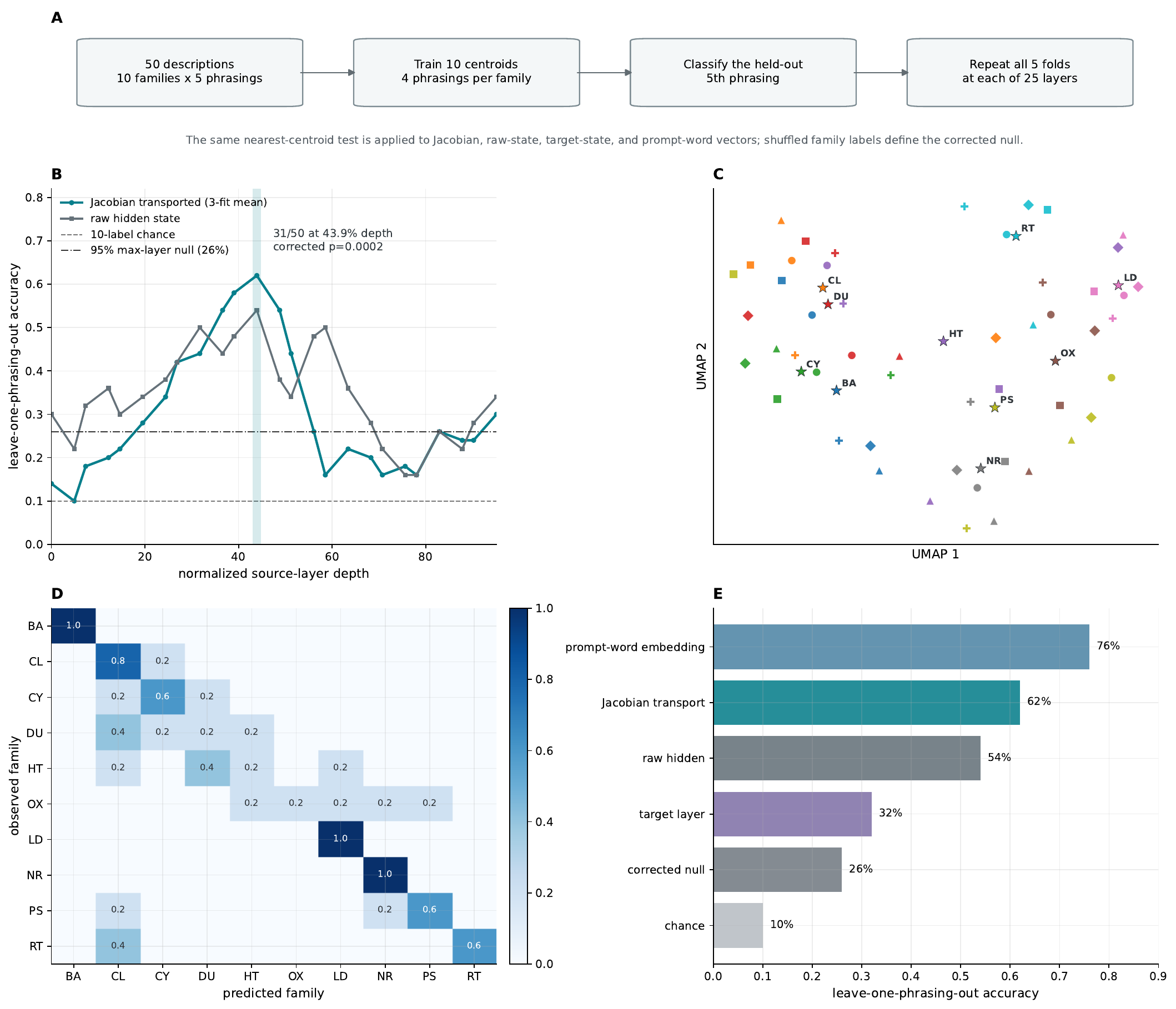}
\caption{\textbf{Do differently worded descriptions of the same mechanism remain close in the model's full state space?} (A) For each of five folds, four phrasings per family define ten centroids and the fifth phrasing is classified; the procedure is repeated at every registered layer and for every baseline representation. (B) Full-dimensional Jacobian classification peaks at 62\% at 43.9\% depth: 31 of 50 descriptions are correct, above the 13-of-50 threshold obtained after shuffling family labels and retaining the best of 25 layers ($p=0.0002$). (C) UMAP is only a display of those 50 best-layer states. Its axes have no physical units and no test uses them; color and two-letter code denote family, marker shape denotes phrasing, and stars are family centroids. (D) The full-dimensional confusion matrix reveals the physical specificity: boundary attack (BA), line-defect motion (LD), and notch resistance (NR) are separated across all five phrasings, while ductile failure (DU), high-temperature deformation (HT), and surface oxidation (OX) overlap. CL, CY, PS, and RT denote cleavage, cyclic damage, particle strengthening, and rapid transformation. (E) The same classifier reaches 54\% on raw hidden states and 76\% on a simple average of the prompt's own word embeddings. The result therefore supports statistically reliable family structure in the transported states, but it does not show that Gemma created that structure independently of the engineering vocabulary in the prompt.}
\label{fig:geometry}
\end{figure}

This experiment asks a different question from word recovery: after the wording changes, do descriptions of the same physical mechanism still occupy nearby regions of the model's full 2,560-dimensional state space? At each layer, we used four phrasings from each family to define that family's average location, then assigned the held-out fifth phrasing to whichever family average was closest. We repeated this five times so that every description was held out once (Figure~\ref{fig:geometry}A). This ``train on four, test on the fifth'' design prevents a description from helping to define the centroid used to classify itself.

After the target-free results were known, we froze this separate exploratory geometry protocol before extracting hidden vectors. For all 50 descriptions and 25 registered source layers, we retained the final-prompt hidden state, transported it through each of the three Jacobian maps, and averaged the normalized transported states. Classification used the original 2,560 coordinates, not the two-dimensional UMAP display and not the declared word list. The identical nearest-centroid procedure was applied to raw hidden states, target-layer states, and prompt-word embeddings so that the comparison changes the representation rather than the classifier.

Classification peaked at 62\% at source layer 18 (43.9\% normalized depth; Figure~\ref{fig:geometry}B): 31 of 50 descriptions were assigned to the correct family even though each test description was excluded when its family centroid was formed. A balanced 5,000-permutation test retained the maximum across all 25 layers on every shuffle; the 95th percentile of that corrected null was 26\% (13 of 50), and the plus-one $p$-value was 0.0002. Individual lens fits all peaked at 62\%. The raw hidden state reached 54\% (27 of 50) at the same layer, and the target-layer state reached 32\% (16 of 50). Thus Jacobian transport exposes a middle-layer family geometry beyond chance and modestly above the raw state under this classifier.

The confusion matrix identifies where that geometry is strongest. All five boundary-attack, line-defect-motion, and notch-resistance descriptions were correctly classified, and four of five cleavage descriptions were correct. Cyclic damage, particle strengthening, and rapid transformation each gave three of five. Only one of five descriptions was correct for each of ductile failure, high-temperature deformation, and surface oxidation. The result is therefore not a uniformly separated materials map: some mechanisms form robust cross-phrasing clusters, whereas morphologically related or lexically variable families overlap.

The lexical baseline is the essential restraint: the normalized mean of the prompt's input-token embeddings classified 76\% (38 of 50). Because the descriptions intentionally contain different physical evidence, their words already reveal family identity. The transported geometry does not outperform that information. Figure~\ref{fig:geometry}E should therefore be read in two steps: 62\% versus the corrected 26\% null is evidence that middle-layer states carry family structure; 76\% versus 62\% shows that the present design cannot separate that structure from the engineering wording that entered the model. Moreover, the best same-family versus other-family cosine margin to the frozen discovered-word vectors was negative at every layer (maximum $-0.0108$). The full vectors organize by family, but they do not simply converge toward the few displayed candidate words.

Panel C uses UMAP only to display the 50 states at the best layer \citep{mcinnes2018umap}; no inference uses its two axes. The main run's five-neighbor trustworthiness was 0.911, and five alternative seeds ranged from 0.917 to 0.945. The UMAP display was chosen after a joint all-layer diagnostic proved dominated by depth progression, so it is post hoc and visually descriptive. The classification, confusion matrix, permutation test, distance ratios, and baselines all use the full-dimensional vectors.

\subsection{Physical-equivalence geometry shifts late but remains answer-scaffold sensitive}

\hlyellow{The experiments are conducted as follows: (1) an initial frozen cohort tested the broad 38-92\% endpoint, which failed; (2) its retained late-layer rise was treated as post hoc discovery; (3) a disjoint cohort tested a prospectively frozen 80-96\% primary endpoint, which failed, and a frozen late-minus-middle secondary endpoint, which passed; and (4) the subsequent answer-scaffold audit was post hoc.}

Family classification can succeed because two descriptions share either physics or words. We therefore constructed a harder test in which those signals conflict. Each of 24 frozen triplets contained (i) an anchor, (ii) a differently worded and unit-converted paraphrase with the same scientific answer, and (iii) a near-verbatim counterfactual with the opposite answer. For example, reducing non-shearable obstacle spacing and describing the same narrowing in micrometers both imply higher Orowan stress, whereas reversing the numerical change implies lower stress. Word and character TF-IDF both selected the physical counterfactual as the anchor's closer lexical neighbor in all 24 triplets before model execution.

At each layer we centered the final-normalized decoder-basis states across all 72 prompts and calculated
\[
m_\ell=\cos(a_\ell,p_\ell)-\cos(a_\ell,c_\ell),
\]
where $a$, $p$, and $c$ denote anchor, physical paraphrase, and lexical counterfactual. Negative $m_\ell$ means that surface wording dominates; positive $m_\ell$ means that physical equivalence dominates. The initial cohort's registered 38-92\% Jacobian mean was $-0.690$ ($95\%$ interval $-0.917$ to $-0.474$), with 0/24 triplets and 0/6 family means positive. Direct decoding was similarly negative at $-0.662$ ($-0.908$ to $-0.441$). Thus the broad physical-equivalence hypothesis failed decisively.

The fully retained layer curve nevertheless rose sharply near the output. We treated that pattern as discovery, froze an 80-96\% late window, and then ran a disjoint cohort containing obstacle spacing/Orowan stress, porosity/modulus, pearlite spacing/strength, dislocation density/strength, stiff-particle fraction/modulus, and crosslink density/rubbery modulus. The prospective primary late-window Jacobian margin was $+0.365$ ($-0.159$ to $+0.791$), with 17/24 triplets and 4/6 family means positive. It therefore failed its registered confidence and breadth gates. The frozen secondary change from the 38-70\% middle window to the 80-96\% late window was much more consistent: $+1.402$ ($0.951$-$1.793$), positive in all 24 triplets and all six families. Clean registered-pair accuracy was 87.5\%, and 16/24 triplets gave all three scientifically consistent answers.

This transition is not a Jacobian-specific advantage. Direct decoding produced a nearly identical late-minus-middle change of $+1.403$ ($0.961$-$1.791$), and the late-window Jacobian-minus-direct contrast was only $+0.0039$ ($-0.0092$ to $+0.0153$). The result therefore concerns a broad change in Gemma's late state under two matched vocabulary readouts, not superiority of the fitted transport.

A post hoc audit further limits interpretation. We compared the same scientific stems immediately before any answer choices with their ordinary final states after the prompt explicitly supplied the semantic choices such as \code{higher}/\code{lower}. These quantities are positive-minus-negative decoder-logit differences, equivalently pairwise log-odds units, rather than the centered-cosine margins reported above. At 95.1\% depth before the choices, relation separation was $+0.144$ log-odds units ($-0.109$ to $+0.380$) under direct decoding and $+0.152$ ($-0.093$ to $+0.386$) under the Jacobian ensemble. After the ordinary answer scaffold, final-state separation was $+11.757$ ($5.687$-$17.307$); the paired gain over the pre-choice direct state was $+11.612$ ($5.813$-$17.030$) and was positive in every family. The compared positions and suffixes differ by design, so this audit cannot assign the gain uniquely to answer words, final-position computation, or instruction following. It does show that the replicated late transition alone is insufficient evidence for an option-free internal physical variable.

\begin{figure}[ht!]
\centering
\includegraphics[width=0.98\linewidth]{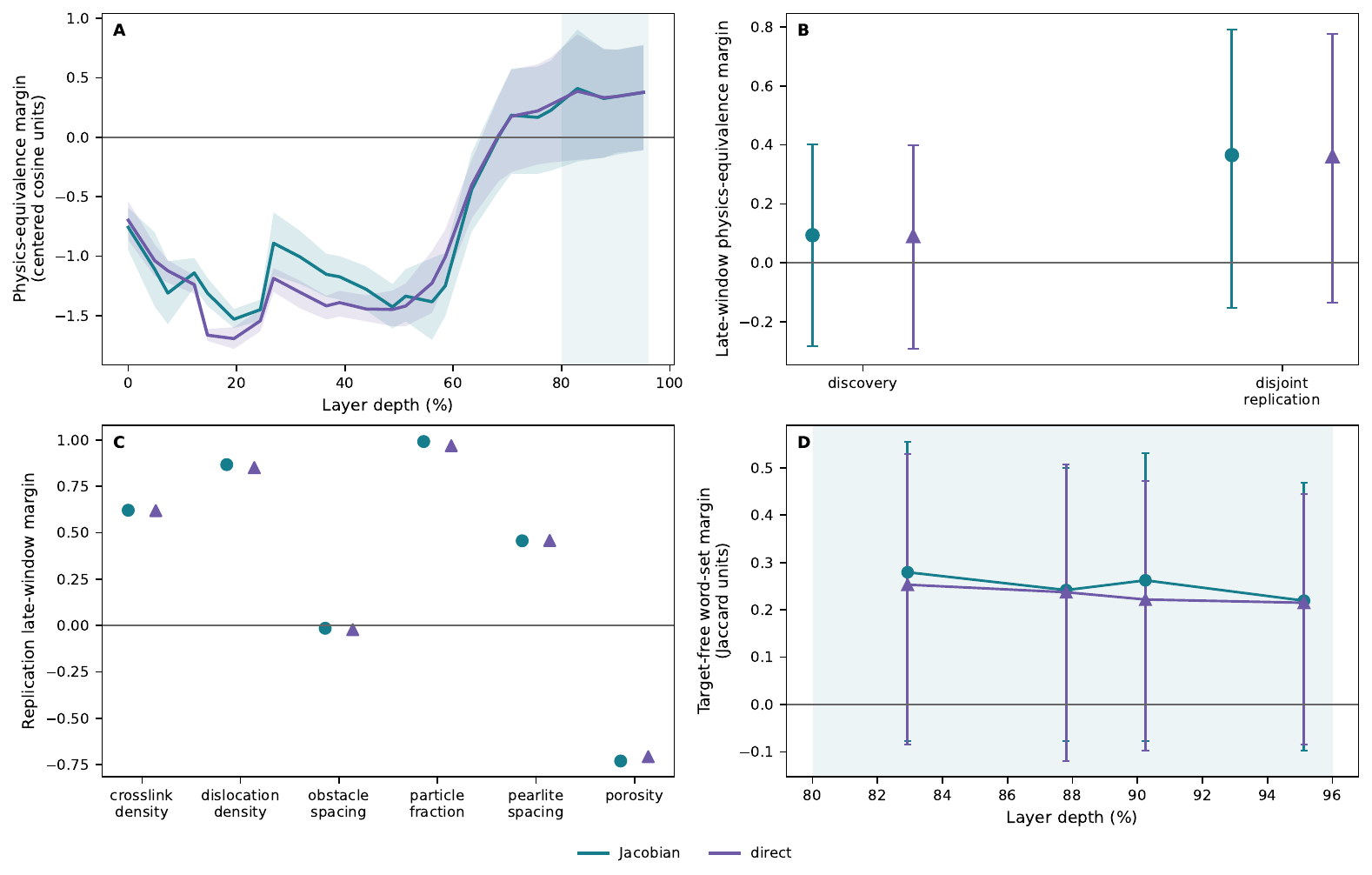}
\caption{\textbf{Does physical equivalence overcome deliberately stronger wording similarity?} Every triplet compares an anchor with a differently worded prompt that has the same physical answer and a near-verbatim prompt whose physical relation is reversed. Positive margin means that the same-physics pair is closer; negative margin means that wording dominates. (A) In the disjoint replication cohort, both Jacobian and direct decoder-basis states begin strongly wording-aligned and rise sharply after roughly 60\% depth. The shaded region is the prospectively frozen 80-96\% late window. The frozen late-minus-middle Jacobian change is $+1.402$ ($0.951$-$1.793$), positive in 24/24 triplets and 6/6 families. (B) The absolute late-window means are positive in both the discovery and disjoint cohorts, but their family-hierarchical intervals cross zero; the replication's registered primary endpoint therefore failed. (C) Four replication families end positive, obstacle spacing is approximately neutral, and porosity remains wording-aligned, demonstrating real mechanism heterogeneity. (D) A target-free comparison of decoded word sets also trends positive late, but its intervals cross zero. Teal and purple nearly coincide throughout, showing that the transition is not uniquely produced by Jacobian transport. This is a representational similarity test, not a causal intervention or a literal chain of thought.}
\label{fig:physical-equivalence}
\end{figure}

\subsection{Option-free graphs reveal conditional comparative structure}

The failed one-to-one triplet endpoint above asks whether an anchor is closer to its own paraphrase than to its near-verbatim counterfactual. A graph asks a different population question: when a prompt is compared with other material cases, which cases become its nearest neighbors? Every node is a complete question, represented by Gemma's 2,560 internal numbers at one depth; nodes are not individual words and edges are not reasoning steps. Within each supplied mechanism family, a node selects the most similar different material case in each of the other two surface-variant groups, producing 144 directed edges. An edge is provisionally called correct when it joins cases with the same registered physical-outcome label. No target vocabulary word is selected.

The original graph was measured after the prompt displayed semantic answer choices. It was strong (81.9\% same-direction edges and all-candidate AUC 0.816), but that state could already attend to the answer words. We therefore froze a positional robustness test on the same 72 scientific stems and captured the natural final token of the complete question, with no choices, answer words, A/B code, response instruction, or checkpoint marker. At this natural boundary, 97 of 144 Jacobian edges (67.4\%) preserved the registered binary relation, compared with a 55.6\% exact structured-null mean ($p=0.02195$). Full-candidate AUC was 0.642 ($p=0.01235$), and five of six mechanism-family AUCs exceeded 0.5. Direct unembedding reached AUC 0.628 and raw states 0.594, so the option-free comparative structure is a property of Gemma's state geometry rather than a Jacobian-specific gain. The identifiability audit below shows that the registered within-family relation is exactly aliased with numerical direction.

\begin{figure}[ht!]
\centering
\includegraphics[width=0.99\linewidth]{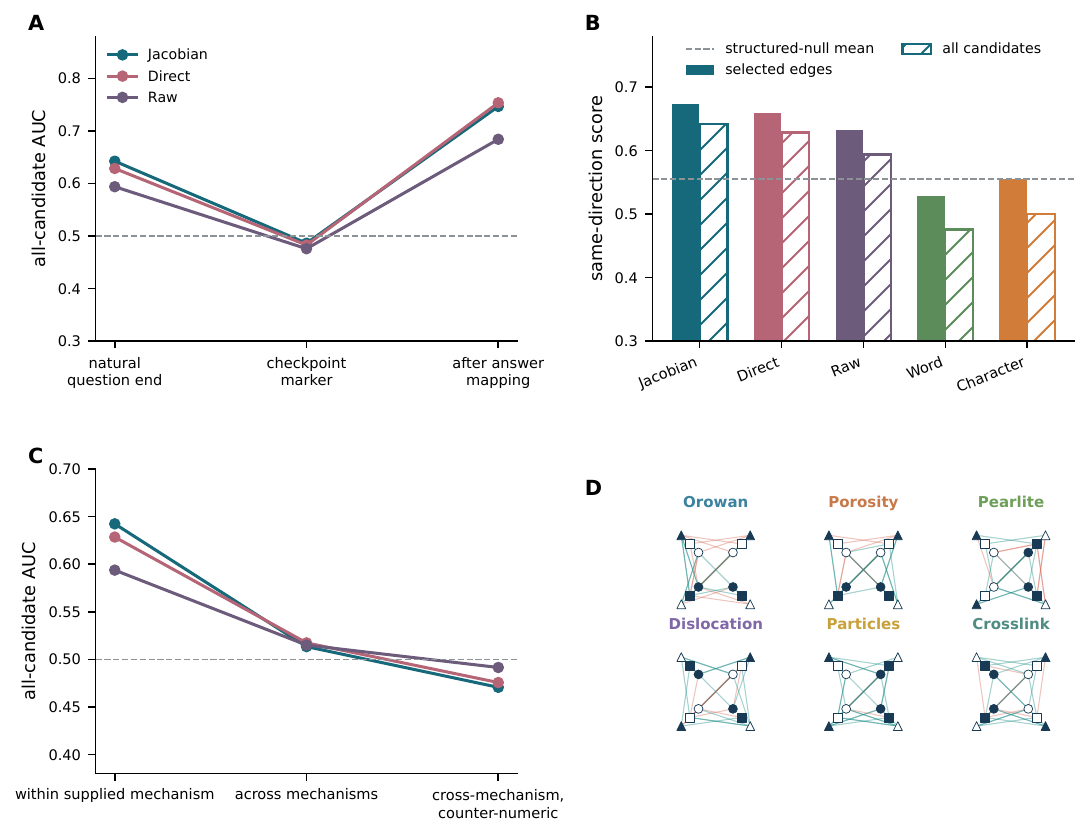}
\caption{\textbf{Where does option-free comparative graph structure appear, and where does it fail?} Every node is one complete materials question and every selected edge points to the most similar eligible different material case. AUC uses all eligible candidates rather than only the selected edge; 0.5 denotes no ordering information. (A) The same 72 scientific stems are read at three prompt boundaries. At the natural question end, no answer word or response scaffold is present and Jacobian AUC is 0.642. A checkpoint-suffixed variant shows no detectable signal, while the explicit answer-mapping condition reaches 0.816. Because suffix and token position differ across conditions, this comparison localizes prompt-boundary sensitivity rather than a unique causal effect of the added words. Direct unembedding and raw states follow the same pattern. (B) At the natural question end, solid bars show the fraction of 144 selected edges that preserve the registered binary relation and hatched bars show all-candidate AUC. The dashed line is the exact case-preserving null mean. Jacobian selected-edge precision is 0.674 ($p=0.02195$) and AUC is 0.642 ($p=0.01235$); word and character TF-IDF are at or below the null. (C) The positive result depends on supplying the governing mechanism. When candidates come from other mechanisms, AUC falls to 0.513; when raw numerical direction deliberately conflicts with physical outcome, it falls to 0.471. (D) The actual 72-node natural-question graph contains six supplied mechanism islands and 144 selected edges. Circles, squares, and triangles are anchor, paraphrase, and near-verbatim counterfactual prompts. Filled and open nodes are the two registered orientations; teal and coral edges preserve or reverse the registered binary relation. Ninety-seven edges are teal. This figure demonstrates an option-free comparative regularity; Figure~\ref{fig:graph-identifiability} determines what that regularity can identify.}
\label{fig:relation-graph}
\end{figure}

Position matters sharply (Figure~\ref{fig:relation-graph}A). The checkpoint-suffixed condition before any answer mapping had Jacobian AUC 0.486, whereas the full answer-mapping condition reached 0.816. Thus the final-state result is not fabricated by displayed answer order (the original answer-order-only control was 0.333) but its magnitude is answer-conditioned. A scientifically meaningful yet weaker neighborhood is already present at the natural question end, the checkpoint-suffixed condition is null, and explicit answer choices strongly amplify the measured relation. Because suffix and token position change together, the comparison does not isolate a unique cause.

Figure~\ref{fig:relation-graph}D displays the actual natural-question graph. The six islands are not discovered communities: they appear because the candidate rule supplies the mechanism family. Circles, squares, and triangles denote anchors, paraphrases, and counterfactuals; filled and open nodes denote the two registered physical-outcome orientations; teal edges preserve the registered binary relation and coral edges do not. The complete 25-layer atlas contains 3,600 selected edges, and its union contains 391 unique directed edges. Both are shown in the Supplementary Information so that the graph can be inspected as a network rather than inferred only from a scalar summary.

The supplied-family restriction motivates the first generalization test. We compared every prompt with cases governed by different mechanisms, including nine mechanism pairs for which the relationship between raw numerical direction and property direction is reversed. Within a supplied mechanism, Jacobian AUC was 0.642; across mechanisms it fell to 0.513, and in the counter-numeric subset it was 0.471 (both exact $p=0.10$; Figure~\ref{fig:relation-graph}C). Only 3 of 9 opposite-response mechanism pairs exceeded 0.5. Direct and raw states failed similarly. The graph therefore does not establish one universal ``property increases'' axis across different governing laws.

\subsection{Graded comparison structure transfers without identifying physical polarity}

The preceding graph result initially appears to organize a physical outcome. A complete identifiability audit reveals a more precise interpretation. For node $i$ in mechanism $f$, let $x_{fi}=+1$ for a stated numerical increase and $-1$ for a decrease, $s_f=+1$ for a direct-response law and $-1$ for an inverse-response law, and $y_{fi}$ denote the registered physical outcome. The prompt manifest obeys $y_{fi}=s_fx_{fi}$ exactly. Consequently, for two cases governed by the same mechanism,
\begin{equation}
y_{fi}y_{fj}=x_{fi}x_{fj}.
\end{equation}
Within a family, ``same versus different physical outcome'' is therefore exactly the same target as ``same versus different numerical direction.'' Moreover, flipping the unobserved law orientation $s_f$ reverses every absolute physical label without changing the unlabeled graph. The graph can reveal a comparative two-group structure, but it cannot by itself say which group means ``property increases.'' This is a proof about what this benchmark can identify, not a claim that Gemma lacks physical information.

An additional post hoc continuous-state check prevents overgeneralizing that algebraic result. Across all nine direct-law-inverse-law mechanism pairs, the Jacobian contrast between same-physical/opposite-numerical similarity and same-numerical/opposite-physical similarity was $-0.000204$ (95\% pair-bootstrap interval $-0.000679$ to $+0.000298$; exact $p=0.445$). Direct, raw, and lexical representations were likewise null. The exact within-family alias therefore limits what the graph endpoint identifies; it does not show that numerical direction universally dominates continuous state geometry.

We tested that distinction without generating new prompts or rerunning Gemma. First, label-blind graph matching learned a case correspondence from anchor-paraphrase similarities and evaluated it on held-out counterfactual similarities before revealing labels. Similarity form transferred across all 15 mechanism pairs: mean held-out correlations were 0.976 for Jacobian, 0.973 for direct decoder-basis, and 0.992 for raw states. Yet the Jacobian maps agreed with physical case identity only 60.0\% of the time ($p=0.135$), no mechanism pair was exactly isomorphic under the surface-variant labels, and the approximate-isomorphism result did not survive multiplicity correction ($q=0.0861$). The transferable object is a graded comparison pattern, not a common physical map.

Second, we trained graph isomorphism networks while holding out an entire mechanism. The absolute-polarity network received graph statistics, surface-variant identity, and the stated numerical increase/decrease sign, but no prompt text, mechanism identity, material identity, answer words, or physical label. Predicting the supplied numerical sign back from these features was a pipeline positive control and was necessarily perfect (mean AUC 1.000; 6/6 held-out mechanisms); it is not evidence that graph topology discovered that sign. The scientific test was whether the network could combine the supplied sign with the graph to determine whether the property rises or falls under an unseen governing law. It could not: absolute physical-polarity AUC averaged 0.259, with only 1/6 mechanisms above 0.5. A separate polarity-free same/different relation network excluded numerical sign and surface variant as well as all semantic metadata. It evaluated the 72 eligible ordered different-case, cross-variant pairs within each held-out 12-node mechanism graph and also failed its frozen gate (mean AUC 0.507; 2/6 mechanisms above 0.5). The two results are therefore not contradictory: the absolute network reads back an input it was given, whereas the relation network asks whether sparse topology alone preserves the within-mechanism pairing. Below-chance physical AUC is consistent with recovering a coherent split but assigning the two unnamed sides the wrong orientation in a new law.

Third, we asked whether the full weighted graph forms a unique physical partition. Retaining all graded similarities improved label-blind spectral agreement from ARI 0.267 for the sparse top-one graph to 0.674, but direct states matched 0.674 and raw states reached 0.817; none of the 12 density-by-representation tests survived false-discovery correction (best $q=0.092$). An exhaustive search over all 462 balanced six-versus-six partitions then found that the registered split was not uniquely preferred in this cohort: Jacobian mean ARI was 0.022 ($p=0.087$), and fitting anchors/paraphrases before assigning held-out counterfactuals gave $-0.027$ ($p=0.620$). These finite-cohort results do not prove absence: six mechanisms provide limited population power, and the near-threshold $q=0.092$ and $p=0.087$ leave open a moderate effect in a larger, prospectively balanced mechanism set. They do show that positive pairwise ranks, visually coherent edges, and a favorable spectral projection need not imply one transferable global physical partition in the present data.

\begin{figure}[ht!]
\centering
\includegraphics[width=0.99\linewidth]{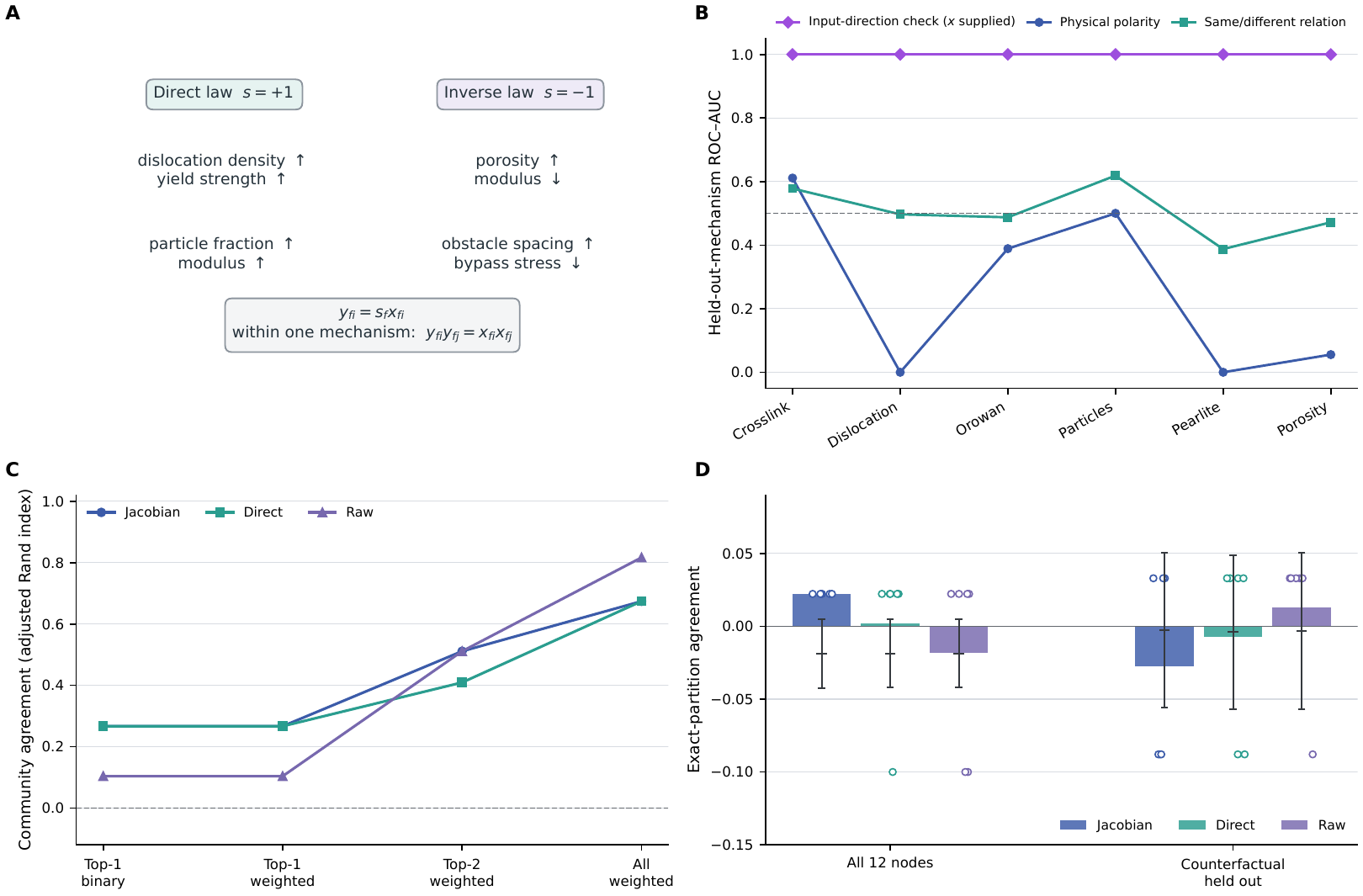}
\caption{\textbf{What can the option-free graph identify?} (A) A numerical increase raises the property for a direct law but lowers it for an inverse law. In this prompt set, physical polarity is exactly $y=sx$; within one mechanism, the same/different physical relation collapses algebraically to the same/different numerical relation. (B) A graph isomorphism network is trained on four mechanisms, selected on a fifth, and tested on an unseen sixth. The purple AUC of 1.0 is a positive-control input readback: numerical direction $x$ was deliberately supplied to the absolute-label network, so this curve does not show graph discovery. The actual test asks whether graph structure can combine that supplied $x$ with an unseen law to infer physical polarity $y$; it fails, as does a separate same/different relation network that receives neither $x$ nor surface-variant identity. Points are held-out mechanisms and connecting lines guide the eye. (C) Community agreement rises when graded similarities are retained instead of only the strongest edge, showing that sparsification discards structure. Direct and especially raw states are at least as strong as the Jacobian states, so this organization is not lens-specific. (D) Exhaustively evaluating every balanced partition shows that the registered physical split is not the uniquely preferred division, including when counterfactual phrasing is held out. Bars are means, dots are mechanisms, and intervals show the corresponding blockwise null distribution. Together, the panels support transferable comparative geometry while rejecting a universal, Jacobian-specific physical-sign graph.}
\label{fig:graph-identifiability}
\end{figure}

\subsection{Matched state changes reveal neutral-anchored constitutive orientation}

The preceding audit concerns absolute states and unlabeled similarity structure. It does not test whether a physical abstraction is expressed as a reproducible transformation between controlled states. We therefore constructed a different benchmark around the elementary materials relation
\begin{equation}
\text{physical-response sign}
=
\text{constitutive-law sign}
\times
\text{numerical-change sign}.
\end{equation}
The same numerical increase must move the response upward for a direct law, downward for an inverse law, and leave it unchanged for a physically neutral relation. The prompt supplied the equation and asked Gemma to perform these two stages silently; it did not request or expose a chain of thought. The 60-law cohort contained 20 direct, 20 inverse, and 20 neutral relations spanning mechanics, fracture, fatigue, transport, thermal, electrical, magnetic, optical, interfacial, microstructural, corrosion, fluid, and kinetic domains. Each law generated 16 prompts by crossing two algebraically equivalent equation surfaces, two material cases, numerical increase and decrease, and two answer orders.

\begin{figure}[ht!]
\centering
\includegraphics[width=0.99\linewidth]{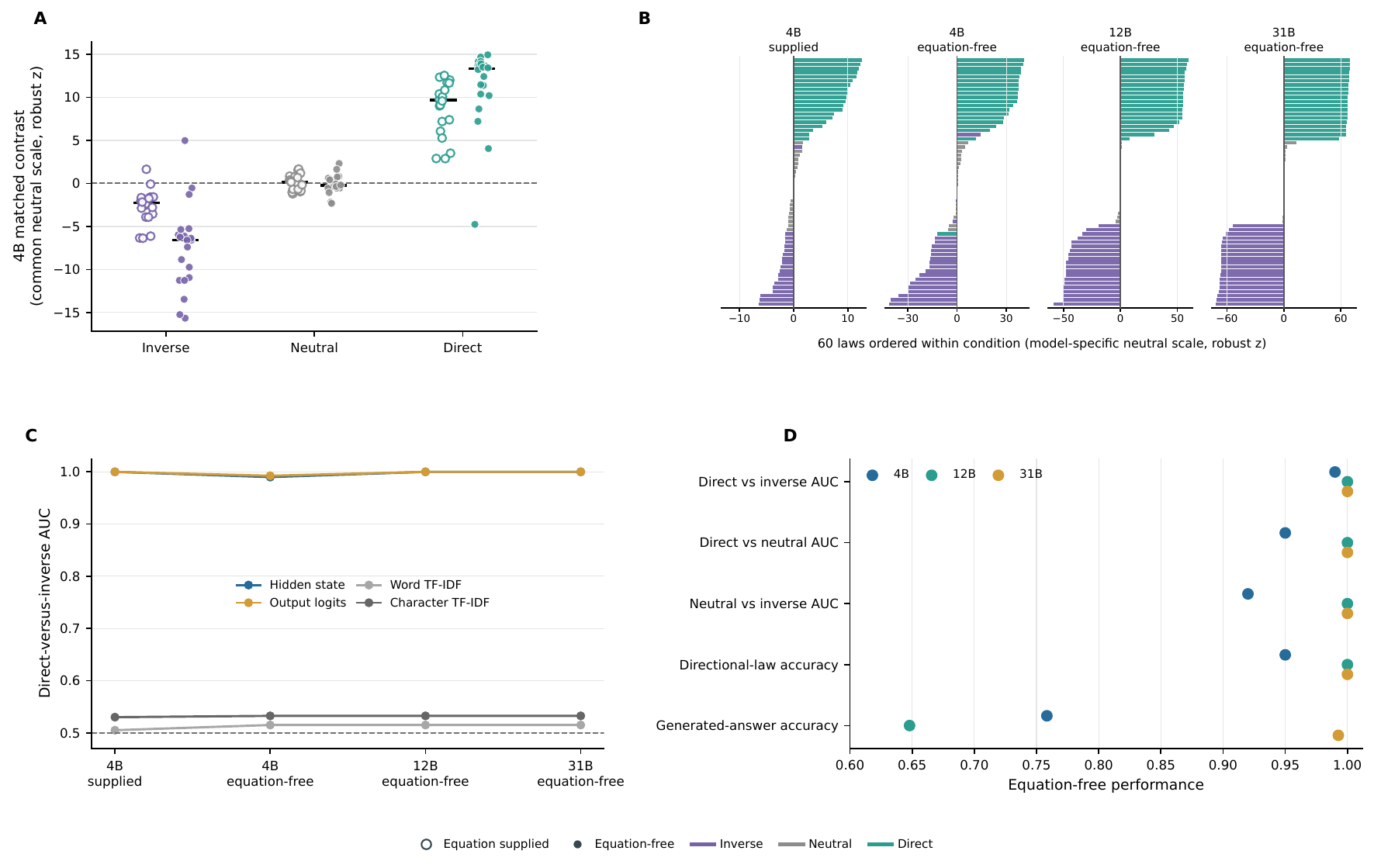}
\caption{\hlyellow{\textbf{Constitutive orientation persists without a supplied equation across Gemma scale.} Every law contributes eight matched up-minus-down state contrasts along a model-specific direction fitted only on the same earlier development laws. (A) The matched 4B prompt-format comparison. Open markers supplied the equation and prescribed sign composition; filled markers used the corresponding equation-free questions. Both use the original 4B neutral scale, and black segments are medians. (B) The complete 60-law distributions for every available condition. Each waterfall independently orders all laws from most inverse to most direct; horizontal bars begin at that condition's calibrated neutral zero. The 4B supplied and equation-free facets are the matched format pair. The 12B and 31B facets contain equation-free tests only. Each facet has its own horizontal robust-$z$ scale, so signs and category ordering are comparable across facets but absolute bar lengths are not common hidden-state units. Purple, gray, and teal denote inverse, neutral, and direct relations. (C) Direct-inverse AUC for the hidden-state direction, output-logit change, and lexical controls across the same four available conditions. Hidden and output separation survives scaffold removal and reaches 1.000 at 12B and 31B, whereas word and character controls remain near chance. (D) The three registered pairwise AUCs, neutral-cut directional accuracy, and exact generated-answer accuracy for the equation-free cohort. Both larger checkpoints orient 40 of 40 directional laws and reach ceiling on all three relational AUCs; only 31B converts that representation into nearly perfect generated answers. The comparison supports a scale-robust ordinal constitutive orientation, not an invariant law-by-law magnitude, unfamiliar-law generalization, or a literal reasoning trace.}}
\label{fig:relational-physics}
\end{figure}

Before defining the 60-law protocol, we froze the matched-reversal endpoint and tested it on a disjoint 12-law confirmation cohort. It separated direct from inverse relations with AUC 0.972 (exact balanced-label $p=0.004329$); only afterward were the 60-law cohort, neutral calibration, endpoints, and random seeds frozen. The readout was therefore fixed before this cohort was executed. A unit direction at layer 34 was defined by the difference between positive- and negative-physical-outcome centroids on 16 earlier laws; no 60-law state was used to refit it. For each equation-surface, material-case, and answer-order cell, we subtracted the projected score of the numerical-decrease prompt from its matched numerical-increase prompt. Eight such comparisons were averaged within each law. This subtraction holds the equation, material wording, numerical values, and answer order fixed; only the stated direction of numerical change reverses. This is a paired contrast between two controlled prompt states; it requires neither a network layout nor a nearest-neighbor graph. \hlyellow{Figure~\ref{fig:relational-physics}A shows the complete 4B supplied-equation and equation-free law distributions on one common calibration.}

Raw zero was not assumed to mean physical neutrality. Ten calibration-neutral laws defined a median matched contrast of 0.01360 and a robust scale of 0.00854 ($1.4826\times$ the median absolute deviation). Ten different neutral laws were then evaluated without contributing to either quantity. Relative to this physical null, category-median robust scores were $-2.266$ for inverse laws, $+0.108$ for all neutral laws, and $+9.669$ for direct laws \hlyellow{(Figure~\ref{fig:relational-physics}B, first waterfall).} Direct versus inverse AUC was 1.000 (balanced-law bootstrap 95\% interval 1.000-1.000); direct versus validation-neutral AUC was 1.000; and validation-neutral versus inverse AUC was 0.935 (0.820-1.000). The neutral-median cutoff classified 39 of 40 directional laws. The single sign error was the classical nucleation-barrier relation versus undercooling.

The result is not a formatting or lexical artifact under the registered controls. Direct-inverse AUC was 1.000 on the explicit equation surface and 0.975 on the rearranged surface. Across inverse, validation-neutral, and direct laws, the neutral-scaled score followed the registered ordinal relation with Spearman $\rho=0.910$ (100,000-label one-sided Monte Carlo permutation $p=0.000010$). Word and character TF-IDF directions trained on the same earlier prompts reached only 0.505 and 0.530 direct-inverse AUC, whereas the model's output-logit change reached 1.000 (Figure~\ref{fig:relational-physics}C). Exact answer production was less reliable: direct-law answers were 90.0\% correct, but inverse-law answers only 53.8\%. The relational ordering inside the model is therefore substantially more stable than its conversion into the requested discrete inverse-law answer.

\hlyellow{We next performed a prompt-format transfer test to determine whether this ordering depended on being shown the equation and the prescribed two-stage decomposition. Before executing the corresponding model runs, we replaced every prompt with a natural materials question that removed the equation, law name, direct/inverse/neutral label, and instruction to compose two signs. The material scenario, response and control quantities, numerical endpoints, all-other-conditions-fixed clause, answer vocabulary, and answer order were retained. The same layer-34 direction and midpoint were applied without refitting. Because the 60 law identities were reused, this is a complete scaffold-removal test over the established cohort rather than an independent fresh-law replication.}

\hlyellow{The relational ordering survived this removal. Equation-free direct-inverse AUC was 0.990 (bootstrap 95\% interval 0.9625-1.000), direct-validation-neutral AUC was 0.950, and validation-neutral-inverse AUC was 0.920. The ordinal association remained strong ($\rho=0.877$, permutation $p=0.000010$), and the neutral cutoff oriented 38 of 40 directional laws correctly. The two exceptions were capillary rise versus surface tension, which had transferred under the supplied equation but reversed under natural wording, and nucleation barrier versus undercooling, which failed in both formats. Output-logit change reached direct-inverse AUC 0.9925, whereas word and character TF-IDF remained near chance at 0.515 and 0.5325. Strict generated-answer accuracy was lower at 75.8\% overall (92.2\% direct, 70.3\% inverse, and 65.0\% neutral); all 45 off-format continuations were retained as errors. The stable state transformation therefore transfers beyond the explicit reasoning scaffold more reliably than the model converts it into a compliant answer.}

\hlyellow{This finding adds a narrower but deeper representational level to the study. Vocabulary readouts identify which scientific terms are readable, and absolute graphs identify which complete prompts become neighbors. The matched comparison instead asks how the state changes under a controlled physical reversal. Transfer across equation forms and then across complete scaffold removal supports an operational abstraction of monotonic constitutive orientation: Gemma's late state changes differently under the same numerical reversal for direct, neutral, and inverse relations, even when their equations are no longer printed in the prompt. The result uses a task-elicited raw-state direction rather than Jacobian transport and does not show that Gemma stores one context-free ``inverse law'' vector or discovers an unfamiliar law. It also does not recover coefficients, units, conservation laws, a full equation, or a hidden verbal rationale. What it shows is that the relevant qualitative relationship is more consistently visible in a controlled transformation between states than in their absolute global arrangement.}

\hlyellow{We then tested whether this equation-free relational measurement persists across model scale. The identical 128-prompt development manifest and 960-prompt, 60-law equation-free test manifest were applied to the 12B and 31B Gemma-4 checkpoints. A model-specific direction was fitted only on the unchanged 16 development laws at the layer nearest 83\% normalized depth (layers 34, 39, and 49 for 4B, 12B, and 31B) and applied without test refitting. At 12B and 31B, direct-inverse, direct-validation-neutral, and validation-neutral-inverse AUC were all 1.000, all 40 directional laws had the expected sign, and ordinal $\rho$ was 0.930 ($p=0.000010$). The corresponding 4B equation-free values were 0.990, 0.950, and 0.920, with 38 of 40 correct signs and $\rho=0.877$. Paired law bootstraps place the 12B/31B improvement over 4B at $+0.010$ [0.000, 0.0375] for direct-inverse AUC, $+0.050$ [0.000, 0.150] for direct-neutral AUC, $+0.070$ [0.000, 0.185] for neutral-inverse AUC, and $+0.050$ [0.000, 0.125] for directional accuracy. Thus the tested larger checkpoints remove the two 4B sign errors, while the intervals retain equality at their lower boundary because the finite 60-law cohort is already near ceiling.}

\hlyellow{Figure~\ref{fig:relational-physics}B restores the complete ordered-law view for all four available test conditions. The 4B supplied-equation and equation-free waterfalls form the only matched prompt-format pair. The 12B and 31B waterfalls are equation-free replications because the complete supplied-equation 60-law test was not run at those sizes.}

\hlyellow{The scale result is specifically relational; word and character TF-IDF remain at 0.515 and 0.5325 direct-inverse AUC because the prompt corpus is identical and those controls do not use Gemma. Output-logit change is 0.9925, 1.000, and 1.000 across 4B, 12B, and 31B, showing that the ordinal variable is also available to each checkpoint's answer decoder. Exact generated-answer accuracy does not follow a monotonic scale trend: it is 75.8\%, 64.8\%, and 99.3\%, respectively; the 12B model orients 95.9\% of both direct and inverse prompts but produces the required neutral answer for only 2.5\%. Overall cross-scale law-score Spearman correlations are high (0.854-0.897), but they fall to 0.218-0.439 after removing the three category means. The robust result from this experiment is therefore the ordinal constitutive orientation and not an invariant law-by-law magnitude or a monotonic claim about exact answer behavior.}

\subsection{Automated blinded interpretation finds equal semantic accuracy}

\begin{figure}[tbp]
\centering
\includegraphics[width=0.98\linewidth]{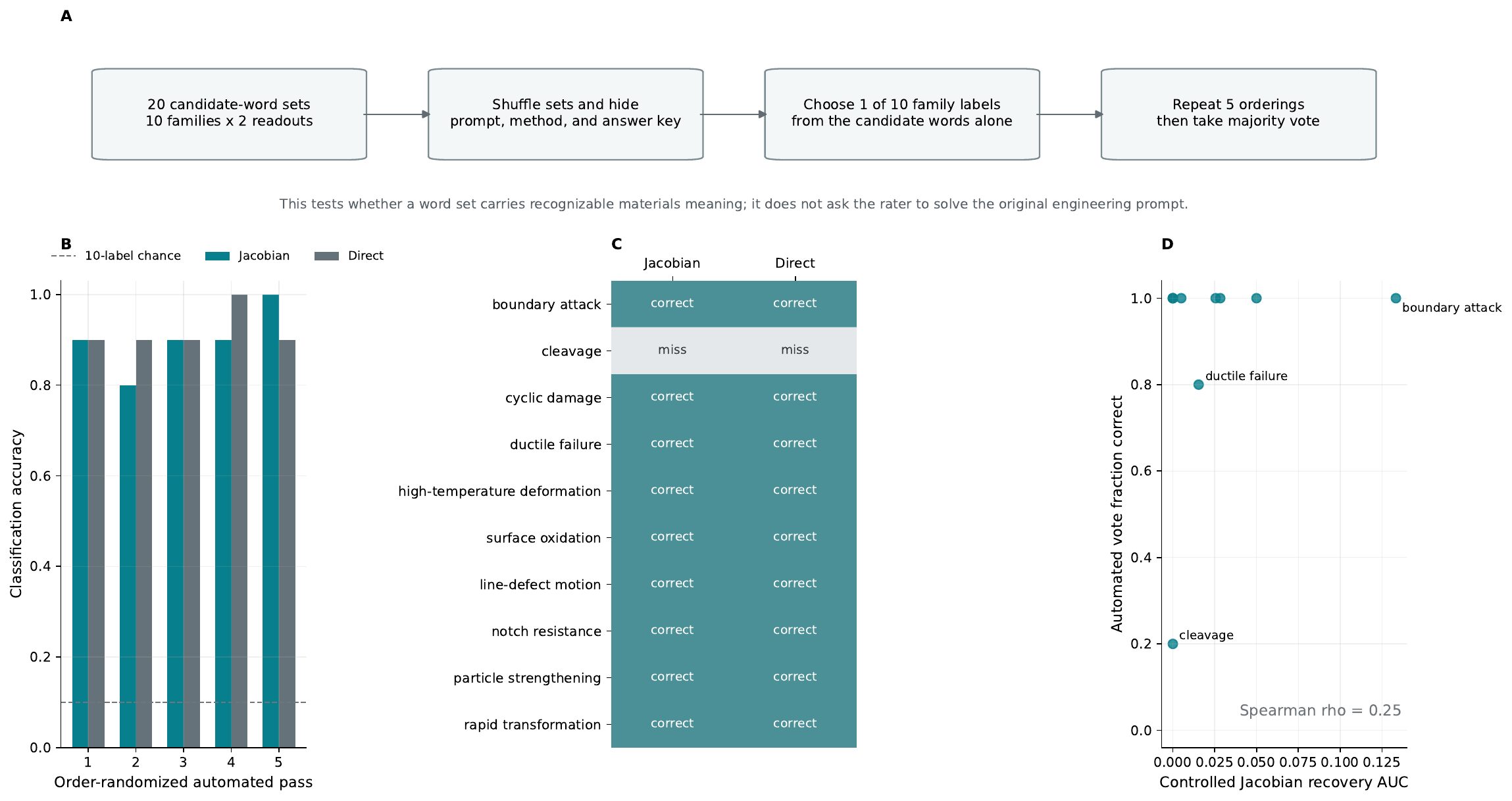}
\caption{\textbf{Can an automated interpreter recognize a materials mechanism from discovered words alone?} (A) Twenty candidate sets (ten mechanism families from each of two readouts) were shuffled and stripped of the original prompt, method identity, and answer key. The automated interpreter selected one of ten family labels for each set in five order-randomized passes; it did not solve the original engineering question. (B) Accuracy stays far above the 10\% chance level on every pass. (C) Majority-vote correctness for every family and readout: both columns are correct for nine families and both miss cleavage. Thus both naturally assembled vocabularies carry recognizable materials meaning, with no Jacobian-specific advantage in this secondary test (paired difference 0, $p=1.0$). (D) Each point is one Jacobian family. The horizontal coordinate is predeclared exact-term recovery; the vertical coordinate is the fraction of automated passes assigning the target-free words correctly. Cleavage is weak on both measures, but most families are semantically identified even when exact targets are rarely top-ranked, producing only a weak relationship ($\rho=0.254$). Controlled recovery and target-free recognition therefore measure complementary properties. The five passes are repeated judgments from one automated system, not independent scientific expert (hence this figure is limited to reproducible machine-based semantic validation).}
\label{fig:blinded}
\end{figure}

As a reproducible secondary semantic analysis, we asked a narrow question: if an automated interpreter sees only eight naturally decoded words, can it tell which materials mechanism produced them? We formed 20 word sets (one Jacobian and one direct set for each of ten families) shuffled them, hid the original descriptions, readout names, and answer key, and supplied only the ten allowable family labels (Figure~\ref{fig:blinded}A). One OpenAI model classified every set in five different orderings. This tests whether the word sets contain recognizable mechanism information; the interpreter never answered the original engineering prompts.

\hlyellow{Because all five passes came from one automated system, we treat this experiment as secondary semantic corroboration rather than independent expert validation; the fixed cross-phrasing classifier in Figure~\ref{fig:targetfree-classification} provides the quantitative target-free assessment.}

Majority accuracy was 9 of 10 families for both Jacobian and direct candidate sets (Figure~\ref{fig:blinded}). Their paired family difference was exactly zero ($p=1.0$ over all 1,024 label swaps). Individual-pass Jacobian accuracy ranged from 80\% to 100\%; direct accuracy ranged from 90\% to 100\%. Agreement over the five repeated passes was high (overall Fleiss $\kappa=0.899$). Both methods failed on cleavage, consistent with its weak open-vocabulary stream. A 100,000-shuffle label null gave $p<10^{-5}$ for each method, so both sets contain strong materials-family information, but the secondary test provides no evidence of a uniquely Jacobian advantage.

The family-level relation between controlled Jacobian recovery and automated vote fraction was weak (Spearman $\rho=0.254$, family-bootstrap interval $-0.333$ to $0.708$). Controlled term recovery and target-free semantic identification therefore appear complementary rather than interchangeable.

\hlyellow{Even without direct circularity, GPT-5.5 and Gemma may share semantic priors inherited from web-scale training, so their agreement does not necessarily substitute for blinded assessment by materials experts; future work could replicate this analysis using independently trained models and an expert panel.}

\subsection{Two cases reveal what the population summary compresses}

\begin{figure}[ht!]
\centering
\includegraphics[width=0.98\linewidth]{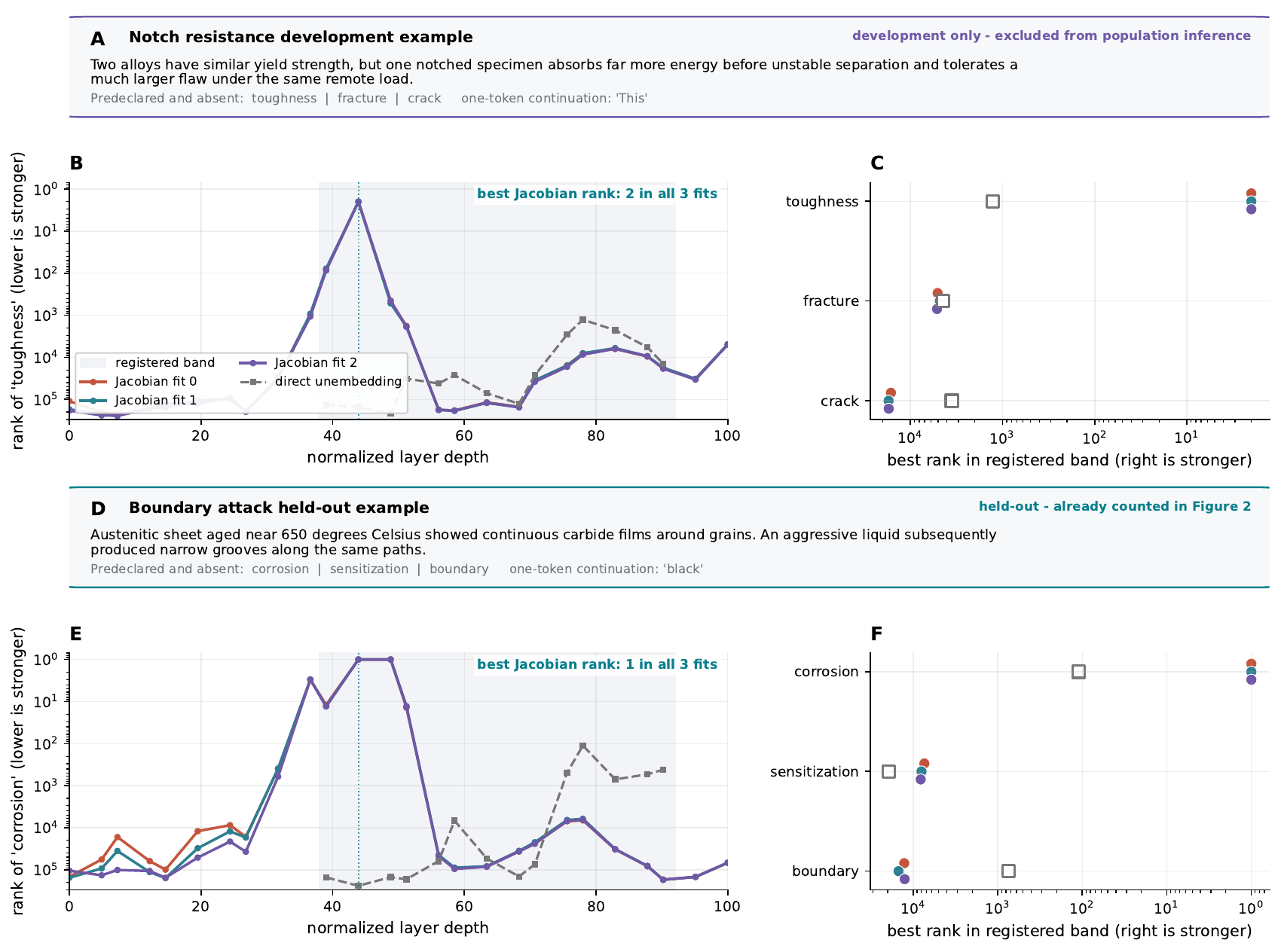}
\caption{\textbf{Two magnified examples of the controlled-rank measurement.} The panels print the exact prompts, predeclared absent words, and one-token continuations. (A-C) The notch-resistance example is an earlier development item excluded from all held-out inference. Panel B plots \code{toughness}'s full-vocabulary rank at every sampled layer for three independent Jacobian fits and direct unembedding; upward is stronger because rank 1 is best. The registered 38-92\% band is shaded. Panel C reduces each word to its best rank within that band, showing that the localized rank-2 event is specific to \code{toughness}. (D-F) The boundary-attack example is one of the 50 held-out prompts and was already counted in Figure~\ref{fig:controlled}; it is not a new test. All three lenses sustain \code{corrosion} at rank 1 over two sampled layers, while direct unembedding reaches rank 111. Panel F shows that \code{sensitization} and \code{boundary} do not receive the same advantage. Thus Figure~\ref{fig:controlled} asks how often declared concepts are recovered across the frozen population, whereas this figure shows the layer location, fit-to-fit agreement, duration, and word specificity that its aggregate statistic necessarily discards. Neither trajectory is a literal chain of thought.}
\label{fig:cases}
\end{figure}

Figures~\ref{fig:controlled} and~\ref{fig:cases} use the same controlled-rank measurement at different scales. Figure~\ref{fig:controlled} compresses every held-out prompt-concept pair to its best registered-band rank and then combines those ranks into recovery curves, prompt AUCs, and family-level inference. Figure~\ref{fig:cases} removes that aggregation. For two examples it preserves the complete rank-versus-depth trajectory, shows the three independently fitted lenses separately, retains the matched direct readout, and compares the primary word with two related predeclared alternatives. It asks where a strong event occurs, whether lens refits reproduce it, and whether all related terms rise indiscriminately. It does not estimate how frequently such events occur and adds no population-level test.

The first row is the earlier notch-resistance development prompt: ``Two alloys have similar yield strength, but one notched specimen absorbs far more energy before unstable separation and tolerates a much larger flaw under the same remote load.'' It is excluded from all 50-prompt statistics. The terms \code{toughness}, \code{fracture}, and \code{crack} were fixed before execution and absent from both the prompt and one-token continuation. At source layer 18 (43.9\% depth), all three lenses placed \code{toughness} at rank 2; direct unembedding never improved beyond rank 1,276. The leading Jacobian neighborhood was \code{robustness}, \code{toughness}, \code{capability}, \code{abilitas}, and \code{ductility}. By contrast, \code{fracture} remained at ranks 4,788-5,138 and \code{crack} at 16,249-17,214. The peak is reproducible and property-specific, but sharply localized: the rank falls away at the adjacent sampled layers and fails the registered two-layer sustained-top-5 rule.

The second row magnifies one observation already included in Figure~\ref{fig:controlled}: ``Austenitic sheet aged near 650 degrees Celsius showed continuous carbide films around grains. An aggressive liquid subsequently produced narrow grooves along the same paths.'' The predeclared absent terms were \code{corrosion}, \code{sensitization}, and \code{boundary}. All three lenses placed \code{corrosion} at rank 1 at 43.9\% and 48.8\% depth, compared with direct rank 111. The unrestricted peak neighborhood (\code{corrosion}, \code{destructive}, \code{damaged}, \code{damaging}, \code{damage}, and the fragment \code{corro}) is coherent with degradation. The readout did not simply elevate every family label: \code{sensitization} remained at ranks 7,371-8,145, and direct unembedding actually ranked \code{boundary} better than the Jacobian lenses (740 versus 12,600-14,885). This case therefore demonstrates sustained access to a broad process word, not complete reconstruction of the textbook mechanism.

The boundary-attack case was selected post hoc because it is the clearest held-out rank event; it was not added as a new observation and was already counted once in the population analysis. The development case is excluded from that analysis altogether. We place both magnifications after the population, target-free, geometry, and blinded results so that selected vivid examples explain the registered statistic without leading or strengthening the paper's inference.

\subsection{Mechanism directions causally and selectively steer materials answers}

Here steering means a controlled numerical perturbation of the model's inner hidden states; not fine-tuning the model and not adding a verbal hint to the prompt. We first ran the frozen prompt to a prespecified layer. At only the final prompt token, we then added or subtracted a small vector representing one materials mechanism and allowed the otherwise unchanged Gemma network to complete its forward pass. Finally, we compared the probabilities of two explicit engineering outcomes. The five doses, from $-4\%$ to $+4\%$ of the unperturbed state's norm, are dimensionless perturbation sizes rather than physical changes in temperature, grain size, or composition. This is analogous to a local sensitivity experiment: if a mechanism direction participates in the decision, moving the hidden state along that direction should predictably change the relative support for the two materials outcomes.

The vectors were constructed from mechanism vocabulary that did not contain the outcome words. For corrosion, the positive set was \code{sensitization}/\code{corrosion}/\code{anodic}/\code{depletion} and the negative set was \code{passivation}/\code{protected}/\code{resistant}/\code{passive}; the scored answers \code{grooves} and \code{clean} were withheld. For martensitic transformation, a quenched, shear-driven, tetragonal direction was contrasted with a diffusion-assisted, annealed equilibrium direction, while the answers were \code{hard} and \code{soft}. For grain-size strengthening, boundary blocking and strengthening were contrasted with coarse-grain slip and mobility, while the answers were \code{higher} and \code{lower}. Consequently, success cannot be obtained simply by adding the answer token itself. It requires a mechanism-defined change at an intermediate state to propagate through the remaining layers into an independently specified engineering judgment.

\subsubsection{Broad mechanism-steering screen}

We first tested whether that propagation generalized across 30 new conditions: ten each for intergranular corrosion, martensitic transformation, and grain-size strengthening. The cases represented recognizable materials situations rather than paraphrases of the direction words. Examples included a Type 304 insert held for 80 hours at $620\,^{\circ}$C with chromium-depleted grain boundaries, a laser-austenitized steel track that self-quenches fast enough to bypass diffusion, ECAP refinement of aluminum from 35 to $3.5\,\mu$m, and overannealing of alpha brass from 11 to $68\,\mu$m. The corresponding questions asked whether the qualification surface would show \code{grooves} or remain \code{clean}, whether the transformed region would be \code{hard} or \code{soft}, and whether yield strength would be \code{higher} or \code{lower}. Each prompt was run in both answer orders. The matched direction was compared at the same layer and dose with two unrelated materials directions, a direction constructed without Jacobian transport, and ten random directions.

\begin{figure}[ht!]
\centering
\includegraphics[width=0.98\linewidth]{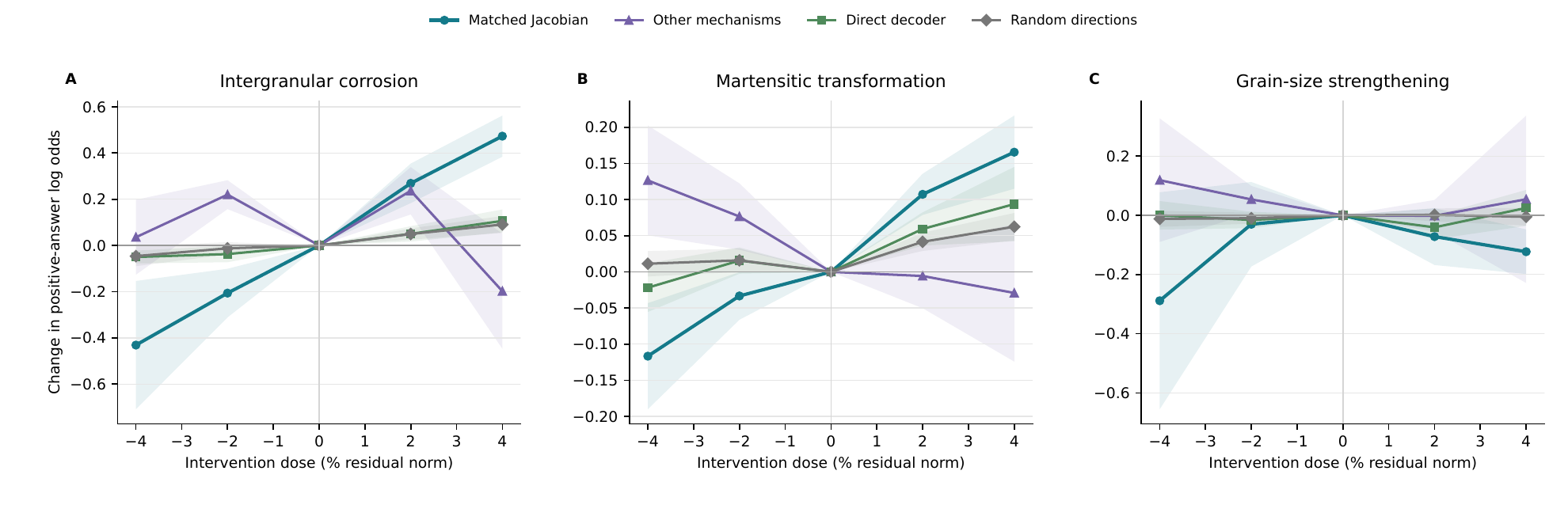}
\caption{\textbf{Broad steering is mechanism dependent.} Each panel averages ten new physical conditions and plots how an intermediate-state intervention changes the log odds of the first scientific answer relative to the second: \code{grooves} versus \code{clean} for corrosion (A), \code{hard} versus \code{soft} for transformation (B), and \code{higher} versus \code{lower} for grain size (C). Zero is the unperturbed model; the horizontal coordinate is the added direction's size relative to the final-prompt residual norm. Points are condition-weighted means and shaded bands are one standard error across physical conditions. The teal curve applies the scientifically matched Jacobian direction. Purple applies the two unrelated materials directions at the identical layer, green constructs the matched direction without Jacobian transport, and gray averages ten random directions. Corrosion is the only family that passes every preregistered gate: its matched curve is strong, monotonic, and distinct from all controls. Transformation trends in the intended direction but lacks registered specificity. The grain curve averages refinement, for which \code{higher} is correct, with coarsening, for which \code{lower} is correct; its cancellation motivated the relation-aware experiment in Figure~\ref{fig:steering}. The figure therefore shows both the causal promise and the non-universality of mechanism steering.}
\label{fig:steering-screen}
\end{figure}

Figure~\ref{fig:steering-screen} plots the change from the unperturbed answer log odds as dose is swept. A rising matched curve means that adding the mechanism direction favors the first listed outcome: \code{grooves}, \code{hard}, or \code{higher}; a falling curve favors \code{clean}, \code{soft}, or \code{lower}. The corrosion result is the clearest materials example. Increasing the sensitization/depletion direction substantially favors attack grooves, whereas random and direct controls remain near zero and the unrelated-mechanism curve does not reproduce the monotonic effect. Transformation has the expected positive trend, but its advantage over the wrong mechanisms is too uncertain to pass every gate. The unaligned grain average is weak because refinement and coarsening are physically opposite relations: a useful direction should favor \code{higher} after refinement but \code{lower} after coarsening, causing the raw family mean to cancel rather than remain uniformly positive.

Across all 30 conditions, the matched Jacobian direction changed positive-versus-negative answer log odds by $+0.451$ (95\% interval $0.160$-$0.726$) from the negative to positive dose endpoint. Its paired advantage was $+0.386$ over random directions ($0.134$-$0.623$; Wilcoxon $p=0.0040$), $+0.602$ over unrelated mechanism directions ($0.198$-$0.985$; $p=0.0066$), and $+0.352$ over the direct direction ($0.052$-$0.646$; $p=0.0384$). However, only intergranular corrosion passed every preregistered family gate. Transformation failed the specificity interval and grain size failed its original fixed-sign intervals and sign-count gate. The broad screen therefore establishes an important boundary condition: a mechanism can be readable without providing reliable, context-independent control.

\subsubsection{Prospective relational confirmation}

After the broad screen, an explicitly post hoc reorientation toward each already frozen correct answer exposed a more informative grain-size pattern: all four refinement conditions moved toward \code{higher}, whereas all four coarsening conditions moved toward \code{lower}. We treated that observation only as hypothesis generation and froze a new matched-pair study before inspecting any corresponding output. The prospective cohort contained titanium alloy, magnesium alloy, low-carbon steel, silver, cobalt alloy, and bronze. Every material contributed a refinement and a coarsening prompt with the same material identity, two grain sizes, covariates, and answer words; only the direction of change was reversed. For example, titanium changed either from 64 to 8 micrometers or from 8 to 64 micrometers while composition, texture, precipitates, porosity, and dislocation density were held fixed. The same grain direction, layer 16, five doses, controls, scoring rule, pair-level statistical unit, and pass criteria were inherited without tuning.

This matched design asks a stronger materials question than whether a direction simply favors the word \code{higher}. Under the ordinary Hall-Petch regime, refinement increases boundary area available to impede dislocation motion and should favor higher yield strength; reversing the process to coarsening should favor lower strength. A context-sensitive grain mechanism should therefore reverse its output effect even though the numerical sizes, alloy identity, other microstructural variables, answer words, layer, and intervention vector are unchanged.

\begin{figure}[ht!]
\centering
\includegraphics[width=0.98\linewidth]{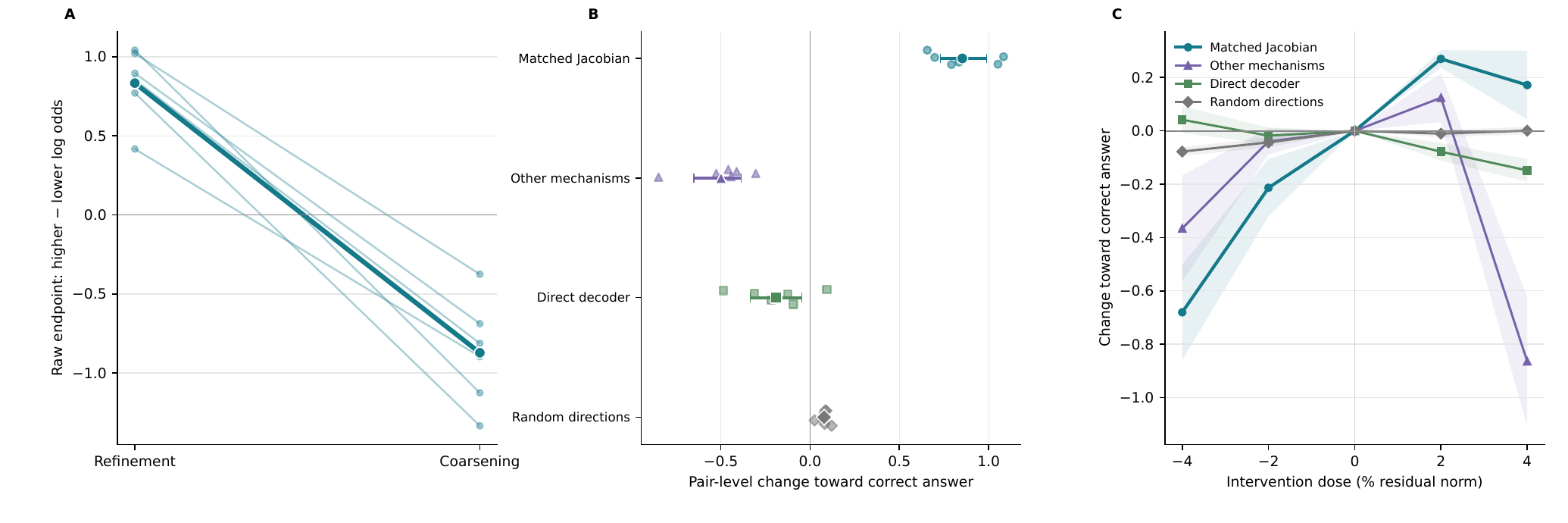}
\caption{\textbf{Can one frozen grain-size mechanism direction change a scientific answer in the relation-appropriate direction?} Six new materials were each written as a matched pair: grain refinement and grain coarsening with alloy identity, the two grain sizes, controlled covariates, and answer words held fixed. At the final prompt token in layer 16, the same frozen Jacobian direction was swept from $-4\%$ to $+4\%$ of the state's norm; the full strings \code{higher} and \code{lower} were scored at each dose. (A) For every material, the raw change in \code{higher}-minus-\code{lower} log odds is positive for refinement and negative for coarsening. Each connecting line therefore crosses zero: the direction responds to the relation rather than always favoring one word. (B) Signs are aligned so that positive always means movement toward the physically correct answer, then refinement and coarsening are averaged within material. Every circle is one independent matched pair; diamonds and bars are the mean and pair-clustered 95\% interval. The matched Jacobian effect is $+0.852$ and exceeds random, unrelated-mechanism, and direct controls in all six pairs. (C) The mean dose curves show the reversal continuously rather than only at endpoints. The experiment prospectively confirms a localized context-dependent causal pathway; it does not reveal a prose chain of thought or establish general materials reasoning. Exact prompts, both answer orders, clean outputs, all 2,400 intervention rows, and the full protocol are printed or linked in the Supplementary Information.}
\label{fig:steering}
\end{figure}

The prospective test passed every registered criterion (Figure~\ref{fig:steering}). The raw \code{higher}-minus-\code{lower} effect was positive after refinement and negative after coarsening in all six material pairs. After orienting each condition toward its physically correct answer, all 12 conditions were positive and all 6 pairs had both relations correct. The pair-level matched effect was $+0.852$ log odds (95\% interval $0.729$-$0.988$). Matched minus random was $+0.774$ ($0.630$-$0.926$), matched minus unrelated mechanisms was $+1.351$ ($1.156$-$1.533$), and matched minus direct was $+1.043$ ($0.944$-$1.136$); each paired Wilcoxon test gave $p=0.03125$, the smallest attainable two-sided value for six nonzero pairs. Both answer-word orders agreed in all 12 conditions, the registered answer remained the global top beginning in every matched row, and the median endpoint correlation across lens fits was 0.966. Because the same direction changes sign when only the scientific relation is reversed, this result cannot be explained as a fixed bias toward the token \code{higher} within this design.

A post hoc audit of the already prospectively collected five-dose curves showed that the effect was not confined to the registered $\pm4\%$ endpoints. From $-2\%$ to $+2\%$, the pair-level matched effect was $+0.484$ log odds ($0.357$-$0.658$), positive in all six material pairs and larger than every matched control in all six pairs (exact two-sided $p=0.03125$ for each pair-level contrast). All 72 stored condition-answer-order-lens-fit trajectories had a positive full-curve slope, but only 10 were strictly monotone at every adjacent dose. The robustness result therefore supports a consistent local slope and half-dose effect, not a generally linear dose response.

\subsubsection{Answer-vocabulary and regime transfer fail}

The prospective result is strong within its frozen \code{higher}/\code{lower} decision format, but a scientific control operator should transfer beyond those words and beyond the conventional Hall-Petch regime. We therefore froze two additional six-pair cohorts using the unchanged grain direction and the complete answer strings \code{increase}/\code{decrease}. The first used new conventional polycrystals, where refinement should increase strength and coarsening should decrease it. Only 7 of 12 conditions moved in the correct direction and only 1 of 6 pairs was correct for both relations; the registered gates failed. The second explicitly stated that each nanocrystalline material lay below its inverse Hall-Petch crossover, where refinement should decrease strength. Only 6 of 12 conditions were correct and no pair was correct in both directions.

Across both transfer cohorts, 23 of 24 conditions moved toward the manifest's designated positive answer word, regardless of whether that word was physically appropriate. This failure changes the causal claim. The earlier refinement/coarsening reversal is not an unrestricted grain-size operator; it is a localized context-and-vocabulary-dependent pathway that works convincingly for one decision format and does not automatically transfer to new output words or a reversed physical regime. The complete dose curves and prompts are reported in the Supplementary Information.

\subsubsection{Exploratory counterfactual activation patching}

The steering experiment adds a deliberately constructed mechanism direction. Hence we next asked a complementary question that does not use a lens to modify the model: is the naturally occurring hidden state itself sufficient to carry the refinement-versus-coarsening decision? Consider the titanium pair. The receiver says that grains change from 64 to 8 micrometers and correctly favors \code{higher}; its matched donor says that the same alloy changes from 8 to 64 micrometers and correctly favors \code{lower}. At each registered layer, we ran both prompts and replaced only the receiver's final-prompt-token residual vector with the donor's entire vector from the same layer. We then completed the receiver forward pass and measured how far its \code{higher}-minus-\code{lower} log odds moved toward the donor's answer. The same operation was repeated in both directions, both answer orders, and all six materials.

The prediction is simple; if the transplanted state carries only surface wording or material identity, replacing it should not consistently move a refinement answer toward \code{lower} or a coarsening answer toward \code{higher}. If the state carries the physical relation in a form used by the remaining layers, the reversed-relation transplant should redirect the receiver toward the donor's conclusion. A reverse state from another alloy tests whether that relation transfers across materials. Equal-distance states that preserve the receiver's relation test whether any sufficiently large replacement would produce the same effect. Thus the experiment tests causal sufficiency: whether introducing a naturally occurring counterfactual state is enough to change the downstream scientific decision. It does not test whether that state is the only representation the model can use or whether every coordinate in it is required.

This is a stronger test than reading a word from a state because it changes the state and measures the downstream consequence. It is also more specific than simply replacing a vector with a distant vector. We compared the reversed-relation donor with three alternatives: another material expressing the same reversed relation, another material preserving the receiver relation, and the identical condition with only answer order changed. Because reverse donors initially lay farther from the receiver than some controls, a separately frozen falsification analysis rescaled the relation-preserving and order-only directions to exactly match the reverse donor's Euclidean distance.

\begin{figure}[ht!]
\centering
\includegraphics[width=0.99\linewidth]{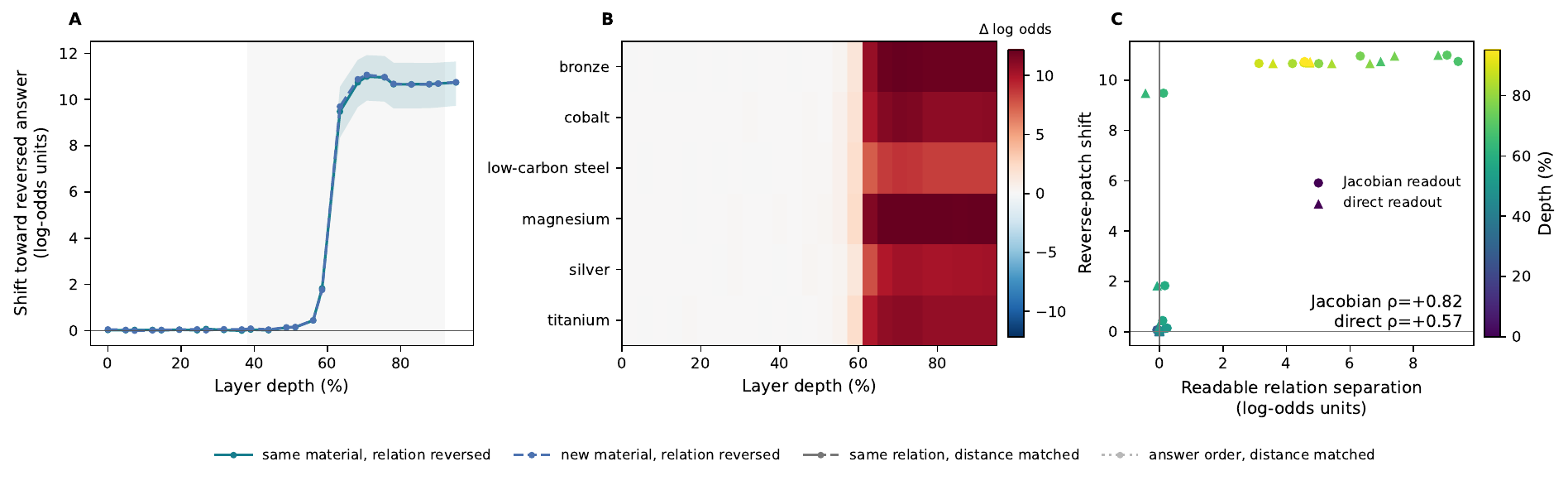}
\caption{\textbf{Can a hidden state from the opposite grain-size relation causally transfer its answer?} A receiver prompt and donor prompt use the same answer words but describe opposite refinement/coarsening relations. At one layer, the receiver's final-prompt-token state is replaced by the donor state; positive values mean movement toward the answer appropriate to the donor's reversed relation. If the state contains only wording or alloy identity, the reversed transplant should not consistently redirect the answer. (A) Read from left to right, same-material and cross-material reverse patches have little effect early but produce a sharp late-layer shift of roughly 11 log-odds units at the peak. Relation-preserving and answer-order-only directions are rescaled to the same receiver-donor distance but remain much smaller. The shaded vertical region is the fixed 38-92\% analysis band; the band-mean same-material reverse effect is $+6.249$ ($5.663$-$6.795$). (B) The same-material reverse effect appears in all six matched alloy pairs, excluding a result driven by one material. (C) Layers where refinement and coarsening are more separable by a vocabulary readout tend also to show larger patching effects. The Jacobian level correlation is $\rho=0.818$ and the direct correlation is $\rho=0.571$, but onset and first-difference checks support only a broad late-stage correspondence, not exact layer-by-layer localization. The experiment establishes causal sufficiency in this constrained decision: a naturally occurring reversed-relation state can redirect the answer and can transfer between alloys. It does not identify a unique one-dimensional mechanism, prove necessity, or reveal a prose chain of thought. It is called exploratory because it reused a grain-size cohort whose strong effect was already known, not because the intervention is noncausal.}
\label{fig:patching}
\end{figure}

The reverse-relation patch had essentially no effect in early and middle layers, then changed sharply near 60\% depth (Figure~\ref{fig:patching}A). Averaged over the fixed 38-92\% band and six material pairs, the same-material reverse patch shifted the receiver toward the donor answer by $+6.249$ log-odds units (95\% pair-bootstrap interval $5.663$-$6.795$). All 12 physical conditions had the expected sign descriptively, and all six independent material-pair means were positive (exact two-sided pair sign test $p=0.03125$). A reversed-relation state from a different material transferred equally well: $+6.283$ ($5.925$-$6.558$). By contrast, the equal-distance same-relation control was $+0.082$ ($-0.874$-$1.085$), and the equal-distance answer-order control was $+1.055$ ($0.779$-$1.262$). Matched reverse minus those controls was $+6.166$ ($5.425$-$7.015$) and $+5.193$ ($4.804$-$5.618$), respectively. The effect therefore follows reversal of the physical relation, transfers across alloys, and is not explained by answer order or perturbation magnitude within this cohort.

Panels B and C localize, but do not over-localize, that conclusion. Every material shows the late-layer effect. Across the 25 layers, the aggregate causal curve correlates with the readable refinement-coarsening separation under the Jacobian lens ($\rho=0.818$) more strongly than with direct unembedding ($\rho=0.571$). However, post hoc first-difference correlations are weak and the patch effect begins before either readout's registered onset. The defensible interpretation is broad stage agreement: the relation becomes both causally potent and vocabulary-readable late in the network, not that the Jacobian lens identifies the exact causal layer.

\subsubsection{Cross-mechanism patching separates physical outcome from numerical direction}

The grain patch proves causal sufficiency inside a known positive cohort, but refinement, \code{higher}, and decreasing grain size are locked together. We therefore froze a harder factorial study before generating any patching output. Twenty-four natural questions from six governing mechanisms served as receivers. At layers 16, 24, 32, and 37, the receiver's final-question-token state was replaced by a state from every eligible other-family donor. The receiver question itself remained unchanged. Donors crossed \code{higher}/\code{lower} with \code{greater}/\code{smaller}, and nine mechanism pairs reversed whether an increasing input produces an increasing property. This separates three hypotheses: copying a particular answer token, transferring the raw numerical trend, and transferring the physical property outcome.

Across all 15 unordered mechanism pairs, the donor-aligned physical-outcome contrast was $+0.385$ (pair-sign exact $p=0.00897$; structured donor-label exact $p=0.03194$; Supplementary Figure~S25). Transfer remained positive when donor and receiver used different answer vocabularies ($+0.491$) and when their numerical-to-property orientations were reversed ($+0.382$). Thus the patch does more than copy one vocabulary token. However, the numerical-direction contrast was stronger overall ($+0.571$), and only four of six donor mechanism families had positive physical-outcome transfer, failing the frozen five-of-six breadth gate. Orowan and porosity donors were especially revealing: they transferred the numerical increase/decrease feature with the wrong physical sign.

The causal conclusion is therefore precise. Full-state patching transfers a late answer-relevant numerical/decision feature that is aggregate-aligned with physical outcome across vocabularies, but heterogeneous across governing mechanisms. It is not a universal materials-relation state. Because the patch replaces all 2,560 coordinates and no Jacobian map constructs the intervention, this experiment establishes sufficiency of a distributed late state, not necessity, a one-dimensional constitutive-law feature, or a special causal power of the lens.

\subsubsection{Disjoint relational-coordinate intervention across scale}

\hlyellow{The preceding patches establish sufficiency by introducing a complete counterfactual state. To test whether a narrower representation contributes during ordinary inference, we defined one physical-outcome coordinate per Gemma scale using only an independent 16-law development cohort, then intervened on 24 target laws absent from that fit. Each equation-free relation was tested in both numerical directions, with \code{higher}/\code{lower} and \code{greater}/\code{smaller}, and in both answer orders, giving 192 prompts per checkpoint. At the same prespecified late normalized depth used for the relational scale comparison, we either set only the coordinate to its development midpoint, replaced it with the matched reversed-input value, replaced the complete final-token state, or restored the original coordinate inside that counterfactual state. Six equal-length controls per operation used activation-derived axes with the physical component removed.}

\begin{figure}[ht!]
\centering
\includegraphics[width=0.99\linewidth]{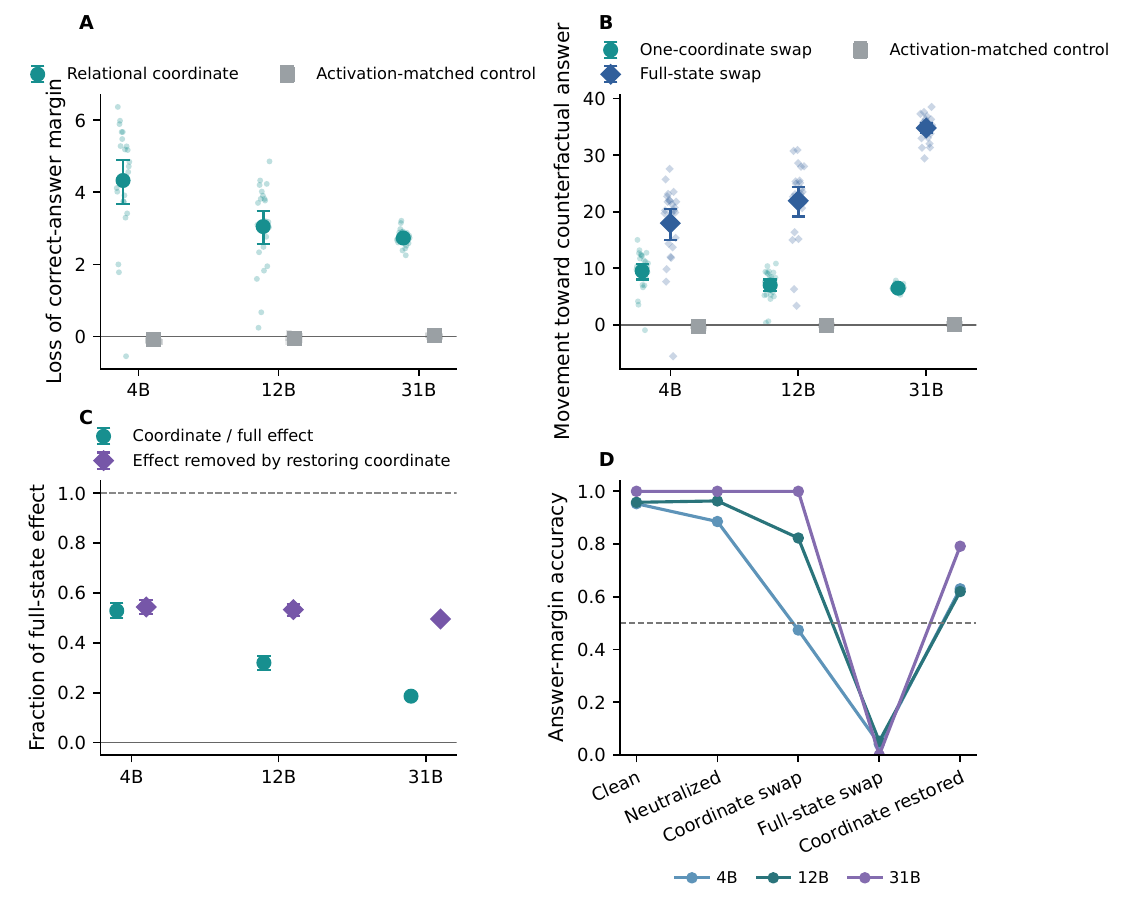}
\caption{\textbf{A development-fitted relational coordinate makes a selective, context-dependent causal contribution to equation-free physical decisions.} (A) Setting only the coordinate to its development midpoint reduces the correct-answer margin at all three scales; equal-length activation-matched controls remain near zero. Pale points are 24 law identities; large markers and bars are law means and 95\% law-bootstrap intervals. (B) Replacing only the coordinate with its value from the matched reversed-input prompt moves the decision toward the counterfactual answer. Full-state replacement shows the total transferable effect; matched controls remain near zero. (C) The coordinate-only share of full-state transfer decreases with scale, but restoring the original coordinate within the counterfactual state removes approximately one-half of the total effect at every scale. (D) Neutralization changes discrete accuracy chiefly at 4B. Full-state counterfactuals reverse almost all original decisions, while coordinate restoration recovers many of them. The experiment establishes selective causal involvement, not a unique or universally necessary representation.}
\label{fig:causal-knockout}
\end{figure}

\hlyellow{Neutralizing the coordinate reduced the correct-answer margin by 4.33, 3.05, and 2.73 logits at 4B, 12B, and 31B, whereas the corresponding control means were $-0.09$, $-0.05$, and $+0.02$ (Figure~\ref{fig:causal-knockout}A). One-coordinate counterfactual replacement moved the margin toward the reversed answer by 9.50, 7.02, and 6.47 logits, with the expected physical-minus-control sign in 23/24, 24/24, and 24/24 laws (panel B; law-level permutation $p=10^{-5}$ at each scale). The coordinate alone accounted for a decreasing fraction of complete-state transfer as scale increased, but restoring it inside the full counterfactual state removed 54.4\%, 53.3\%, and 49.6\% of that effect (panel C). Neutralization changed 15 initially correct 4B decisions to errors and repaired two existing errors (McNemar $p=0.00235$), but did not reduce discrete accuracy at 12B or 31B (panel D). Thus the coordinate selectively contributes to normal decisions across scale and is behaviorally necessary for some 4B cases; it is not the unique representation required for every answer, and the larger checkpoints retain sufficient additional evidence to preserve their choices.}

\section{Discussion}

The five questions posed in the Introduction receive five bounded answers. First, the Jacobian lens is a highly reproducible but selective measurement instrument. Independent WikiText fitting samples yield nearly identical term ranks, and several individual prompts show large, physically coherent gains. Reproducibility does not imply universal superiority: \hlyellow{in the primary 4B suite,} 36 of 50 controlled prompts are zero under both readouts, the family-level interval includes no mean advantage, and some mechanisms favor direct unembedding. Earlier work reported broader Jacobian advantages on general-language distributions \citep{gurnee2026workspace}; the present open-model materials study reproduces the method, not that numerical pattern.

Predetermined and discovered vocabulary remain complementary. Controlled recovery can audit a specific physical hypothesis such as \code{corrosion} or \code{toughness}, while target-free decoding exposes whatever words assemble naturally. Both readouts produce word sets that an automated secondary rater recognizes for 9 of 10 mechanism families. Yet prompt TF-IDF remains competitive for family classification, zero of 523 target-free words survives the network enrichment correction, and the frozen complete-sequence test succeeds for only 7 of 10 multi-token prompts. Readability is therefore meaningful but sparse, and no word list by itself establishes causal use.

Second, much of the useful geometry is Gemma-general rather than Jacobian-specific. Mechanism families are separable in full states, and every case in the disjoint lexical-adversarial cohort shifts from wording similarity toward physical equivalence between middle and late layers. Direct unembedding follows almost the same trajectory, and prompt embeddings can classify families even better. A separated UMAP is therefore a descriptive map, not proof that the model discovered a new ontology.

Third, the graph experiments sharpen this boundary and expose a benchmark-design lesson. At the natural end of a complete scientific question, different material cases governed by the same supplied mechanism form above-null comparative neighborhoods without answer choices. However, the registered same/different physical target is algebraically identical to same/different numerical direction within every family: if $y=sx$, then $y_i y_j=x_i x_j$. The original graph result is therefore real but not independently diagnostic of physics. Its correct interpretation is that Gemma preserves comparative structure, which cases move in the same stated direction, inside a supplied governing law. Applying that law requires the additional orientation $s$, which determines whether an input increase raises or lowers the material property.

The audits converge on this interpretation. Graded label-blind comparison structure transfers almost perfectly across mechanisms, yet the transferred maps do not preserve physical case identity above the structured null. Even when numerical direction is supplied explicitly, a graph network fails to assign absolute physical polarity for an unseen mechanism; the perfect numerical-direction readback is only a positive control confirming that the supplied feature remains available. Spectral agreement improves when graded affinities are retained, while exhaustive partitions show that the registered split is not the unique global division. Direct states match the Jacobian topology and raw states are often stronger. The six displayed islands are supplied strata, not discovered laws, and the cross-mechanism counter-numeric endpoint fails. The useful general insight is not a universal materials axis but a symmetry-aware criterion for interpretability: a scientific graph cannot identify a label that can be globally flipped without changing its unlabeled structure, and a relation endpoint should not be called independent physics when it collapses exactly to a prompt variable.

\hlyellow{The neutral-anchored experiment changes the measured object and therefore does not contradict that negative absolute-graph result. The earlier graph asks where individual states lie and how they neighbor one another; its polarity is subject to a global sign ambiguity. The matched benchmark instead compares the two endpoints of a controlled numerical reversal while holding the law identity, material, prompt surface, and answer order fixed. Subtraction removes much of the fixed context, and separate neutral laws provide an empirical zero. The supplied-equation condition first establishes that inverse, neutral, and direct laws produce an ordered state transformation when the relation is stated explicitly. The equation-free condition then shows that nearly the same ordering remains when the equation and prescribed sign decomposition are removed. This is evidence for a task-elicited, transferable computation of qualitative constitutive orientation rather than merely a readback of printed algebra. It is deeper than word readability because it concerns how a state transforms under a physical counterfactual, but narrower than a complete physics model because the law identities are reused and the experiment does not recover coefficients, dimensions, conservation, unfamiliar laws, or a spontaneous multistep rationale. The gap between latent orientation and exact output is model- and condition-dependent. In the supplied-equation 4B benchmark, the law-aggregated hidden-state score oriented 39 of 40 directional laws although prompt-level inverse accuracy was only 53.8\%. In the equation-free tests, directional-law accuracy was 38/40, 40/40, and 40/40 at 4B, 12B, and 31B, whereas exact overall answer accuracy was 75.8\%, 64.8\%, and 99.3\%, respectively. Thus, representational success does not guarantee reliable output conversion, although the 31B result shows that both can approach ceiling together.}

\hlyellow{The matched 12B and 31B replications show that this equation-free relational result is not peculiar to the 4B checkpoint. Both larger checkpoints preserve the complete inverse-neutral-direct ordering and orient all 40 directional laws on the same frozen test prompts. The paired uncertainty intervals nevertheless touch zero because the 4B measurement was already near ceiling, so the data establish replication across scale more directly than a population scaling law. The measurement is refitted in each model's own coordinates and does not imply a shared vector across different hidden widths. Even within 31B, independently fitted directions at 67.8\% and 83.1\% depth have cosine 0.352 while their 60 law scores correlate at Pearson $r=0.996$ and Spearman $\rho=0.949$ (Supplementary Figure~S28). The relational variable can therefore remain stable while its coordinate axis changes with model depth.}

\hlyellow{The neutral-anchored matched-state benchmark is the most directly portable procedure because it uses within-model normalization, a model-specific development direction, and an empirical neutral calibration rather than assuming shared hidden coordinates. The disjoint relational-coordinate intervention could likewise be repeated after refitting its direction within each model. By contrast, the exact grain-steering vector, selected layer and perturbation dose, as well as target-free vocabulary ranks, are likely to be more architecture sensitive because tokenizers, output heads, residual coordinates, layer schedules, and training corpora differ. The present 4B-12B-31B results therefore establish within-Gemma scale robustness; evaluating Llama, Mistral, or OLMo would be required to establish cross-architecture transfer.}

Three otherwise distinct measurements show broad late-stage convergence. Physical-equivalence geometry rises sharply after approximately 60\% depth, grain-size patching reaches 10\% and 50\% of its peak at 58.5\% and 63.4\% (Supplementary Figure~S24), and the disjoint 12-law relational sweep becomes continuously strong across the latter half of the network (Supplementary Figure~S19). Because the tasks, states, and endpoints differ, and the relational sweep contains isolated early peaks, this agreement supports late-stage stabilization rather than one shared feature or an exact emergence layer.

\hlyellow{The absolute-versus-relational distinction suggests a broader and more scientifically natural graph framework. The 60-law matched-comparison benchmark demonstrates one way to break the exact alias $y=sx$ by crossing numerical direction, law orientation, and physical outcome and by introducing a separately calibrated neutral class. The equation-free transfer removes the explicit two-stage scaffold but retains known response and control quantities, the same law identities, and constrained answer words. A future graph should go further by holding out entire constitutive-law families, inferring the governing mechanism without a supplied class, testing option-free or free-form responses, and extending from signed monotonicity to graded equation-predicted response.} It should then be hierarchical and multiplex rather than flat. One layer would connect prompts by an inferred governing mechanism; a second would encode signed and graded response; a third would connect compatible quantities, units, boundary conditions, and uncertainty; and the same nodes would be followed through network depth. Hyperedges could represent process-structure-mechanism-property chains that require more than pairwise similarity. Full weighted affinities should be primary because the present sparsification audit shows that a top-edge graph discards much of the graded structure. The current graph atlases demonstrate that such visualization is technically possible, but a universal materials graph still requires a prospectively standardized, option-free cohort with entire governing laws and model families held out.

Fourth, intervention converts selected claims from correlation to causation, but not to universality. The broad screen passes every family gate only for corrosion. The prospective grain study is stronger: one frozen layer-16 direction reverses its output effect between refinement and coarsening, remains effective at half the registered dose, and exceeds random, unrelated-mechanism, and direct-direction controls in all six material pairs. However, the same direction fails new \code{increase}/\code{decrease} cohorts and the inverse Hall-Petch boundary. Its impressive causal effect is real within the tested \code{higher}/\code{lower} context, but it is vocabulary and regime dependent.

\hlyellow{The disjoint 24-law intervention adds the complementary ablation test requested by this distinction. Neutralizing one development-fitted relational coordinate selectively lowers the evidence for the unmodified model's correct answer at every scale, whereas matched activation-derived controls do not. Reversing that coordinate moves evidence toward the counterfactual answer, and restoring it removes approximately half of a full-state counterfactual effect. The 4B accuracy loss demonstrates necessity for some individual decisions; the preserved 12B and 31B choices show that causal contribution need not imply unique necessity when a larger model has stronger margins or redundant evidence.}

Fifth, transfer reveals the boundary of the causal result. Full-state patching moves a distributed late decision feature: in the known positive grain cohort, reversed-relation states redirect answers across alloys and survive equal-distance controls. In the frozen six-mechanism factorial, transfer remains positive across different answer vocabularies and opposite numerical-to-property orientations, ruling out simple copying of one answer token. Numerical direction nevertheless transfers more strongly than physical outcome, Orowan and porosity donors have the wrong physical sign, and the family-breadth gate fails. The patch is therefore causal but not a universal materials-relation operator. It replaces all 2,560 coordinates, proves sufficiency rather than necessity, and is not constructed by the Jacobian lens.

None of these results reveals a literal internal monologue. The ribbons are not thought transcripts, UMAP axes are not reasoning coordinates, a graph edge is not a reasoning step, similarity is not causation, and most decoded words have not been causally tested. Materials-science information is reproducibly readable in Gemma, but readable structure is not automatically constitutive physics: much of the absolute geometry is compatible with lexical and numerical comparison and does not uniquely identify a mechanism-independent physical representation. \hlyellow{The neutral-anchored result provides a more specific positive claim: controlled state transformations carry a transferable monotonic law orientation first under an explicit equation scaffold and then after that equation and prescribed decomposition are removed.} Selected internal directions can also causally steer constrained scientific decisions in context-dependent ways. This story is useful precisely because it is bounded: it distinguishes absolute location, relational transformation, and causal use, and states where each measurement stops.

\subsection{Outlook for representation-aware scientific model training}

The present study uses the Jacobian lens after training, but the same measurements could become candidate feedback signals during fine-tuning. Ordinary supervised fine-tuning and reinforcement learning from human feedback predominantly reward the answer that leaves the model \citep{christiano2017preferences,ouyang2022instructgpt}. That is necessary but incomplete for science: two models can give the same correct engineering answer while one uses the intended mechanism and the other exploits a wording shortcut. A representation-aware objective would preserve the final-answer reward while adding independent tests of the scientific structure that supports it. In schematic form,
\begin{equation}
R = R_{\mathrm{answer}}
  + \lambda_m R_{\mathrm{mechanism}}
  + \lambda_c R_{\mathrm{counterfactual}}
  + \lambda_s R_{\mathrm{specificity}}
  + \lambda_u R_{\mathrm{uncertainty}}
  - \lambda_{\ell} R_{\mathrm{lexical\ shortcut}}.
\end{equation}
The weights would be chosen on development families, then frozen before evaluation. This expression could guide future work.

Specifically, the mechanism term would reward a physically appropriate internal concept family being readable across several paraphrases and independently fitted lenses, not the appearance of one preferred token. A corrosion tutor, for example, could reward a stable neighborhood involving sensitization, chromium depletion, passivity, and boundary attack when the thermal and chemical history warrants it. A fracture tutor could reward the appropriate competition among cleavage, void nucleation, coalescence, plasticity, and notch constraint. Target-free discovery could contribute a softer reward when the model assembles a stable, expert-recognizable semantic neighborhood without reproducing the instructor's exact vocabulary. Agreement across paraphrases, lens fits, and materials-expert labels would matter more than any single decoded word.

Graph structure offers a related reward that does not depend on one vocabulary token, but the identifiability audit shows why a single flat target graph would be too crude. The reward benchmark must first ensure that its physical target is not algebraically identical to numerical wording. A hierarchical reward could then ask whether cases assemble by governing mechanism, whether the correct constitutive-law orientation is applied, and whether signed and graded responses match the governing equation. Cross-mechanism links could test units, invariants, and equation-predicted equivalences; a multiplex version could compare the same nodes across depth, paraphrase, and counterfactual relation. Hyperedges could reward complete process-structure-mechanism-property chains, such as heat treatment $\rightarrow$ precipitate spacing $\rightarrow$ dislocation bypass $\rightarrow$ yield strength, rather than isolated pairwise similarity. Training would score full-candidate ranking before sparse visualizations, persistence across layers and held-out lenses, and failure to follow lexical, answer-order, or raw increase/decrease shortcuts. Whole mechanism families would remain held out for evaluation so that the model cannot obtain reward merely by reproducing the supplied taxonomy.

The counterfactual term would ask the mechanism to change its downstream effect when the physics changes. Matched pairs could reverse refinement and coarsening, oxidizing and reducing environments, heating and cooling paths, tension and compression, loading and unloading, or defect creation and annealing while holding irrelevant words fixed. A grain-size mechanism would earn reward only if it supports the correct relation in both members of the pair; simply favoring \code{higher} would fail. In phase-transformation training, the same idea could require the predicted transformation pathway to change consistently with cooling rate and composition. In mechanics, a constitutive model could be rewarded for reversing signed quantities under a reversed load while preserving invariant quantities such as elastic modulus or fracture energy where appropriate.

The specificity term would turn interventions into an anti-shortcut test. A small frozen mechanism-direction perturbation should change the associated scientific decision in the registered direction, whereas unrelated materials directions and matched random vectors should not. One could reward the difference between the intended intervention and these controls rather than rewarding raw sensitivity, because a model that reacts strongly to every direction is not scientifically selective. Conservation and dimensional constraints supply further science-specific rewards: mass and charge should balance, phase fractions should remain normalized, units should be consistent, and a process-structure-mechanism-property chain should agree across atomistic, microstructural, and continuum descriptions. Such constraints could support curricula in which simple one-mechanism cases precede coupled degradation, fracture, and multiphysics design problems.

Uncertainty should also be rewarded rather than trained away. Near a phase boundary, an inverse Hall-Petch crossover, or an ambiguous failure surface, the desired internal state may preserve two competing mechanisms and produce calibrated probabilities instead of an artificially sharp single word. A useful uncertainty reward could therefore favor agreement between internal competition, output calibration, and known experimental uncertainty. It could penalize confident mechanism claims when small, physically irrelevant prompt changes reverse the readout. This would make representation-aware training relevant not only to question answering but also to experiment selection: a model could propose the measurement that most cleanly distinguishes two internally competing hypotheses.

However direct optimization creates a serious risk of readout hacking: the model may learn to display the rewarded vocabulary without using the associated physics. The answer-scaffold audit gives a concrete version of this danger: a rewarded scientific word can become highly readable after that same word has been supplied among the choices, even when the pre-choice relation contrast is weak. Reward signals should therefore be measured before answer options when possible and must be evaluated on option-free, free-response, and arbitrary-format transfers. The reward instrument must also be separated from the evaluation instrument. A credible experiment would freeze several reward lenses before optimization, reserve different lens fits and mechanism families for final evaluation, and include held-out paraphrases, lexical counterfactuals, reversed relations, unrelated interventions, free-form answers, and expert review. Periodically changing the reward lens or using an evaluator trained on a different corpus may further reduce probe-specific gaming. The decisive comparison would include outcome-only fine-tuning, outcome plus readout reward, outcome plus counterfactual-intervention reward, the combined objective, and a no-update baseline. Success would mean improved held-out scientific behavior and causal consistency and not merely a higher score under the probe used for training. Process supervision and representation-engineering studies make this direction plausible \citep{lightman2023verify,zou2023representation,turner2023activation}, but the safeguards above are likely essential before an internal readout can be treated as a scientific reward.

\subsection{Limitations and next experiments}

\hlyellow{The main-text vocabulary, graph, and original steering studies remain centered on one 4B checkpoint. A secondary controlled-vocabulary audit repeats the same 50 prompts at 12B and 31B, but 31B has only one lens fit and the Jacobian-minus-direct contrast is inconsistent across scale. The equation-free relational endpoint and disjoint relational-coordinate intervention were repeated at 12B and 31B, but use model-specific directions in each hidden space. The former reuses the same 16 development laws, 60 test-law identities, response/control quantities, answer set, and prompt manifest; the latter uses 24 target laws disjoint from the development fit. Together they establish within-architecture scale robustness for known qualitative relations, not a common cross-model vector or a universal scaling law.} WikiText is a public pretraining-like fitting proxy rather than a materials corpus or the proprietary distribution used in the source study. \hlyellow{Because the Jacobian maps at all three scales were fitted on general-domain WikiText-103 rather than a materials corpus, insufficient coverage of activation directions associated with uncommon technical vocabulary may contribute to sparse recovery; the present design cannot separate this measurement limitation from limited accessibility in the underlying model.} The target-free filter still admits fragments and generic words, and the automated rater is one model repeated five times rather than a panel of materials experts. Controlled recovery is zero-inflated. The frozen complete-sequence analysis of \emph{martensite} and \emph{transgranular} fails its breadth gate, so first-token ranks cannot be generalized to technical multi-token words. The lexical-adversarial replication fails its absolute late-window primary endpoint even though its frozen late-minus-middle secondary endpoint is uniformly positive.

\hlyellow{The option-free graph improves the positional evidence but reuses an inspected 72-prompt cohort, supplies mechanism families, and contains an exact label alias: within each family, same physical outcome equals same numerical direction. Its counter-numeric cross-mechanism endpoint fails; direct and raw states are similar or stronger; the artificial-checkpoint result shows sensitivity to prompt boundary; and graph-network, isomorphism, spectral, and exhaustive-partition audits do not identify a universal physical polarity. The neutral-anchored benchmark corrects that alias for a narrow monotonic task. Its original condition used an explicit two-stage instruction, whereas the follow-up removed the equation and that decomposition but reused the same 60 law identities, response/control quantities, and constrained answer set. The scale extension fits a separate development direction in each checkpoint but reuses those same laws and prompts. It therefore demonstrates prompt-format and within-architecture scale transfer, not discovery or independent generalization to unfamiliar laws. The neutral class remains physically designed rather than naturally occurring discourse; capillary rise reverses only in the 4B scaffold-removal comparison, nucleation barrier is one of the two 4B equation-free sign errors, and exact answer accuracy varies from 64.8\% to 99.3\% across checkpoints despite ceiling latent ordering at 12B and 31B. That 60-law score is observational and task-elicited. The disjoint 24-law intervention establishes a selective causal contribution of one linear coordinate, including decision-level necessity in some 4B cases, but uses constrained binary answers, one late layer per scale, and known monotonic relations; it does not establish unique global necessity, quantitative equation recovery, or a universal physics manifold. The 50-prompt mechanism graph has competitive prompt-only baselines, and zero of 523 target-free words survives enrichment correction. The prospective intervention covers six matched grain-size pairs in one answer vocabulary; both later transfer cohorts fail. Grain patching reuses a known positive cohort, and the factorial patch replaces an entire state while failing family breadth. These studies do not establish causality for every displayed word, graph edge, family, layer, token position, or model output.}

\hlyellow{The next representation study should move beyond the present scaffold-removal result by holding out entire law families and removing the constrained answer vocabulary while retaining matched counterfactuals and neutral-calibration safeguards.} Hall-Petch, Griffith, Orowan, Taylor, Arrhenius, mixture-rule, and rubber-elasticity cases should share one option-free prompt boundary, include graded equation-predicted magnitudes decorrelated from raw numerical difference and lexical similarity, and hold out entire law families. Primary endpoints should infer the mechanism without supplying its class, recover constitutive orientation, predict signed and graded response, and test whether the relational direction survives free-form and numerical answers. A multilayer graph can then track the same matched edges across depth, while hyperedges test process-structure-mechanism-property chains. Independent materials-expert assessment of decoded vocabularies would be useful future validation, but it is outside the scope and evidential claims of the present paper. \hlyellow{Future causal tests should extend the present knockout-and-restoration design across layers, prompt positions, unfamiliar law families, and free-form outputs, with a larger prospectively error-enriched cohort.} Repeating the design across Gemma scales, model families, and independently fitted raw, Jacobian, tuned, and sparse-feature readouts would separate model, scale, and measurement effects.

\section{Conclusion}

\hlyellow{Materials-science vocabulary is reproducibly readable in selected Gemma states, although controlled recovery is sparse and does not resolve a family-level Jacobian advantage over direct unembedding.} Much of its absolute geometry is compatible with, or confounded by, lexical and numerical comparison. A narrower physical abstraction becomes visible when the measurement follows a controlled transformation between states, and selected directions can exert causal, context-dependent control over constrained scientific decisions. Five findings support this conclusion. \hlyellow{First, three independent Jacobian fits nearly agree and selected prompts show large gains, but 36 of 50 controlled prompts in the primary 4B suite are null and the family-level Jacobian-direct difference is not significant; separately, target-free word sets support above-chance cross-phrasing family classification for both readouts, without a Jacobian advantage.} Second, the average vocabulary and geometry results are not uniquely Jacobian; direct unembedding, raw states, and prompt words often carry similar information. Third, option-free hidden-state graphs preserve comparative structure within supplied mechanisms (natural-question AUC 0.642), but their absolute physical polarity is not identifiable. When a separate 60-law test instead measures the state change under a matched numerical reversal and calibrates zero with independent neutral laws, it orders inverse, neutral, and direct constitutive relations ($\rho=0.910$), separates direct from inverse laws with AUC 1.000, and classifies 39 of 40 directional laws. \hlyellow{After the equation and prescribed decomposition are removed, the same frozen direction retains direct-inverse AUC 0.990, ordinal $\rho=0.877$, and 38-of-40 directional accuracy. This cross-format result supports a recalled qualitative relationship rather than a readback of printed algebra, while remaining a test of known laws in a constrained answer format.} This is a task-elicited abstraction of monotonic orientation, not a complete law or chain of thought. Fourth, causal control is possible \hlyellow{but context dependent}: one prospectively frozen grain direction reverses correctly for all 12 refinement and coarsening conditions, remains effective at half the registered dose, and exceeds matched controls. \hlyellow{A separate 24-law intervention shows that a development-fitted relational coordinate selectively contributes to ordinary equation-free decisions across 4B, 12B, and 31B: neutralization lowers correct-answer evidence, counterfactual replacement reverses it, and restoration removes about half of full-state transfer. Decision-level necessity appears in some 4B cases but not at the larger scales.} Fifth, that control remains bounded: new grain answer-word and inverse-regime cohorts fail, while cross-mechanism full-state patches transfer a late numerical/decision feature more consistently than a universal physical outcome.

\hlyellow{The equation-free relational result also replicates across Gemma scale: the 12B and 31B checkpoints reach perfect pairwise ordering and orient all 40 directional laws on the frozen cohort, strengthening the conclusion that controlled state transformations reveal a reproducible qualitative constitutive variable. Exact answer generation remains separable from that representation, falling to 64.8\% at 12B while reaching 99.3\% at 31B.}

\hlyellow{Together, these results show that direct and Jacobian readouts recover selected materials-science terms from Gemma's hidden states, matched physical reversals reveal direct-versus-inverse constitutive orientation in the resulting state changes, and neutralizing or reversing the corresponding relational coordinate causally changes answer evidence. The relational signal remains highly stable across model scale even when exact output behavior and the magnitude of a single-coordinate intervention differ.}  

\hlyellow{Follow-up work should extend the present equation-free transfer to fresh governing laws, option-free boundaries, free-form answers, and equation-predicted graded responses while preserving neutral anchors and matched reversals.} Its mechanism, constitutive-orientation, counterfactual, graph-persistence, intervention-specificity, and uncertainty scores could become separate reward signals for fine-tuning, provided evaluation uses different readouts, unseen mechanisms, free-form answers, causal error repair, and expert review. Under those safeguards, representation-aware training could move beyond rewarding only a final answer toward models whose scientific decisions are supported by more stable and testable internal structure. Finally, additional comparisons with other LLMs, including LLMs fine-tuned on materials science concepts, may provide additional insights into how various models process prompts.

\section{Materials and Methods}

\subsection{Model, revisions, and registered layer grid}

We used \code{google/gemma-4-E4B-it} \citep{gemma4modelcard} at immutable revision \code{a4c2d58be94dda072b918d9db64ee85c8ed34e3f}. The model has 42 transformer layers and hidden width 2,560. All held-out evaluations used bfloat16 model inference, a fixed final-prompt-token score position, one greedy continuation token for output-leakage checks, 25 registered source/report layers, and the fixed 38-92\% source-depth band for controlled and target-free endpoints. The vocabulary contains 262,144 tokens.

\hlyellow{The equation-free relational scale extension additionally used the matched 12B checkpoint (48 layers; width 3,840) and 31B checkpoint (60 layers; width 5,376) from the Gemma-4 instruction-tuned family; their exact repository identifiers and revisions are recorded in Supplementary Section~S1. All three checkpoints use Gemma's 262,144-token vocabulary. Their zero-indexed probe layers were 34, 39, and 49, corresponding to 82.9\%, 83.0\%, and 83.1\% normalized depth. These scale runs use unit-normalized raw residual states and model-specific development-law directions; they do not use the Jacobian-lens checkpoints. The disjoint causal-coordinate intervention uses the same three models, layers, and model-specific development directions. The original vocabulary, graph, steering, and patching experiments remain evaluations of the 4B checkpoint; a secondary Supplementary Information audit repeats only the frozen controlled-vocabulary endpoint at 12B and 31B.}

\subsection{Jacobian-lens estimator and direct baseline}

For source layer $\ell$, prompt $x$, valid source position $t$, and target-layer position $t'$, the fitted map estimates
\begin{equation}
J_{\ell}=\mathbb{E}_{x,t}\left[\sum_{t'\geq t}
\frac{\partial h_{40,t'}}{\partial h_{\ell,t}}\right],
\end{equation}
where layer 40 is Gemma's penultimate transformer layer. Causality makes only later or equal target positions contribute. The implementation excludes the first 16 positions, excludes the final position without a next-token target, computes rows in output-dimension batches by reverse-mode automatic differentiation, averages valid source positions within each record, and then averages records. The evaluation readout is
\begin{equation}
s^{J}_{\ell,t}=W_U\,\mathrm{final\_norm}\!\left(J_{\ell}h_{\ell,t}\right),
\end{equation}
where $W_U$ is Gemma's own vocabulary decoder. Direct unembedding uses the matched score
\begin{equation}
s^{D}_{\ell,t}=W_U\,\mathrm{final\_norm}\!\left(h_{\ell,t}\right).
\end{equation}
No separately trained classifier converts states to words. Both methods use the same decoder and differ only in the layer-to-final transport.

Each of three lenses was fitted on 1,000 unique 128-token records drawn from the \code{wikitext-103-raw-v1} training split with corpus seeds 0, 1, and 2. All use the same model revision, penultimate target, 25 source layers, and released Jacobian-lens estimator. The checkpoints for the \hlyellow{lenses of the 4B, 12B and 31B models} are stored in the Hugging Face repository \code{lamm-mit/gemma4-jacobian-lenses} at recorded revision \code{37ba15033a42e72bdfbc04815b3fbd37e516fd59}.

\hlyellow{The 12B and 31B Jacobian lenses used the same paper-scale fitting recipe as used for the 4B case: 1,000 unique 128-token WikiText-103 records per fit, exclusion of the first 16 positions, the model-specific penultimate target layer, and 25 evenly spaced source layers. The 12B checkpoint used corpus seeds 0, 1, and 2 with target layer 46; the 31B checkpoint used seed 0 with target layer 58. The released files are \code{gemma4-12b-it/paper/seed0.pt}, \code{seed1.pt}, and \code{seed2.pt}, and \code{gemma4-31b-it/paper/seed0.pt} in \code{lamm-mit/gemma4-jacobian-lenses}.}

\subsection{Mechanism-direction construction}

Steering used semantic contrasts rather than the answer words being tested. For mechanism family $f$, four prespecified positive concept tokens $P_f$ and four negative concept tokens $N_f$ define
\begin{equation}
q_f(z)=\log\sum_{i\in P_f}\exp u_i(z)-
       \log\sum_{j\in N_f}\exp u_j(z),
\end{equation}
where $u_i(z)$ is token $i$'s logit after Gemma's final normalization and fixed decoder. For each of three frozen direction-fitting prompts, the state at source layer $\ell$ was transported to $z=J_{s,\ell}h_\ell$ for lens fit $s$. We differentiated $q_f$ with respect to $z$, pulled the gradient back through the fitted map, and normalized:
\begin{equation}
d^{J}_{f,s,\ell,r}=\frac{J_{s,\ell}^{\mathsf T}\nabla_z q_f(z)}
{\|J_{s,\ell}^{\mathsf T}\nabla_z q_f(z)\|_2}.
\end{equation}
The three prompt-specific unit vectors were averaged and renormalized to give one direction per family, lens fit, and layer. The matched direct control repeats the procedure through \code{final\_norm} without $J$. Answer outcomes such as \code{higher}, \code{lower}, \code{grooves}, \code{clean}, \code{hard}, and \code{soft} never enter either direction construction.

Candidate layers 10, 16, 20, 24, and 28 were evaluated once in a separate preliminary study using lens seed 0 and disjoint calibration conditions. The frozen selection score was the mean signed $-4\%$ to $+4\%$ endpoint across answer orders minus one population standard deviation, with the lower layer breaking ties. This selected layers 16, 24, and 16 for corrosion, transformation, and grain size. The broad screen reused those layers without selection. The prospective grain experiment then reused the same grain layer and every direction-fitting prompt without further tuning.

\subsection{Localized intervention and scientific-answer scoring}

For an evaluation prompt, intervention was restricted to the final prompt token at the frozen source layer:
\begin{equation}
h'_{\ell,t}=h_{\ell,t}+\frac{a}{100}\|h_{\ell,t}\|_2d,
\qquad a\in\{-4,-2,0,2,4\}.
\end{equation}
All other positions and layers were left unchanged. We scored each complete lowercase answer string with teacher forcing,
\begin{equation}
L(y\mid x,d,a)=\sum_{m=1}^{|y|}\log p(y_m\mid x,y_{<m};d,a),
\end{equation}
using its exact no-leading-space continuation tokenization. For multi-token answers, the same intervention remains at the original final prompt position while subsequent answer pieces attend through the ordinary network. The measured contrast is positive-answer minus negative-answer log probability. The endpoint subtracts this contrast at $-4\%$ from its value at $+4\%$.

Every condition was asked twice, with the two answer words presented in opposite orders. Every target used three matched Jacobian directions, one per lens fit; the other two mechanism directions reconstructed at the identical target layer for all three fits; the matched direct direction; and ten seeded random unit vectors. A random vector was made orthogonal to the seed-0 matched Jacobian direction and to the remaining independent component of the direct direction. Symmetric doses, identical layers, and identical residual-norm scaling prevent a control from receiving a smaller nominal intervention. Clean correctness, probability mass on the answer pair, next-token validity, endpoint choice flips, and clean-to-intervened Kullback-Leibler divergence were audited but never used to exclude a frozen item.

\subsection{Steering cohorts, endpoints, and inference}

The broad screen contained ten conditions per family: four clear positive cases, four clear negative cases, and two deliberately near-threshold cases. Two answer orders produced 60 exact prompts. With three matched lens directions, six wrong-mechanism directions, one direct direction, ten random directions, and five doses, the approach used 6,000 intervention rows. The independent unit was the physical condition after averaging answer orders and, where applicable, lens fits. Family-level 95\% intervals used 30,000 condition bootstrap resamples. A family passed only if at least 8 of 10 condition endpoints were positive, at least 8 of 10 conditions had the same nonzero sign under both answer orders, the matched-minus-random and matched-minus-wrong-mechanism intervals were above zero, and a registered answer beginning remained the global top token in at least 90\% of matched rows. Integrated Wilcoxon tests and the exact binomial sign test were secondary summaries across 30 conditions. Orienting determinate endpoints toward the known correct answer was not preregistered in the broad screen and is reported only as the hypothesis-generating analysis that motivated the prospective confirmation.

The prospective confirmation used six new matched material pairs. Each pair contributed one refinement and one coarsening condition, each in both answer orders, for 24 exact prompts and 2,400 intervention rows. Its primary endpoint equals the raw \code{higher}-minus-\code{lower} endpoint for refinement and its negative for coarsening, so positive always means movement toward the frozen correct answer. The independent unit was the material pair after averaging lens fits and answer orders within condition and then averaging its two relations. Pair-clustered 95\% intervals used 30,000 resamples. The registered pass required at least 80\% positive individual conditions, at least 80\% of pairs correct in both relations, at least 80\% order-sign agreement, all three matched-minus-control bootstrap intervals above zero, and at least 90\% valid global-top answer beginnings. Pairwise Wilcoxon tests were reported descriptively; with six nonzero pairs, $p=0.03125$ is the smallest attainable two-sided value. All exact prompts, frozen concepts, clean answers, amendments, and success rules are reproduced in the Supplementary Information.

A post hoc dose-shape audit reused all five prospectively collected doses without another model forward pass. For each condition-answer-order-lens-fit trajectory, it calculated the least-squares slope against perturbation dose and the aligned difference from $-2\%$ to $+2\%$; strict monotonicity was descriptive. Inference retained the six material pairs as independent units, with 30,000 fixed-seed pair bootstraps and exact $2^6$ sign flips. This is a robustness analysis of prospectively collected data, not a second prospective confirmation.

Two later transfer protocols were each frozen before their model output and reused the unchanged grain direction, layer, doses, scoring, and two answer orders. The conventional cohort contained six new polycrystals with refinement and coarsening but replaced \code{higher}/\code{lower} by the complete strings \code{increase}/\code{decrease}. The inverse Hall-Petch cohort contained six nanocrystalline materials explicitly stated to lie below their material-specific crossover, so the expected relation was reversed. Each protocol required at least 80\% correct condition signs and at least 80\% of material pairs correct in both relations, with the same control and output-validity gates. The raw records, exact prompts, condition rows, pair rows, and failed gates are retained under \path{experiments/candidate-followups-2026-07-16/}.

\subsection{Exploratory full-state activation patching}

Activation patching reused the 24 exact prompts from the prospective grain-size study: six materials, refinement and coarsening, and two answer orders. Its protocol was frozen before any patching output, but after the original steering output had been inspected. At each of the same 25 registered layers, we replaced the final-prompt-token residual after that transformer block with a donor residual from the identical layer. The primary donor was the same material and answer order with the grain-size relation reversed. The transfer donor used the next material in a frozen cyclic order with the relation reversed. Relation-preserving and order-only donors used the next material with the same relation or the identical condition with only answer order changed. No Jacobian map was used to construct or apply a patch.

The endpoint was exact next-token \code{higher}-minus-\code{lower} log odds. Let $s=+1$ for a refinement receiver and $s=-1$ for a coarsening receiver. We defined the counterfactual-aligned shift as $-s$ times patched minus clean log odds, so positive means movement toward the answer appropriate to the opposite relation. The independent unit was the matched material pair after averaging both relations and answer orders. The primary scalar was the mean over the fixed 38-92\% band; 30,000 pair-cluster bootstrap resamples gave 95\% intervals. Layerwise peaks were descriptive.

The reverse donor states were farther from the receiver than some original controls. After seeing that diagnostic, we froze a separate falsification protocol before generating its output. For each receiver and layer, it rescaled the same-relation and order-only direction to have exactly the Euclidean norm of the matched reverse displacement, then repeated the patch and inference unchanged. This post hoc control tests perturbation magnitude but cannot convert an inspected cohort into an independent confirmation.

To compare readability with causal sensitivity, we separately decoded each clean receiver state using the three Jacobian lenses and direct unembedding. Within a material, relation separation was one half of the refinement contrast minus the coarsening contrast, averaged over answer orders and, for the Jacobian readout, lens fits. We compared the 25-layer relation-separation curve with the matched patching curve by Spearman correlation and a 49-member circular-shift null. These level correlations are descriptive. Post hoc onset and first-difference checks were used to prevent a shared late-layer rise from being described as exact causal localization. The complete layerwise patch and readout curves, including these diagnostics, are reported in Supplementary Figure~S24.

\subsection{Frozen cross-mechanism activation patching}

The factorial patching protocol was frozen before any corresponding patch output. It used the natural-question anchor stems for four material cases in each of six mechanisms, for 24 receivers. Each receiver was paired with all four anchor donors in each of the five other mechanisms, giving 480 ordered receiver-donor pairs. At layers 16, 24, 32, and 37, only the receiver's final-question-token residual was replaced by the donor residual, producing 1,920 patches. Neither receiver nor donor contained an answer scaffold. No Jacobian map constructed or applied the patch.

For every receiver, output was scored as its own positive-minus-negative next-token logit using either \code{higher}/\code{lower} or \code{greater}/\code{smaller}. For each ordered donor-family to receiver-family pair and layer, the mean patched receiver margin under the donor family's two negative-outcome cases was subtracted from the mean under its two positive-outcome cases. We then averaged over receivers, the four frozen layers, and both directions of each unordered mechanism pair. Prespecified subsets were all 15 pairs, nine cross-vocabulary pairs, nine opposite-response-orientation pairs, and five pairs satisfying both controls. The secondary numerical-direction contrast instead grouped donors by increasing versus decreasing input.

Two-sided exact inference enumerated all $2^n$ sign flips of the unordered pair effects; 30,000 fixed-seed pair bootstrap resamples supplied 95\% intervals. During execution, before endpoints were calculated, an amendment froze a complementary structured donor-label test. It enumerates all $6^6=46{,}656$ balanced donor-family label assignments while retaining every pair and requires both the pair-sign and structured tests for strong evidence. The strong gate additionally required at least five of six donor-family physical-outcome contrasts above zero. Every receiver, donor, layer, state distance, answer margin, and negative family result is retained in \path{experiments/cross-mechanism-activation-patching-2026-07-18/}.

\subsection{Disjoint relational-coordinate intervention}

\hlyellow{For each Gemma scale, the causal coordinate was fitted only on the existing 16-law development cohort. Final-prompt-token residuals at layers 34, 39, and 49 were normalized to unit length; the unit vector joining the mean negative- and positive-outcome states defined the model-specific axis, and the two development means defined its midpoint and outcome centroids. The target cohort comprised 12 direct and 12 inverse equation-free relations absent from that fit. Each law crossed two numerical directions, two one-token answer vocabularies, and both presentation orders, producing 192 prompts per checkpoint. No Jacobian lens was used.}

\hlyellow{At the target layer we changed only the final-prompt-token residual. Neutralization set its normalized coordinate to the development midpoint while retaining its orthogonal component; coordinate replacement set it to the value from the exact reversed-input prompt; full-state replacement transplanted that prompt's complete residual; and restoration returned the original coordinate inside the transplanted state. Three null axes were built from real paired development transformations after balanced sign shuffling, removal of the physical component, and mutual orthogonalization. Both signs of every null axis were applied at exactly the physical intervention's Euclidean length and averaged within prompt. This activation-matched control was frozen before its output after an earlier Gaussian control was rejected as off-manifold; the superseded rows are retained and excluded. Correct-answer margin was the registered-answer logit minus its alternative. Intervals used 30,000 law-bootstrap resamples and specificity used 100,000 law-level sign permutations. Error repair included every strictly negative clean margin; no prompt was selected by its intervened result. Complete construction, controls, prompts, strata, and validation are reported in Supplementary Section S7A.}

\subsection{Practical command-line workflow for a new experiment}

The repository exposes fitting, prompt execution, raw-data storage, and per-prompt visualization through \code{run\_lens.py}. The examples below are operational commands, not pseudocode. First, a user can fit and save a paper-recipe lens without evaluating any prompt. The new \code{-fit-only} flag exits after writing the weight file and its metadata sidecar. The optional Hugging Face flags upload both files to a repository in the same operation.

\begin{lstlisting}[style=cli]
python run_lens.py \
  -model google/gemma-4-E4B-it \
  -model-revision a4c2d58be94dda072b918d9db64ee85c8ed34e3f \
  -lens lenses/gemma4-e4b-it.my-seed.pt \
  -fit -fit-only -recipe paper \
  -corpus wikitext -corpus-subset wikitext-103-raw-v1 \
  -corpus-seed 0 -dtype bfloat16 -dim-batch 16 \
  -hf-lens-repo OWNER/PRIVATE-REPO \
  -hf-lens-path gemma4-e4b-it/paper/my-seed.pt \
  -hf-upload-lens
\end{lstlisting}

A new controlled prompt is a small JSON record. For example, \path{prompts/my-fatigue-case.json} can contain:

\begin{lstlisting}[style=cli]
{
  "prompts": [{
    "slug": "my-fatigue-case",
    "shape": "ASSOCIATION",
    "domain": "materials",
    "title": "Cyclic shaft damage",
    "text": "A shaft survived millions of alternating load cycles before separation. Concentric arrest marks spread from a surface notch and fine parallel ridges covered the smoother region.",
    "readout_selector": "final_prompt_token",
    "tracked": ["fatigue", "crack", "propagation"],
    "must_be_absent_from_input": true,
    "must_be_absent_from_output": true
  }]
}
\end{lstlisting}

\Needspace{16\baselineskip}
One exploratory prompt is run with the \code{demo} evaluation wrapper because the quantitative \code{paper} wrapper correctly requires at least 50 independent items per evaluated shape. The lens itself remains the 1,000-record paper-recipe checkpoint:

\begin{lstlisting}[style=cli]
python run_lens.py \
  -model google/gemma-4-E4B-it \
  -model-revision a4c2d58be94dda072b918d9db64ee85c8ed34e3f \
  -lens lenses/gemma4-e4b-it.my-seed.pt \
  -recipe demo -prompts prompts/my-fatigue-case.json \
  -shapes ASSOCIATION -domains materials \
  -workspace-band 38,92 -generation-max-new-tokens 1 \
  -layer-readout-top 32 -surprising-top 64 \
  -open-vocab-logit-baseline -tag my-fatigue-case
\end{lstlisting}

This command writes the complete machine-readable record to \path{runs/my-fatigue-case.json}. For each prompt it writes PNG and SVG versions of the token-by-layer grid and fixed-position semantic stream. When tracked terms are present, it additionally writes full-vocabulary rank trajectories, sustained-emergence summaries, and a prompt-position heatmap for the strongest tracked term under \path{figures/my-fatigue-case/materials/}. To run target-free discovery, the user removes the \code{tracked} field and adds \code{-discover 12}; the selected words, their layer trajectories, and the unrestricted candidate lists are then recorded rather than supplied by the user. A single prompt is appropriate for exploration and figure development. Population claims require a frozen multi-prompt manifest, independent phrasings, all three lens fits, and the family-aware analysis used here.

The complete steering studies are executable from their JSON manifests. On a Linux GPU, the following commands regenerate every raw row, statistical table, Markdown report, and publication figure; \code{-device mps} can replace \code{cuda} on Apple silicon. A neutral entry point selects either the broad screen or the prospective matched-pair confirmation and resolves its versioned manifest and output paths.

\begin{lstlisting}[style=cli]
python scripts/run_mechanism_steering.py \
  -study broad-screen -device cuda -dtype bfloat16
python scripts/analyze_mechanism_steering.py \
  -study broad-screen

python scripts/run_mechanism_steering.py \
  -study prospective-grain -device cuda -dtype bfloat16
python scripts/analyze_mechanism_steering.py \
  -study prospective-grain
\end{lstlisting}

The lexical-adversarial discovery, disjoint replication, and post hoc scaffold audit are reproduced as follows. The second command reuses the common runner but supplies the disjoint manifest and output paths. The large compressed state arrays are regenerated locally; the committed JSON, CSV, and figure products permit numerical inspection without them.

\begin{lstlisting}[style=cli]
python scripts/run_lexical_adversarial_representation.py \
  -device cuda -dtype bfloat16 -chunk-size 12
python scripts/analyze_lexical_adversarial_representation.py

python scripts/run_lexical_adversarial_representation.py \
  -protocol experiments/late-physics-representation-replication-2026-07-17/protocol.json \
  -output experiments/late-physics-representation-replication-2026-07-17/raw.json \
  -states-output experiments/late-physics-representation-replication-2026-07-17/representations.npz \
  -device cuda -dtype bfloat16 -chunk-size 12
python scripts/analyze_late_physics_replication.py

python scripts/run_answer_code_binding.py \
  -device cuda -dtype bfloat16 -chunk-size 12
python scripts/analyze_answer_code_binding.py
python scripts/analyze_answer_scaffold_audit.py
\end{lstlisting}

Once the two representation arrays exist, the graph analysis and its independent integrity audit run without another model forward pass. The third command regenerates the SI's complete 72-prompt and 144-primary-edge LaTeX inventories directly from the frozen machine-readable files.

\begin{lstlisting}[style=cli]
python scripts/analyze_graph_topology_rigorous.py
python scripts/audit_graph_topology_rigorous.py
python scripts/build_graph_si_inventory.py

python scripts/run_option_free_question_end_states.py \
  -device cuda -dtype bfloat16 -chunk-size 12
python scripts/analyze_option_free_question_end.py
python scripts/analyze_cross_mechanism_outcome.py
python scripts/plot_relation_graph_robustness.py
\end{lstlisting}

The graph analysis writes the selected edge for every method, every eligible candidate rank, layerwise tables, hard-negative comparisons, target-free word tests, exact-null results, validation checks, and publication figures under \path{experiments/graph-topology-rigorous-2026-07-17/}. The portable manuscript bundle mirrors the complete prompt manifest, protocols, amendments, all-method edge and rank CSVs, and primary edge inventory under \path{paper/graph_data/}. Thus the figure can be audited without reconstructing a plotted value by eye.

The graph-identifiability audit reuses the option-free arrays and therefore requires no additional model inference. The following commands reproduce the label-blind matching, graph-network, symmetry, spectral, exact-partition, figure, validation, and SI products. The Supplementary Information lists the same commands beside the corresponding protocols, seed counts, and output files.

\begin{lstlisting}[style=cli]
python scripts/analyze_graph_isomorphism_generalization.py
python scripts/run_graph_isomorphism_gin.py
python scripts/analyze_graph_gauge_generalization.py
python scripts/run_graph_relation_gin.py
python scripts/analyze_spectral_graph_communities.py
python scripts/analyze_exact_graph_partitions.py
python scripts/plot_graph_generalization_audit.py
python scripts/validate_graph_generalization_audit.py
python scripts/build_graph_generalization_si.py
\end{lstlisting}

\Needspace{14\baselineskip}
The frozen multi-token and cross-mechanism patching follow-ups are reproduced with:

\begin{lstlisting}[style=cli]
python scripts/run_multitoken_sequence_robustness.py \
  -device cuda -dtype bfloat16
python scripts/plot_multitoken_sequence_robustness.py

python scripts/run_cross_mechanism_activation_patching.py \
  -device cuda -dtype bfloat16
python scripts/analyze_cross_mechanism_activation_patching.py
python scripts/audit_cross_mechanism_activation_patching.py
\end{lstlisting}

For analysis-only regeneration from raw data, run \path{scripts/rebuild_review_robustness_outputs.sh}. It rebuilds the positional graphs, cross-mechanism falsification, multi-token figure, patching statistics, actual 25-layer graph atlas, SI inventory, and validation record without another Gemma forward pass.

For a new mechanism, a user copies the broad-screen manifest, defines two four-word concept ensembles, three direction-fitting prompts, two scientific answer strings, a frozen layer, and new confirmation conditions. The exact words being scored as outcomes should not appear in the direction ensembles. Development prompts may be used to choose a layer, but the layer and all controls must then be frozen before the confirmation prompts are run. The raw JSON records every exact rendered prompt, tokenization, direction identity, dose, answer probability, and integrity diagnostic, so a new analysis need not rely on a figure alone.

\subsection{Held-out evaluation design and leakage controls}

The generator and exact 50-prompt manifest were finalized on 14 July 2026 before held-out execution. The manifest contains five prompts in each family listed in Table~\ref{tab:families}. No prompt duplicates a development item. Mean word-5-gram Jaccard overlap with the earlier association suite was 0.0061 and the maximum was 0.0952.

Every tokenizer-resolved declared term had to be absent from the input and the generated one-token continuation. Items failing this rule would be excluded identically for all methods and seeds; none failed. Multi-token terms remained in the prompt inventory but were excluded from the one-token rank endpoint exactly as registered. The complete exact prompts, tracked terms, continuation tokens, and prompt-level results are provided in the companion Supplementary Information and in \code{experiments/MATERIALS\_HELDOUT\_V1\_COMPLETE\_SI.md}.

\subsection{Lexical-adversarial physical-equivalence cohorts}

The first lexical-adversarial protocol was frozen before any corresponding model forward pass. It comprised six mechanism families with four material systems per family: grain size/yield strength, crack size/fracture stress, temperature/diffusivity, fiber angle/axial stiffness, fiber fraction/elastic modulus, and martensite fraction/hardness. Each material system generated an anchor, a physically equivalent paraphrase with changed wording and units, and a near-verbatim counterfactual with the numerical change reversed, for 24 triplets and 72 prompts. The anchor and paraphrase had the same registered one-token scientific answer; the counterfactual had the opposite answer. Both word-level and character 3-5-gram TF-IDF were calculated before model execution. In all 24 triplets, each baseline ranked the counterfactual as more similar to the anchor than the physical paraphrase. No triplet was removed for model error or an unattractive representation.

At the final prompt position we captured the same 25 source layers used elsewhere. For direct decoding, each state was mapped through Gemma's final normalization. For Jacobian decoding, it was first multiplied by the corresponding frozen layer map and then final-normalized. The three transported arrays were retained separately; the ensemble array is their arithmetic mean. At every layer and separately for every representation, states were centered across all 72 prompts and row-normalized. For triplet $i$, method $q$, and layer $\ell$, the registered margin was
\begin{equation}
m_{i,q,\ell}=
\cos\!\left(a_{i,q,\ell},p_{i,q,\ell}\right)
-\cos\!\left(a_{i,q,\ell},c_{i,q,\ell}\right),
\end{equation}
where $a$, $p$, and $c$ denote anchor, physical paraphrase, and lexical counterfactual. Positive values mean physics outranks wording. The prospective primary scalar averaged $m$ over the fixed 38-92\% band for the three-fit Jacobian ensemble. Population intervals used 30,000 two-stage bootstrap resamples: six families were sampled with replacement and four triplets were then sampled within each selected family. The frozen breadth gates required at least 18/24 positive triplets and 5/6 positive family means. Direct, individual fits, raw states, uncentered cosine, clean answer consistency, and target-free top-word Jaccard were secondary controls.

After that primary analysis failed, its retained layer curve showed an unregistered late rise. A second protocol was therefore frozen before running any prompt from a disjoint cohort. It used six new mechanisms and 24 new material systems: obstacle spacing/Orowan stress, porosity/modulus, pearlite spacing/yield strength, dislocation density/yield strength, stiff-particle fraction/modulus, and crosslink density/rubbery modulus. The exact triplet construction and lexical preflight were unchanged. The prospectively selected primary window was 80-96\% depth. The 38-92\% full-band result and the late-minus-middle contrast, defined as mean 80-96\% margin minus mean 38-70\% margin, were frozen secondary endpoints. Inference and breadth gates were unchanged. The completed discovery statistics were stored before the disjoint protocol was executed; the discovery cohort was not rerun or substituted.

\subsection{Graph construction, falsification, and statistical tests}

The graph analyses reuse two full-state datasets but define new endpoints. Their primary protocol was specified and stored before any graph statistic was calculated; the underlying arrays had already been generated and inspected in earlier analyses, so all graph results are labeled post hoc. Two subsequent amendments froze answer-order, exact structured-null, ranking, and robustness tests before those specific calculations. No prompt, node, edge, layer, family, or word was removed after graph output was known.

\paragraph{Cross-phrasing mechanism graph.}
The first graph contains the 50 held-out descriptions, ten mechanism families, five phrasing folds, 25 layers, and three fitted lenses. At each layer, a source prompt selects its most similar target in each of the other four phrasing folds, giving 200 directed edges. The primary graph uses cosine similarity averaged across the fixed 38-92\% band and asks what fraction of edges remain inside the source's mechanism family; balanced chance is 10\%. The same edge rule was applied to the Jacobian ensemble, each individual lens, raw states, target-layer states, mean input-token embeddings, word TF-IDF, and character 3-5-gram TF-IDF. Label permutations were blocked by phrasing fold, and every layerwise scan retained the maximum across all 25 layers inside each of 50,000 permutations.

To test whether prompt wording alone explained the graph, we regressed each pairwise Jacobian similarity on word TF-IDF cosine, character TF-IDF cosine, token-set Jaccard overlap, absolute token-count difference, and a same-phrasing indicator. The residual-similarity graph used the same frozen edge rule and blocked null. A separately frozen hard-negative test selected, in every eligible target fold, the different-family prompt with the highest mean word and character TF-IDF similarity and compared its state cosine with the true-family target. Repeated graph assembly recorded how often every edge appeared across registered layers and three lens fits; pairwise fit Jaccard and ROC-AUC for same-family identification quantified stability. Finally, 523 target-free consensus words were tested for family enrichment with blocked permutations and Benjamini-Hochberg false-discovery control at 0.05. This last analysis returned zero significant words and is retained as a negative result.

\paragraph{Signed-relation graph across different material cases.}
The second graph contains 72 prompts: six mechanism families, four material cases per family, and three surface variants per case (anchor, physical paraphrase, and lexical counterfactual). Every source node was restricted to targets in the same supplied mechanism family but from another material case. It selected one nearest neighbor in each of the other two surface-variant groups, giving $72\times2=144$ directed edges. A correct edge connected two cases with the same registered physical outcome after accounting for the counterfactual sign reversal. The primary similarity was the three-fit Jacobian cosine averaged over the fixed 38-92\% band. We applied the identical candidate set and edge rule to the frozen 80-96\% late band, direct decoder-basis states, raw states, word and character TF-IDF, an answer-order-only indicator, and a numeric-direction oracle.

The standard null independently permuted balanced outcome labels within mechanism and surface-variant blocks 50,000 times. Because an independent shuffle can break the designed relationship among the three variants of one material case, the stricter exact null operated on base cases: it enumerated all $6^6=46{,}656$ balanced assignments across the six families while keeping every anchor-paraphrase-counterfactual bundle and its registered sign transform intact. We additionally tabulated all nine ordered source-to-target surface-variant cells. Six non-diagonal ordered pairs were tested with blocked permutations; the four cells entering or leaving a counterfactual are the key sign-reversal falsification. Their $p$-values were adjusted together by the Benjamini-Hochberg procedure.

Edge precision uses only the highest-similarity candidate. To use the complete eligible ordering, we also calculated pairwise ROC-AUC, mean reciprocal rank of the best same-direction candidate, and leave-one-mechanism-out AUC. Family uncertainty used 50,000 family-bootstrap resamples. Matched method contrasts used all $2^6=64$ family sign assignments with a plus-one convention only where specified by the frozen analysis. All graph nodes, candidates, selected edges, similarities, labels, ranks, protocols, amendments, and validation checks are included in the Supplementary Information bundle.

\paragraph{Natural-question position audit.}
The original 72-prompt signed-relation graph used the final state after a prompt had displayed semantic answer choices. A separately frozen robustness run used the identical 72 scientific stems but stopped at the natural final token of the complete question. It prohibited answer choices, answer words, A/B codes, response-format instructions, and checkpoint markers. Raw residuals, direct decoder-basis states, and each of three Jacobian decoder-basis states were captured at the same 25 registered layers. The frozen primary similarity averaged the 38-92\% band; the edge rule, 144-edge count, all-candidate AUC, and 46,656-assignment case-preserving null were unchanged. Its strong gate required significant selected-edge precision and AUC together with at least four of six family AUCs above 0.5. This run reuses the inspected cohort and endpoint, so it is a positional robustness study rather than an independent replication.

The three-position comparison combines this natural question end with two states from the arbitrary-code cohort: the contextual \code{checkpoint} before the code mapping and the final prompt state after the mapping. Every plotted method uses the same 38-92\% layer band. Prompt suffix and token position therefore differ by design; the comparison maps sensitivity to realistic analysis boundaries but does not isolate a single linguistic cause.

\paragraph{Cross-mechanism counter-numeric falsification.}
The cross-mechanism analysis was frozen before any cross-mechanism statistic was computed, but after the natural-question cohort had been inspected. Every one of 72 sources was ranked against four cases in each of the five other mechanisms and each of three target surface variants, producing 1,080 rankings per method. Each query contained two same-outcome and two opposite-outcome candidates. Three mechanisms have a direct input-property response (crosslink density, dislocation density, particle fraction); three have an inverse response (obstacle spacing, pearlite spacing, porosity). The primary endpoints were mean pairwise AUC across all rankings and across the 648 rankings joining mechanisms with opposite response orientation. The exact null enumerated all $\binom{6}{3}=20$ balanced assignments of direct versus inverse orientation and derived every prompt label from numerical direction under that assignment; the smallest attainable exact $p$ was 0.05. Fifteen unordered mechanism pairs were the units for 30,000 fixed-seed bootstrap resamples. Success required both primary AUCs above 0.5 with exact $p\leq0.05$ and at least six of nine opposite-orientation pairs above 0.5.

\paragraph{Graph-identifiability theorem and audit chronology.}
The broader graph audit was designed after the positional and cross-mechanism results had been inspected and is therefore explicitly post hoc. Its component protocols were specified before their corresponding outputs were calculated. It introduces no new prompts and no new Gemma forward pass; it reuses all 72 natural questions, 25 registered layers, three Jacobian fits, direct decoder-basis states, and raw states without exclusion. For node $i$ in family $f$, numerical direction $x_{fi}\in\{-1,+1\}$, law orientation $s_f\in\{-1,+1\}$, and physical polarity $y_{fi}\in\{-1,+1\}$ were extracted directly from the frozen manifest. Exact enumeration confirmed $y_{fi}=s_fx_{fi}$ for all 72 nodes. Because $s_f^2=1$, the within-family pair label obeys $y_{fi}y_{fj}=x_{fi}x_{fj}$. This finite $Z_2$ symmetry establishes in advance which targets are identifiable from an unlabeled family graph.

The additional continuous cross-law audit was also post hoc and reused the same centered, row-normalized band states. For each of the nine direct-law-inverse-law mechanism pairs, it balanced the nine ordered surface-variant cells and subtracted mean similarity for same numerical direction but opposite physical outcome from mean similarity for the same physical outcome but opposite numerical direction. Thirty thousand mechanism-pair bootstrap resamples supplied the interval, and all $2^9$ sign assignments supplied the exact two-sided test. Identical calculations were applied to direct, raw, word TF-IDF, and character TF-IDF representations.

\paragraph{Label-blind matching and graph isomorphism.}
For each of the 15 unordered mechanism pairs, all $4!=24$ bijections between material cases were scored using only anchor-paraphrase similarities averaged over the registered depth band. The best map was fixed before it was evaluated on the held-out similarities involving counterfactuals and before physical labels were revealed. We retained held-out correlation, physical-label agreement, gauge-adjusted agreement, exact variant-labeled isomorphism, Weisfeiler-Lehman similarity, and three-cycle consistency. Constrained nulls preserved surface-variant and score-surface structure; the seven isomorphism-related tests were corrected together by Benjamini-Hochberg.

\paragraph{Whole-mechanism-held-out graph learning.}
A three-block graph isomorphism network was trained on four mechanisms, selected on a fifth, and tested on the completely unseen sixth, rotating through all six held-out families. Twenty seeds were ensembled after averaging the registered depth band. Each topology-only node input contained a constant, a three-way surface-variant indicator, the explicitly supplied numerical sign $x$, normalized depth, in/out degree, and incoming/outgoing edge strength. It contained no prompt text or embedding, mechanism identity, material identity, answer words, or physical label $y$. Matched graphless multilayer-perceptron and constrained edge-shuffle controls used the same inputs and splits. The numerical-direction configuration changed the target from $y$ to the already supplied $x$; its AUC is therefore an intentional input-readback positive control, not an independent topology endpoint. The scientific absolute-label endpoint asks whether graph structure plus the supplied numerical direction can recover physical polarity under a completely unseen law orientation. A separate polarity-free relation GIN classified all 72 eligible ordered pairs within each 12-node family graph: source and target had to be different material cases and different surface variants. It used only a constant, normalized depth, in/out degree, and edge strengths; it excluded surface variant and numerical direction as well as all semantic metadata. Thus the relation GIN asks whether graph topology alone reconstructs the within-mechanism same/different relation, whereas the numerical AUC merely checks an input supplied to the absolute model. All node-permutation audits were exact. In total, the dataset retains 1,320 absolute-label and 1,080 relation-model seeded fits, logits, validation selections, and held-out AUCs.

\paragraph{Spectral communities, sparsification, and exact partitions.}
Label-blind communities were obtained from the normalized graph Laplacian and scored with adjusted Rand index, which is invariant to swapping the two community names. The implementation was calibrated on 1,000 positive synthetic block graphs and 1,000 negative graphs. Four graph densities were frozen: binary top-one, weighted top-one, weighted top-two, and all eligible candidate affinities. Jacobian, direct, and raw representations produced 12 secondary density tests whose $p$-values were corrected together by Benjamini-Hochberg. To remove dependence on one clustering algorithm, we also evaluated every one of the $\frac{1}{2}\binom{12}{6}=462$ unique balanced partitions of each family graph. A held-out-surface endpoint fitted every balanced anchor/paraphrase partition and then assigned the counterfactual nodes. Its blockwise null preserved every ordered surface-variant block and its empirical weight distribution.

\subsection{Neutral-anchored relational constitutive benchmark}

\paragraph{Chronology and task construction.}
This benchmark was developed after the absolute-state and graph-identifiability results had been inspected. The fixed prompt scaffold asks the model to use the supplied equation to infer whether the response is direct, inverse, or unchanged; infer whether the numerical control rises or falls; silently compose those two signs; and output one allowed lowercase word. It supplies no intermediate answer or generated rationale. Method development used 16 earlier laws and four recorded prompt positions. A centroid direction fitted on labeled development laws was frozen before the 60-law evaluation. A first 16-law fresh cohort obtained mean absolute-state physical-outcome AUC 0.792 but failed the complete frozen gate because exact behavior was 71.5\% and the rearranged-formula AUC was 0.695. Its matched-reversal pattern motivated the relational endpoint. That endpoint was then frozen and passed on a disjoint 12-law cohort (direct-inverse AUC 0.972; exact balanced-label $p=0.004329$). A descriptive layer sweep on the same disjoint cohort, with a separate direction fitted only on the development laws at each depth, is reported in Supplementary Figure~S19; it does not use the final 60-law cohort. Only after the disjoint confirmation did we define the larger neutral-anchored protocol. Its 60 laws, prompts, layer, direction, normalization, endpoints, success gates, 50,000-bootstrap seed, and 100,000-permutation seed were specified before any corresponding model output.

The final cohort balances 20 direct, 20 inverse, and 20 physically neutral relations across 13 broad scientific domains. Each law is rendered with two algebraically equivalent surfaces, two material cases with different numerical scales, both numerical directions, and two answer orders, giving $60\times2\times2\times2\times2=960$ exact prompts and 480 matched comparisons. The two prompts in a matched cell have identical instructions, equation, material wording, endpoint values, and answer order; only the order of the numerical endpoints reverses. Ten neutral laws were designated for calibration before execution and ten different neutral laws for validation. No prompt or law was excluded after output.

\paragraph{Frozen state direction and matched contrast.}
Let $\tilde h=h/\|h\|_2$ be a unit-normalized raw residual state at the final prompt token after layer 34. On the 16-law development set only, let $\mu_+$ and $\mu_-$ be the centroids for positive and negative physical outcomes. The frozen direction, midpoint, and scalar readout are
\begin{equation}
d=\frac{\mu_+-\mu_-}{\|\mu_+-\mu_-\|_2},
\qquad
m=\frac{\mu_++\mu_-}{2},
\qquad
r(h)=d^\mathsf{T}(\tilde h-m).
\end{equation}
This is a raw-state centroid readout, not a Jacobian-transported or vocabulary-decoded state, and it is never refitted on the 60-law cohort. For law $f$ and matched cell $k$, the contrast is
\begin{equation}
\Delta r_{f,k}=r(h^{\mathrm{up}}_{f,k})-r(h^{\mathrm{down}}_{f,k}),
\qquad
\overline{\Delta r}_f=\frac{1}{8}\sum_{k=1}^{8}\Delta r_{f,k}.
\end{equation}
A direct law predicts $\overline{\Delta r}_f>0$, an inverse law predicts $\overline{\Delta r}_f<0$, and a neutral relation predicts no systematic change. Because raw zero need not transfer between prompt distributions, the calibration-neutral center is $c=\operatorname{median}_{f\in\mathcal N_c}\overline{\Delta r}_f$ and the robust scale is $s=1.4826\,\operatorname{median}_{f\in\mathcal N_c}|\overline{\Delta r}_f-c|$. The reported robust score is $(\overline{\Delta r}_f-c)/s$. Direct, inverse, and validation-neutral labels do not affect $c$ or $s$.

\paragraph{Endpoints, controls, and inference.}
Frozen primary endpoints were pairwise ROC-AUC for direct versus inverse, direct versus validation-neutral, and validation-neutral versus inverse laws; direct/inverse accuracy at the calibration-neutral median; direct-inverse AUC on each equation surface; and Spearman correlation between registered law sign $-1,0,+1$ and neutral-scaled contrast. Fifty thousand class-stratified law bootstraps supplied AUC intervals. The ordinal test used 100,000 fixed-seed label permutations with the plus-one convention and the directional alternative. Word and character TF-IDF controls were fitted on development prompts only, projected on a development physical-outcome centroid direction, and evaluated with the identical matched subtraction and neutral calibration. The output-head control used the clean final \code{higher}-minus-\code{lower} logit difference. Exact answer accuracy was reported only for direct and inverse laws; the neutral answer \code{unchanged} is multi-token under this tokenizer and was not part of that one-token behavior diagnostic. It does not affect any hidden-state endpoint.

\paragraph{\hlyellow{Equation-free prompt-format transfer.}}
\hlyellow{After the supplied-equation result was known, we constructed a controlled prompt ablation over the same 60 law identities. The prompt manifest and protocol were fixed before the corresponding model outputs. Every natural prompt removed the constitutive equation, law name, law-orientation label, two-stage monotonicity decomposition, and instruction to compose the law sign with the numerical sign. It retained the named response and control quantities, material scenario, two numerical endpoints, statement that all other relevant conditions were fixed, and the ordered one-word answer set. Two natural surface phrasings, two material cases, both numerical directions, and two answer orders again produced 960 prompts and 480 matched comparisons. No prompt or law was excluded after execution. This design isolates transfer across prompt format; because the law identities were reused, it is not a fresh-law generalization test.}

\hlyellow{We recorded the unit-normalized final-prompt-token residual at layer 34 and applied the unchanged development-law direction and midpoint defined above. The up-minus-down matched contrast and eight-cell law average were computed identically. The ten preassigned natural calibration-neutral laws established a natural center of 0.01124797 and robust scale of 0.00317143; the other ten neutral laws remained validation-only. Threshold-free class AUCs, neutral-cut directional accuracy, the two surface AUCs, ordinal Spearman correlation, 50,000 class-stratified bootstraps, and 100,000 fixed-seed label permutations followed the supplied-equation analysis. As a stricter cross-format check, directional accuracy was also calculated using the original supplied-equation neutral center and scale; it remained 38 of 40. Panel A of Figure~\ref{fig:relational-physics} uses that original center and scale for both 4B formats so that their displayed magnitudes have a common denominator. Panel B instead uses each condition's own calibration-neutral center and robust scale and independently orders its 60 laws; its horizontal axes are therefore condition-specific.}

\hlyellow{The output-head and development-trained word and character TF-IDF controls used the same matched subtraction and natural-neutral calibration. Greedy generation was allowed three tokens so that the multi-token answer \code{unchanged} could be evaluated. A completion counted only when its stripped lowercase text began with exactly \code{higher}, \code{lower}, or \code{unchanged}; all other continuations, including 45 that began an explanation despite the one-word instruction, were retained as incorrect. The complete prompt manifest, human-readable prompt inventory, raw continuations, answer logits, 960-by-2,560 state array, pair table, law table, controls, statistics, and validation reports are retained under \code{experiments/equation-free-relational-physics-2026-08-21/}.}

\paragraph{\hlyellow{Equation-free comparison across Gemma scales.}}
\hlyellow{The 12B and 31B runs reused the exact development and equation-free manifests above. At the prespecified nearest layer to 83\% normalized depth, each checkpoint supplied one unit-normalized final-prompt residual per prompt. A model-specific physical-outcome direction was defined as the positive-minus-negative centroid difference over the 128 development states and unit normalized; no 60-law test state contributed to that direction. Projection, matched up-minus-down subtraction, eight-cell law averaging, natural-neutral calibration, registered AUCs, surface and answer-order strata, ordinal permutation test, lexical controls, output-logit control, and strict three-token answer parsing were then repeated unchanged. The 12B protocol, model revision, layer, manifests, and success criteria were frozen before its model outputs. The cross-scale synthesis was specified after the completed 4B and 31B runs were available and is therefore a matched robustness analysis rather than a prospective scaling-law test.}

\hlyellow{Scale differences were paired by law identity. One hundred thousand fixed-seed bootstrap replicates independently resampled 20 direct, 20 neutral, and 20 inverse law identities with replacement and applied the same resampled indices to both checkpoints. We report percentile intervals for changes in the three AUCs and directional-law accuracy. Cross-scale law-score correlations use the raw mean matched contrast over the same 60 identities; category-centered correlations first subtract each checkpoint's direct, neutral, or inverse mean to determine whether agreement extends beyond the registered ordinal categories. The descriptive 31B depth comparison independently refits the development direction at layers 40 and 49 and evaluates the same equation-free cohort at each depth. It was defined after both depth runs existed and is not a continuous onset analysis.}

\hlyellow{The archived states and outputs reproduce the follow-up analysis, integrity audit, and combined Figure 10 without another model forward pass:}
\begin{lstlisting}[style=revisioncli]
python scripts/analyze_equation_free_relational_benchmark_amended.py
python scripts/validate_equation_free_relational_benchmark_postrun.py
python scripts/analyze_gemma_equation_free_multiscale.py
python scripts/analyze_gemma31b_equation_free_layers.py
python scripts/plot_gemma31b_equation_free_direction_rotation.py
\end{lstlisting}

The states reproduce the analysis, figure, and complete SI inventory without another model forward pass:
\begin{lstlisting}[style=cli]
python scripts/analyze_neutral_anchored_relational_benchmark.py
python scripts/build_relational_physics_si.py
\end{lstlisting}
To recapture the frozen layer-34 states from the model before running those two commands:
\begin{lstlisting}[style=cli]
conda activate PyTorch
python scripts/run_neutral_anchored_relational_benchmark.py
\end{lstlisting}

\subsection{Answer-scaffold and arbitrary-code audits}

The answer-scaffold comparison was defined after both lexical-adversarial cohorts had been inspected and is therefore post hoc. It combines two already stored states from the disjoint scientific stems. The pre-choice state is layer 39 (95.1\% depth) at a contextual \code{checkpoint} placed after the complete scientific question but before any answer mapping. The ordinary state is the final output state from the disjoint replication after the prompt supplied its semantic answer pair. Both were scored with the same family-specific positive-minus-negative decoder-logit contrast. Because a difference of two logits equals the log ratio of their softmax probabilities, all values in this audit are pairwise log-odds units, not centered-cosine units. Within each triplet, that contrast was signed so the anchor answer was positive, and relation separation was one half of the anchor-plus-paraphrase contrast minus the counterfactual contrast. The paired gain subtracts the pre-choice value from the ordinary final-state value. The same two-stage family/triplet bootstrap supplied descriptive intervals. Because layer, token position, and prompt suffix differ together, the comparison does not attribute the gain uniquely to answer-word exposure.

The pre-choice states came from a separately frozen arbitrary-code falsification. In that design, anchor and physical paraphrase shared a scientific answer but were assigned different A/B codes, while anchor and physical counterfactual had opposite scientific answers but both required A. The contextual checkpoint preceded the mapping and therefore could not attend to the future assignment. The first execution attempt found no marker because isolated and in-context tokenization differed; it stopped before any model forward pass. The failure, original protocol, and a tokenization-only amendment were preserved. The frozen study subsequently failed: forced-pair A/B accuracy was 62.5\%, only 5/24 triplets produced the complete registered \code{ABA} pattern, and A or B was never the global top next token. Its representational endpoints are retained in the Supplementary Information as an inconclusive negative control, not evidence for staged binding.

\subsection{Controlled recovery and population inference}

For each tokenizer-resolved declared term, we recorded its one-indexed full-vocabulary rank at every registered layer in the 38-92\% band and retained the best rank. At cutoff $k$, pass@$k$ equals the fraction of declared prompt-concept pairs with rank at most $k$. We evaluated $k\in\{1,2,5,10,20,50,100\}$ and calculated normalized trapezoidal area under pass@$k$ versus $\log k$. Jacobian AUC was first averaged over the three fitted lenses within prompt. Lens fits are repeated measurements, not population replicates.

The primary contrast is prompt-level Jacobian mean AUC minus direct AUC. The 20,000-resample hierarchical bootstrap samples mechanism families with replacement and then phrasings within sampled families. The exact sign-flip test enumerates all $2^{10}=1,024$ sign assignments of family-mean effects. Pairwise Spearman rank correlations use all 150 resolved prompt-concept rows; 5,000 bootstrap intervals resample whole families.

\hlyellow{For the secondary cross-scale audit in Supplementary Figure~S29, the exact 50 prompt strings, 150 predeclared one-token concepts, final-prompt-token position, normalized 38-92\% layer band, seven rank cutoffs, and direct-unembedding baseline were held fixed. The 4B and 12B Jacobian endpoints average three independently fitted lenses within prompt; only one 31B lens was available, so the 31B result does not estimate lens-fit variability. The same 20,000-resample family-hierarchical procedure supplies the displayed within-checkpoint intervals. This post-review analysis is descriptive and is separate from the equation-free relational scale comparison.}

A post hoc influence audit removed one complete mechanism family at a time, recomputed the equally weighted mean of the remaining nine family AUC differences, and recomputed all three pairwise lens-fit Spearman correlations over the remaining prompt-concept rows. It introduced no new threshold or hypothesis test.

\subsection{Complete-sequence scoring for multi-token technical terms}

The controlled rank endpoint excludes terms that Gemma tokenizes into multiple pieces because the rank of the first piece is not the probability of the complete word. A separately frozen robustness study scored two exact contrasts across all five held-out prompts in their respective families: \code{transgranular} versus \code{intergranular}, each split into two pieces, and \code{martensite} versus \code{bainite}, each split into three. At the original final-prompt position, the first-piece score was the direct or Jacobian logit; remaining pieces were scored by teacher-forced conditional log probability under the unchanged model. The sequence margin was the target complete-word score minus the equal-piece alternative. It was averaged over the registered 38-92\% band and then over five prompts per family. The frozen pass required positive family means for all three lens fits, positive Jacobian-minus-direct means for both families, and at least 8 of 10 positive prompt margins. All 10 prompts were retained; the observed 7 of 10 failed the breadth gate.

\subsection{Target-free candidate generation}

For each prompt and method, the algorithm scanned unrestricted top-1 decoded tokens over every prompt position and registered source layer in the fixed band. It removed tokens present in the input or one-token continuation and retained the highest-scoring 64 candidates per stored run. A prompt-level candidate survived only if it appeared in all three stored lens records. The same three-record consensus rule was applied to the matched direct-unembedding lists stored during those evaluations; direct scores themselves are identical across records, while prompt-level retained lists can reflect the common extraction pipeline.

After a frozen lowercase alphabetic filter and target-agnostic English function-word list, family candidates were ranked by the mean three-record consensus score multiplied by $\log(50/f_w)$, where $f_w$ is the number of prompts containing candidate $w$, and averaged over the five family phrasings. Declared terms were not used to generate, filter, retain, or rank candidates. Exact overlap was added only as a descriptive annotation after ranking.

\subsection{Target-free cross-phrasing classification}

This secondary analysis was specified after the held-out readout results were known but before its classifier output was calculated. Each prompt was represented by its complete filtered three-fit consensus candidate list. Candidate weights were the stored consensus score multiplied by an inverse-document-frequency factor calculated only from the 40 training prompts in that fold; the feature vocabulary was also restricted to training prompts. A cosine nearest-centroid classifier trained on four phrasings of each family and predicted the held-out fifth, yielding five balanced folds.

We evaluated Jacobian and direct candidates under three frozen filters: the original target-agnostic function-word filter; additional removal of exact lowercase prompt words; and additional removal of candidate/input pairs of at least five letters for which either word begins with the other. The last conservative rule removes simple morphological relatives when its prefix condition is met. A training-fold TF-IDF bag of input words and balanced label permutations served as baselines. We report accuracy, macro-F1, per-family confusion, exact paired McNemar tests, and 30,000 family-bootstrap intervals. Ten thousand label permutations independently shuffled the ten balanced family labels inside every phrasing fold and retained the maximum accuracy over the six readout/filter configurations; reported permutation values therefore account for filter selection. No declared term was used by the classifier.

\subsection{Semantic-stream construction}

Semantic streams are retrospective displays, not inferential endpoints. Figure~\ref{fig:streams} intentionally pairs two deterministic renderings of five selected frozen prompts. The unrestricted rendering uses fit 0 and assigns a layer score $(W-r)/W$ to each normalized leading token at rank $r$ in a stored list of width $W$; its seven displayed words have the largest depth-integrated score. The strict rendering first excludes input and continuation words, the frozen 214-word function list, and the globally common target-free scaffold. At each layer it intersects the three fits' remaining lists and scores a surviving word by its mean reciprocal within-list rank. Its seven displayed words have the largest integrated score inside the 38-92\% band. The five cases were selected after the population analyses to illustrate four physically informative neighborhoods and the cleavage failure case, so they support interpretation rather than a prevalence claim. Every selected word, prompt, score, and filter is stored in \code{experiments/materials-heldout-v1\_semantic\_streams.json}; the protocol is stored in \code{experiments/materials-heldout-v1-semantic-stream-protocol.md}. Ribbon heights are display scores with no probability interpretation.

\subsection{Exploratory latent geometry}

The inferential geometry protocol was frozen before hidden-vector extraction, but after the held-out vocabulary results were known. It is therefore post hoc exploratory. We extracted the raw final-prompt state $h_{p,\ell}$ for 50 prompts and 25 layers, then computed the L2-normalized transport
\begin{equation}
z_{s,p,\ell}=\frac{\mathrm{final\_norm}(J_{s,\ell}h_{p,\ell})}
{\|\mathrm{final\_norm}(J_{s,\ell}h_{p,\ell})\|_2}
\end{equation}
for lens fit $s$. We averaged the three normalized transports and renormalized. The array contains 3,750 transported 2,560-dimensional vectors, plus raw, target-layer, lexical, and discovered-word vectors.

At every layer, a nearest-centroid classifier used cosine similarity. The five folds are phrasing indices 1-5; in each fold, one phrasing from every family is held out and the other four form the ten class centroids. The raw state, target-layer state, and normalized mean input-token embedding use the same classifier. The corrected null independently permutes the ten balanced labels within each phrasing fold, calculates all 25 layer accuracies, and retains only the maximum. We used 5,000 permutations and a plus-one correction. We also calculated between-family divided by within-family cosine distance, pairwise lens-fit cosine distance, and alignment to frozen target-free word vectors.

For the display in Figure~\ref{fig:geometry}C, the 50 best-layer vectors were reduced to 25 principal components (85.1\% explained variance) and then embedded by UMAP with cosine metric, 10 neighbors, minimum distance 0.20, and fixed seed 20260715 \citep{mcinnes2018umap}. Five additional UMAP seeds and a two-dimensional PCA sensitivity plot are recorded. The original frozen projection plan used a joint all-layer display; after that diagnostic was dominated by layer progression, we chose the best-layer view for legibility. This post hoc visual choice does not affect any classification or permutation result.

\subsection{Automated blinded secondary rating}

The 20 candidate sets (ten per method) were shuffled, assigned opaque identifiers, and separated from the answer key before rating. Five order-randomized passes used the OpenAI Responses API model \code{gpt-5.5}. Each pass received candidate words with support counts and the ten allowed family labels, and returned one label per set. It did not receive prompts, method identities, or the key. Majority accuracy, individual-pass accuracy, Fleiss $\kappa$, a 100,000-shuffle label null, and all $2^{10}=1,024$ paired family swaps were prespecified in the automated secondary script. Raw prompts, raw responses, parsed labels, randomization seeds, and the initial unscored operational failure are retained. These are repeated judgments from one automated system and cannot be interpreted as five independent experts.

\section*{Data and code availability}
Code, exact prompt manifests, versioned protocols, raw model outputs, analysis scripts, figure-generation code, and manuscript sources are retained in the accompanying \code{https://github.com/lamm-mit/Substrates} repository. Versioned machine-readable manifests link each reported result to its corresponding experimental inputs and outputs. The three fitted lens bundles are stored in the Hugging Face repository \code{lamm-mit/gemma4-jacobian-lenses}. The raw, transported, target, lexical, and word vectors used in the exploratory latent-geometry analysis, together with their metadata, exact prompt manifest, and protocol, are stored in the public Hugging Face dataset \code{lamm-mit/gemma4-materials-latent-vectors}. The Supplementary Information includes additional details including prompts. Prompts are also available via \code{lamm-mit/gemma4-materials-mechanism-prompts} for easy access.


\section*{Acknowledgements}

\hlyellow{This work was supported by MIT's Generative AI Impact Consortium (MGAIC).}

\section*{Author contributions}
M.J.B. conceived and designed the study, developed the methodology and software, analyzed and interpreted the results, prepared the visualizations, and wrote and revised the manuscript.

\section*{Competing interests}
The author declares no competing interests.

\bibliographystyle{naturemag}

\bibliography{references}

\end{document}